\documentclass[12pt]{report}
\usepackage[dvipsnames]{xcolor} 
\usepackage{def}  
\usepackage{amsmath,amsfonts,amsthm}
\usepackage{booktabs}       
\usepackage{graphicx,subcaption,caption}
\graphicspath{{Chapter3/images/}{Chapter4/images/}{Chapter5/images/}{Chapter6/images/}{Chapter7/images/}}
\usepackage{fancyhdr}
\usepackage{lscape}
\usepackage{stfloats}
\usepackage{url}
\usepackage{indentfirst}
\usepackage{multirow}
\usepackage{array}
\usepackage{rotating}
\usepackage{bm}
\usepackage{setspace}
\usepackage{tikz}
\usepackage{algorithm}
\usepackage{algorithmic}
\usepackage{longtable}
\usepackage{caption}
\usepackage{comment}
\usepackage{xspace}
\usepackage{microtype}
\usepackage{adjustbox}
\usepackage{tabularx}
\usepackage[colorinlistoftodos]{todonotes}

\usepackage[normalem]{ulem}

\usepackage[numbers]{natbib}
\usetikzlibrary{shapes,arrows,positioning}
\newcommand{\PreserveBackslash}[1]{\let\temp=\\#1\let\\=\temp}
\newcolumntype{C}[1]{>{\PreserveBackslash\centering}p{#1}}
\newcolumntype{L}[1]{>{\PreserveBackslash}p{#1}}
\newcolumntype{R}[1]{>{\PreserveBackslash\flushright}p{#1}}

\addtocounter{secnumdepth}{1}
\addtocounter{tocdepth}{1}

\newcommand{\namelasyn}{LaSyn\xspace}

\newcommand{\rev}[1]{#1}

\newcommand{\del}[1]{}

\newcommand{\namens}{JoGANIC}
\newcommand{\name}{\namens\xspace}
\newcommand{\fullnamens}{Journalistic Guideline Aware News Image Captioning}
\newcommand{\fullname}{\fullnamens\xspace}
\newcommand{\tellns}{Tell}

\newcommand{\tellfullns}{Transform and Tell}
\newcommand{\tellfull}{\tellfullns\xspace}

\newcommand{\neeshort}{NEE\xspace}
\newcommand{\neeshortns}{NEE}
\newcommand{\good}{GoodNews\xspace}
\newcommand{\nyt}{NYTimes800k\xspace}

\begin{document}

\title{Exploring External Knowledge for Accurate modeling of Visual and Language Problems}

\author{Xuewen Yang}

\gradyear{2021}

\convocationdate{August 2021}

\degree{Doctor of Philosophy}

\fieldofstudy{Electrical Engineering}

\fileforabstract{abstract}
\fileforacknowledgement{acknowledgement}

\doublespacing

\begin{singlespace}
\prefatorypages
\end{singlespace}

\setlength{\headheight}{15.2pt}
\setlength{\baselineskip}{22pt}

\chapter{Introduction and Background} \label{ch:Background}


\section{Artificial Intelligence}
We humans have the nature or intelligence to acquire new knowledge and skills.
This intelligence has helped us to complete our most important achievements, from hunting and farming in the wild to modern medicine and spacecraft, by creating handful tools like bows and arrows, machines, microscopes, robots, etc.
However, we humans also have the innate limitations, such as we are prone to be tired.
It's getting more and more difficult for us to keep updated with the massive data everyday.
The question for now is ``what if we apply our intelligence to create new tools that can learn from the increasing amount of data in more efficient and effective ways?''
This question is central to the field of artificial intelligence. 
In the rapidly developing subfield of machine learning, specifically, we set out to acquire new knowledge and skills to build machines that themselves can acquire new knowledge and
skills. 
The aim of this work is further advancement of the field of artificial intelligence, and through this means, to increase the probability of a future
that is bright.

\section{Deep Generative Models}
As Richard Feynman said ``What I cannot create, I do not understand'', for a model to understand the visual and language world, it has to be capable to create new things based on what it has.
Generative models are the models that are able to achieve this goal.
The term ``generative model'', in this work, refer to any model that takes training data, consisting of samples drawn from a distribution $p_{data}$, and learns to represent an estimate of that distribution somehow.
The result is a probability distribution $p_{model}$.

One might legitimately wonder why generative models are worth studying and why they are important to AI, especially generative models that are only capable of generating data rather than providing an estimate of the density function. 
After all, when applied to images, such models seem to merely generate more images, and the world has no shortage of images.
There are several reasons to study generative models.
In this work, the author lists three of them, including:
\begin{itemize}
    \item Training and sampling from generative models is an excellent test of our ability to represent and manipulate high-dimensional probability distributions, which are important objects in a wide variety of applied math and engineering domains.
    \item Generative models can be trained with missing data and can provide predictions on inputs that are missing data. One particularly interesting case of missing data is semi-supervised learning, in which the labels for many or even most training examples are missing.
    Generative models are able to perform semi-supervised learning reasonably well.
    \item Many tasks intrinsically require realistic generation of samples from some distribution, including image super-resolution and image-to-image translation applications, etc.
\end{itemize}

All of these and other applications of generative models provide compelling
reasons to invest time and resources into improving generative models.
In this work, we target on image-to-image translation and how to use domain labels as the external knowledge to solve the multi-domain image-to-image translation problem.


\section{Sequence to Sequence Models}

Despite their flexibility and power, Convolutional Neural Networks (CNNs) \cite{xuewen1,xuewen2,xuewen3,xuewen4,xuewen5,xuewen6,xuewen7,xuewen8,xuewen9,xuewen10,xuewen11,xuewen12,xuewen13,xuewen14,xuewen15} can only be applied to problems whose inputs and targets can be sensibly encoded with vectors of fixed dimensionality. 
It is a significant limitation, since many important problems are best expressed with sequences whose lengths are not known a-priori.
For example, speech recognition and machine translation are sequential problems. 
Likewise, question answering can also be seen as mapping a sequence of words representing the question to a sequence of words representing the answer. 
It is therefore clear that a domain-independent method that learns to map sequences to sequences would be useful.

A straightforward application of the Recurrent Neural Networks (RNNs), including the following Long Short-Term Memory (LSTM) architectures and Transformers can solve general sequence to sequence problems.
The idea is to use one RNN to read the input sequence, one time step at a time, to obtain large fixed dimensional vector representation, and then to use another RNN to extract the output sequence from that vector. 
The second RNN is thus conditioned on the input sequence. 
In the current applications, LSTMs and Transformers are more widely used than the vanilla RNNs because of their ability to successfully learn on data with long range temporal dependencies, which makes them a natural choice for this sequence to sequence application due to the considerable time lag between the inputs and their corresponding outputs.


\chapter{Image-to-Image Translation}
\section{Introduction}

In this work, we define multi-domain as multiple datasets or several subsets of one dataset that are applied to complete the same task, but these datasets (or subsets) have different statistical biases. As some examples, images taken at Alps in the summer  and in the winter are considered as two different domains, while faces with hair and faces with eyeglasses form another two different domains. Under this domain definition, for faces with black hair and faces with yellow hair, the black hair and yellow hair are two different \textit{attributes} of the same domain. In \textit{multi-domain learning}, each sample $\boldsymbol{x}$ is drawn from a domain $d$ specific distribution $\boldsymbol{x}\sim p_{d}(\boldsymbol{x})$ and has a label $y\in \{0,1\}$, with $y=1$ signifying $\boldsymbol{x}$ from domain $d$, $y=0$ signifying $\boldsymbol{x}$ not from domain $d$.

Image-to-image translation is the task of learning to map images from one domain to another, e.g., mapping grayscale images to color images \cite{DBLP:conf/pkdd/CaoZZY17}, mapping images of low resolution to images of high resolution \cite{Ledig17}, changing the seasons of scenery images \cite{8237506}, and reconstructing photos from edge maps \cite{Isola2017ImagetoImageTW}. The most significant improvement in this research field came with the application of Generative Adversarial Networks (GANs) \cite{NIPS2014_5423,NIPS2016_6544}.

\begin{figure}[!t]
\begin{minipage}{1\linewidth}
\centering
\subfloat[]{\label{m:a}\includegraphics[scale=.25]{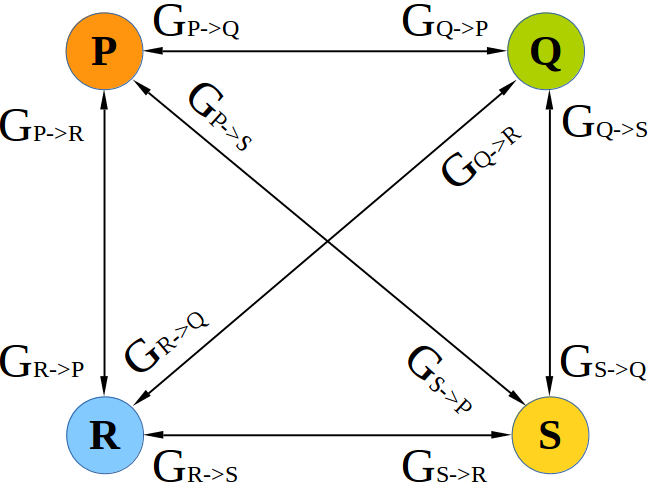}}
\subfloat[]{\label{m:b}\includegraphics[scale=.25]{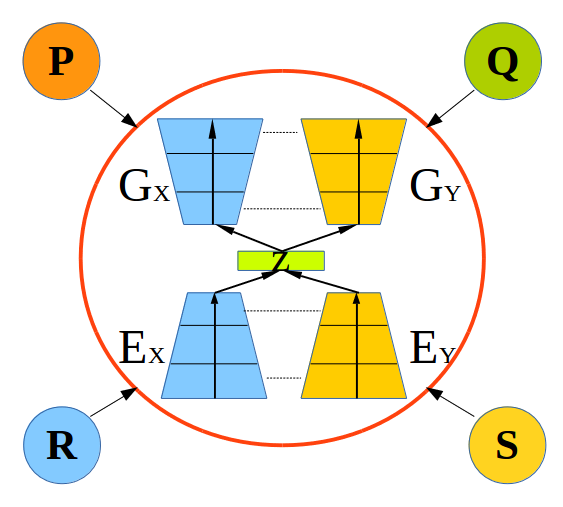}}
\end{minipage}\par
\caption{Image-to-image translation of 4 domains. (a) CycleGAN needs $4\times 3$ generators. (b) Our model only needs 2 encoder-generator pairs. In every iteration, we randomly pick two domains, and sample two batches of training data from the domains to train the model. The two encoders first encode domain information into a latent code $\boldsymbol{z}$ using two encoders $E_X$ and $E_Y$ and then generate two samples of the two domains using the generators $G_X$ and $G_Y$.}
\label{different_models}
\end{figure}

The image-to-image translation can be performed in supervised \cite{Isola2017ImagetoImageTW} or unsupervised way \cite{8237506}, with the unsupervised one becoming more popular since it does not need to collect ground-truth pairs of samples. Despite the quick progress of research on image-to-image translation, state-of-the-art results for unsupervised translation are still not satisfactory. In addition, existing research generally focuses on image-to-image translation between two domains, which is limited by two drawbacks. First, the translation task is specific to two domains, and the model has to be retrained when there is a need to perform image translation between another pair of similar domains. Second, it can not benefit from the features of multiple domains to improve the training quality. We take the most representative work in this research field CycleGAN \cite{8237506} as an example to illustrate the first limitation. The translation between two image domains $X$ and $Y$ is achieved with two generators, $G_{X\rightarrow Y}$ and $G_{Y\rightarrow X}$. However, this model is inefficient in completing the task of multi-domain image translation. To derive mappings across all $n$ domains, it has to train $n(n-1)$ generators, as shown in Fig.~\ref{m:a}.

To enable more efficient multi-domain image translation with unsupervised learning where image pairing across domains is not predefined, we propose Crossing-Domain GAN (CD-GAN), which is a multi-domain encoding generative adversarial network that consists of a pair of encoders and a pair of generative adversarial networks (GANs). We would like the encoders to efficiently encode the information of all domains to form a high-level feature space with an encoding process, then images of different domains will be translated by decoding the high-level features with a decoding process. CD-GAN achieves this goal with the integrated use of three techniques. First, the two encoders are constrained by a \textit{weight sharing} scheme, where the two encoders (or the two generators) share the same weights at both the highest-level layers and the lowest-level layers.
This ensures that the two encoders can encode common high-level semantics as well as low-level details to obtain the feature space, based on which generators can decode the high-level semantics and low-level details correctly to generate images of different domains. 
Second, we use a selected or existing label to guide the generator to generate images of a corresponding domain from the high-level features learnt. Third, we propose an efficient training algorithm that jointly train the model across domains by randomly selecting two domains to train at each iteration.

Different from ~\cite{NIPS2016_6544} where only weights at high-level layers of generators are shared, in CD-GAN, we propose the concurrent sharing of the lowest-level and the highest-level layers at both the encoders and the generators to improve the quality of image translation between {\em any two} domains. The sharing of highest layers between two encoders helps to enable more flexible cross-domain image translation, while the sharing of the lowest layers across domains helps improve the training quality by taking advantage of the transferring learning across domains.

The contributions of our work are as follows:
\begin{itemize}
\item We propose CD-GAN that learns mappings across multiple domains using only two encoder-generator pairs.
\item We propose the concurrent use of weight-sharing at highest-level and lowest-level layers of both encoders and generators to ensure that CD-GAN generates images with sufficient useful high-level semantics and low-level details across all domains.
\item We leverage domain labels to make a conditional GAN training that greatly improves the performance of the model.
\item We introduce a cross-domain training algorithm that efficiently and sufficiently trains the model by randomly taking samples from two of domains at a time. CD-GAN can fully exploit data from all domains to improve the training quality for each individual domain.
\end{itemize}

Our experiment results demonstrate that when trained on more than two domains, our method achieves the same quality of image translation between any two domains as compared to directly training for translation between the pair. However, our model is established with much less training time and can generate better quality images for a given amount of time. We also show how CD-GAN can be successfully  applied  to a variety of unsupervised multi-domain image-to-image translation problems.

The remainder of this chapter is organized as follows. Section~\ref{relatedwork} reviews the relevant research for image-to-image translation problems. Section~\ref{mod} describes our model and training method in details. Section~\ref{exp} presents our evaluation metrics, experimental methodology, and the evaluation results of the model's accuracy and efficiency on different datasets. Finally, we discuss some limitations of our work and conclude our work in Section~\ref{conclusion}.

\section{Related Work}
\label{relatedwork}

\subsection{Generative Adversarial Networks (GANs)}

GANs \cite{NIPS2014_5423} were introduced to model a data distribution using independent  latent variables. Let $\boldsymbol{x} \sim p(\boldsymbol{x})$ be a random variable representing the observed data and $\boldsymbol{z} \sim p(\boldsymbol{z})$ be a latent variable. The observed variable is assumed to be generated by the latent variable, i.e., $\boldsymbol{x} \sim p_{\boldsymbol{\theta}}(\boldsymbol{x} \vert \boldsymbol{z})$, where $p_{\boldsymbol{\theta}}(\boldsymbol{x} \vert \boldsymbol{z})$ can be explicitly represented by a generator in GANs. GANs are built on top of neural networks, and can be trained with gradient descent based algorithms.

The GAN model is composed of a discriminator $D_{\boldsymbol{\phi}}$, along with the generator $G_{\boldsymbol{\theta}}$. The training involves a min-max game between the two networks. The discriminator $D_{\boldsymbol{\phi}}$ is trained to differentiate `fake' samples generated from the generator $G_{\boldsymbol{\theta}}$ from the `real' samples from the true data distribution $p(\boldsymbol{x})$. The generator is trained to synthesize samples that can fool the discriminator by mistaking the generated samples for genuine ones. They both can be implemented using neural networks.

At the training phase, the discriminator parameters $\boldsymbol{\phi}$  are firstly updated, followed by the update of the generator parameters $\boldsymbol{\theta}$. The objective function is given by:

\begin{equation}
\begin{aligned}
\min_{\boldsymbol{\theta}}\max_{\boldsymbol{\phi}}V(D,G)=&\mathbb{E}_{x\sim p(x)}[\log D_{\boldsymbol{\phi}}(x)] \\
& + \mathbb{E}_{\boldsymbol{z}\sim p(\boldsymbol{z})}[\log (1-D_{\boldsymbol{\phi}}(G_{\boldsymbol{\theta}}(\boldsymbol{z})))]
\end{aligned}
\end{equation}

The samples can be generated by sampling $\boldsymbol{z}\sim p(\boldsymbol{z})$, then $\hat{\boldsymbol{x}}=G_{\boldsymbol{\theta}}(\boldsymbol{z})$, where $p(\boldsymbol{z})$ is a prior distribution, for example, a multivariate Gaussian.

\subsection{Image-To-Image Translation}

\textit{Image-to-image translation} problem is a kind of image generation task that given an input image $\boldsymbol{x}$ of domain \textit{X}, the model maps it into a corresponding output image $\boldsymbol{y}$ of another domain \textit{Y}. It learns a mapping between two domains given sufficient training data \cite{Isola2017ImagetoImageTW}. Early works on image-to-image translation mainly focused on tasks where the training data of domain $X$ are similar to the data of domain $Y$ \cite{Gupta:2012:ICU:2393347.2393402,Liu:2008:IC:1409060.1409105}, and the results were often unrealistic and not diverse.

In recent years, deep generative models have shown increasing capability of synthesizing diverse, realistic images that capture both fine-grained details and global coherence of natural images \cite{2015arXiv150204623G,2015arXiv151106434R,Kingma2014}.
With Generative Adversarial Networks (GANs) \cite{Isola2017ImagetoImageTW,8237506,pmlrv70kim17a}, recent studies have already taken significant steps in image-to-image translation. In \cite{Isola2017ImagetoImageTW}, the authors use a conditional GAN on different image-to-image translation tasks, such as synthesizing photos from label maps and reconstructing objects from edge maps. However, this method requires input-output image pairs for training, which is in general not available in image-to-image translation problems. For situations where such training pairs are not given, in \cite{8237506}, the authors proposed CycleGAN to tackle unsupervised image-to-image translation. With a pair of Generators $G$ and $F$, the model not only learns a mapping $G: X\rightarrow Y$ using an adversarial loss, but constrains this mapping with an inverse mapping $F: Y\rightarrow X$. It also introduces a cycle consistency loss to enforce $F(G(X)) \approx X$, and vice versa. In  settings where paired training data are not available, the authors showed promising qualitative results. The authors in \cite{pmlrv70kim17a} and \cite{8237572} use similar idea to solve the unsupervised image-to-image translation tasks.

These approaches only tackle the problems of translating images between two domains, and have two major drawbacks. First, when applied to $n$ domains, these approaches need $n(n-1)$ generators to complete the task, which is computationally inefficient. To train all models, it would either require a significant amount of time to complete if the training is performed on one GPU, or it will require a lot of hardware and computing resources if training is run over multiple GPUs.
Second, as each model is trained with only two datasets, the training cannot benefit from the data of other domains.

Our work is inspired by \textit{multimodal learning} \cite{Ngiam:2011:MDL:3104482.3104569}, which shows that data features can be better extracted using one modality if multiple modalities are present at feature learning time. The intuition of our method is that if we can encode the information of different domains together and generate a high-level feature space, it would be possible to decode the high-level features to build images of different domains. 
In this work, rather than generating images from random noise, we incorporate an encoding process into a GAN model. The image-to-image translation can be achieved by first encoding real images into high-level features, and then generating images of different domains using the high-level features through a decoding process. The encoding process and the decoding process are constrained by a weight-sharing technique that both the highest layer and the lowest layer are shared across the two encoders as well as the two generators. Sharing the high-level layers makes sure that the generated images are semantically correct, while sharing the low-level layers ensures that important low-level features be captured and transferred between domains. Our model is trained end-to-end using data from all $n$ domains.

\section{Background}

\section{Cross-Domain Generative Adversarial Network}
\label{mod}

To conduct unsupervised multi-domain image-to-image translation, a direct approach is to train a CycleGAN for every two domains. While this approach is straightforward, it is inefficient as the number of  training models increases quadratically with the number of domains. If we have $n$ domains, we have to train $n(n-1)$ generators, as shown in Fig.~\ref{m:a}. In addition, since each model only utilizes data from two domains \del{$X$ and $Y$} to train, the training cannot benefit from the useful features of other domains.

To tackle these two problems, a possible way is to encode useful information of all domains into common high level features, and then to decode the high-level features into images of different domains. Inspired by work \cite{Ngiam:2011:MDL:3104482.3104569} from \textit{mutimodal learning}, where training data are from multiple modalities, we propose to build a multi-domain image translation model that can encode information of multiple domains into a set $Z$ of high-level features, and then use features in $Z$ to reconstruct data of different domains or to do image-to-image translation. The overview of the model applied to 4 domains is shown in Fig.~\ref{m:b}, where only one model is used.

In this section, we first present our proposed CD-GAN model, then describe how  image translation can be performed across domains, and finally introduce our cross-domain training method. 





\subsection{CD-GAN with Double Layer Sharing}

We first describe how to apply our model to multi-domain image-to-image translation in general then illustrate it using two domains as an example.
As shown in Fig.~\ref{mm:b}, our proposed CD-GAN model consists of a pair of encoders followed by a pair of  GANs. Taking domain $X$ and $Y$ as an example, the two encoders $E_X$ and $E_Y$ encode domain information from $X$ and $Y$ into a set of high-level features contained in a set $Z$. Then from a high-level feature $z$ in space $Z$,
we can generate images that fall into domain $X$ or $Y$. The generated images are then evaluated by the corresponding discriminators $D_X$ and $D_Y$ to see whether they look real and cannot be identified as generated ones. For example, following the red arrows, the input image $\boldsymbol{x}$ is first encoded into a high-level feature $\boldsymbol{z}_x$, then $\boldsymbol{z}_x$ is decoded to generate the image $\hat{\boldsymbol{y}}$. The image $\hat{\boldsymbol{y}}$ is the translated image in domain $Y$. Similar processes exist for image $\boldsymbol{y}$.


Our model is also constrained by a reconstruction process shown in Fig.~\ref{mm:c}. For example, following the red arrows, the input image $\boldsymbol{x}$ is first encoded into a high-level feature $\boldsymbol{z}_x$, then $\boldsymbol{z}_x$ is decoded to generate the image $\boldsymbol{x}\prime$, which is a reconstruction of the input image. Similar processes exist for image $\boldsymbol{y}$.

Learning with deep neural networks involves hierarchical feature representation.  In order to support flexible cross-domain image translation and also to improve the training quality, we propose the use of {\em double-layer sharing} where the highest-level and the lowest-level layers of the two encoders share the same weights and so does the two generators. By enforcing the layers that decode high-level features in GANs to share weights, the images generated by different generators can have some common high-level semantics. The layers that decode low-level details then map the high-level features to images in individual domains.

Sharing weights of low-level layers has the benefit of transferring low-level features of one domain to the other, thus making the image-to-image translation more close to real images in the respective domains. Besides, sharing layers reduces the complexity of the model, making it more resistant to the over-fitting problem.


\subsection{Conditional Image Generation}
\label{cig}
In state-of-the-art techniques, like CycleGAN, each domain is described by a specific generator, thus there is no need 
to inform the generator which domain the input image is generated to. However, in our model, multiple domains share two generators. For an input image, we have to include an auxiliary variable to guide the generation of image for a specific domain. The only information we have is the domain labels. To make use of this information, the inputs of the model are not images $\boldsymbol{x}$, $\boldsymbol{y}$, but image pairs $(\boldsymbol{x}, \boldsymbol{l}_y)$ and $(\boldsymbol{y}, \boldsymbol{l}_x)$ where the labels $\boldsymbol{l}_y$ and $\boldsymbol{l}_x$ inform the generators which domains to generate an image for. These image pairs are not the same as the image pairs of supervised image-to-image generation tasks, which are $(\boldsymbol{x}, \boldsymbol{y})$. Thus no matter which domain images are the input, the model can always generate images of a domain of interest.


\begin{figure}[!t]
\begin{minipage}{1\linewidth}
\centering
\subfloat[]{\label{mm:b}\includegraphics[scale=.25]{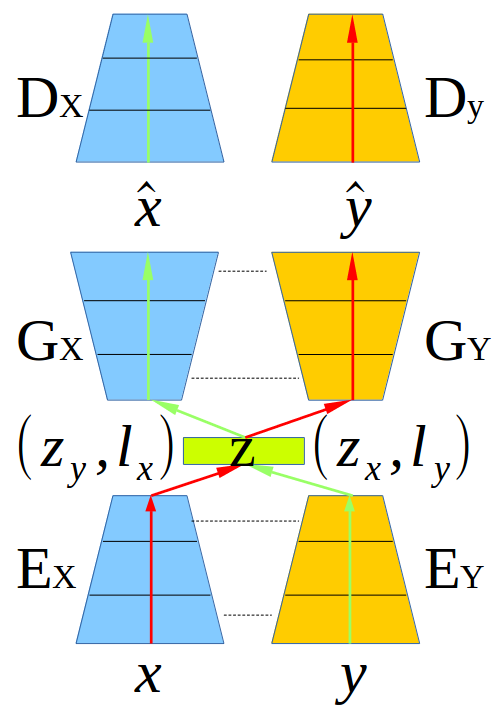}}
\subfloat[]{\label{mm:c}\includegraphics[scale=.25]{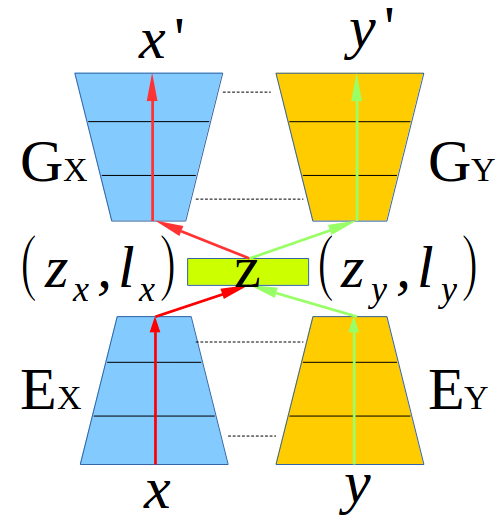}}
\end{minipage}\par
\caption{The proposed CD-GAN model. (a) The translation mappings: the input image $\boldsymbol{x}$ is first encoded as a latent code $\boldsymbol{z}_x$ through $E_X(\boldsymbol{x})$, which is then decoded into a translated image $\hat{\boldsymbol{y}}$ through $G_Y(\boldsymbol{z}_x, \boldsymbol{l}_y)$. The process is identified with red arrows. There is a similar process for the image $\boldsymbol{y}$. $D_X$ and $D_Y$ are adversarial discriminators for the respective domains to evaluate whether the translated images are realistic. (b) The reconstruction mappings: the input image $\boldsymbol{x}$ is first encoded as a latent code $\boldsymbol{z}_x$ through $E_X(\boldsymbol{x})$, which is then decoded into a reconstructed image $\boldsymbol{x}\prime$ through the generator $G_X(\boldsymbol{z}_x, \boldsymbol{l}_x)$. The process is signified in red arrows. A similar process exists for image $\boldsymbol{y}$. Note: the dashed lines indicate that the two layers share the same parameters.}
\label{model:loss}
\end{figure}

We denote the data distributions as $\boldsymbol{x}\sim p(\boldsymbol{x})$ and $\boldsymbol{y}\sim p(\boldsymbol{y})$. As illustrated in Fig. \ref{model:loss}, our model includes four mappings, two translation mappings $X\rightarrow Z\rightarrow Y$, $Y\rightarrow  Z\rightarrow X$ and two reconstruction mappings $X\rightarrow Z \rightarrow X$, $Y\rightarrow Z\rightarrow Y$. The translation mappings constrain the model by a GAN loss, while the reconstruction mappings constrain the model by a reconstruction loss. To further constrain the auxiliary variable, we introduce a classification loss by applying a classifier to classify the real or generated images into different domains. The intuition is that if images are generated with the guidance of the auxiliary variable, then it can be correctly classified into the domain specified by the auxiliary variable. Next, we introduce these model losses in more details as follows.

\textbf{GAN Losses} Following the translation mapping $X\rightarrow Z\rightarrow Y$, we can translate image $\boldsymbol{x}$ from domain $X$ to $\hat{\boldsymbol{y}}$ of domain $Y$ using $\boldsymbol{z}_x = E_{X}(\boldsymbol{x})$, $\hat{\boldsymbol{y}}=G_{Y}(\boldsymbol{z}_x, \boldsymbol{l}_y)$. With the purpose of improving the quality of the generated samples, we apply adversarial loss. We express the objective as:

\begin{equation}
\begin{aligned}
\mathcal{L}_{GAN_{Y}}&=\mathbb{E}_{\boldsymbol{y}\sim p(\boldsymbol{y})}\log (D_{Y}(\boldsymbol{y})) \\ &
 + \mathbb{E}_{\boldsymbol{x}\sim p(\boldsymbol{x})}\log (1-D_{Y}(G_{Y}(E_{X}(\boldsymbol{x}),\boldsymbol{l}_y)))
\end{aligned}
\end{equation}
where $G_Y$ tries to generate images $\hat{\boldsymbol{y}}=G_{Y}(\boldsymbol{z}_x,\boldsymbol{l}_y)$ that look similar to images from domain $Y$, while $D_{Y}$ aims to distinguish between translated samples $\hat{\boldsymbol{y}}$ and real samples $\boldsymbol{y}$. The similar adversarial loss for $Y\rightarrow  Z\rightarrow X$ is

\begin{equation}
\begin{aligned}
\mathcal{L}_{GAN_{X}}&=\mathbb{E}_{\boldsymbol{x}\sim p(\boldsymbol{x})}\log (D_{X}(\boldsymbol{x})) \\ &
 + \mathbb{E}_{\boldsymbol{y}\sim p(\boldsymbol{y})}\log (1-D_{X}(G_{X}(E_{Y}(\boldsymbol{y}),\boldsymbol{l}_x)))
\end{aligned}
\end{equation}

The total GAN loss is:
\begin{equation}
\mathcal{L}_{GAN}=\mathcal{L}_{GAN_{X}} + \mathcal{L}_{GAN_{Y}}
\end{equation}

\textbf{Reconstruction Loss}
The reconstruction mappings $X\rightarrow Z \rightarrow X$, $Y\rightarrow Z\rightarrow Y$ encourage the model to encode enough information to the high-level feature space $Z$ from each domain. The input can then be reconstructed by the generators. The reconstruction process of domain $X$ is $\boldsymbol{z}_x = E_{X}(\boldsymbol{x})$, $\boldsymbol{x}\prime=G_{X}(\boldsymbol{z}_x, \boldsymbol{l}_x)$. Similar reconstruction process exists for domain $Y$. With $l_2$ distance as the loss function, the reconstruction loss is:

\begin{equation}
\begin{aligned}
\mathcal{L}_{rec}&=\mathbb{E}_{\boldsymbol{x}\sim p(\boldsymbol{x})}(\vert \vert \boldsymbol{x} - G_{X}(E_{X}(\boldsymbol{x}),\boldsymbol{l}_x) \vert \vert_{2}) \\
&+ \mathbb{E}_{\boldsymbol{y}\sim p(\boldsymbol{y})}(\vert \vert \boldsymbol{y} - G_{Y}(E_{Y}(\boldsymbol{y}),\boldsymbol{l}_y) \vert \vert_{2})
\end{aligned}
\end{equation}

\textbf{Latent Consistency Loss}
With only the above losses, the encoding part is not well constrained. We constrain the encoding part using a latent consistency loss. Although $\boldsymbol{x}$ is translated to $\hat{\boldsymbol{y}}$, which is in domain $Y$, $\hat{\boldsymbol{y}}$ is still semantically similar to $\boldsymbol{x}$. Thus, in the latent space $Z$, the high-level feature of $\boldsymbol{x}$ should be close to that of $\hat{\boldsymbol{y}}$. Similarly, the high-level feature of $\boldsymbol{y}$ in domain $Y$ should be close to the  high-level feature of $\hat{\boldsymbol{x}}$ in domain $X$.  
The latent consistency loss is the following:

\begin{equation}
\begin{aligned}
\mathcal{L}_{lcl}&=\mathbb{E}_{\boldsymbol{x}\sim p(\boldsymbol{x})}(\vert \vert E_{X}(\boldsymbol{x}) - E_{Y}(G_{Y}(E_{X}(\boldsymbol{x}),\boldsymbol{l}_y)) \vert \vert) \\
&+\mathbb{E}_{\boldsymbol{y}\sim p(\boldsymbol{y})}(\vert \vert E_{Y}(\boldsymbol{y}) - E_{X}(G_{X}(E_{Y}(\boldsymbol{y}),\boldsymbol{l}_x)) \vert \vert)
\end{aligned}
\end{equation}

\textbf{Classification Loss}
We consider $n$ domains as $n$ categories in the classification problems. We use a network $C$, which is an auxiliary classifier, on top of the general discriminator $D$ to measure whether a sample (real or generated) belongs to a specific fine-grained category. The output of the classifier $C$ represents the posterior probability $P(c\vert \boldsymbol{x})$. Specifically, there are four classification losses, i.e., for real data $\boldsymbol{x}$, $\boldsymbol{y}$, and generated data $\hat{\boldsymbol{x}}$, $\hat{\boldsymbol{y}}$. For image-label pairs ($\boldsymbol{x}$, $\boldsymbol{l}_{x}$) and ($\boldsymbol{y}$, $\boldsymbol{l}_{y}$) with $\boldsymbol{l}_{x}\sim p(\boldsymbol{l}_{x})$ and $\boldsymbol{l}_{y}\sim p(\boldsymbol{l}_{y})$ our goal is to translate $\boldsymbol{x}$ to $\hat{\boldsymbol{y}}$ with label $\boldsymbol{l}_{y}$, and to translate $\boldsymbol{y}$ to $\hat{\boldsymbol{x}}$ with label $\boldsymbol{l}_{x}$. The four classification losses are:

\begin{equation}
\begin{aligned}
\mathcal{L}_{c}&=-\mathbb{E}_{\boldsymbol{x}\sim p(\boldsymbol{x}), \boldsymbol{l}_{x}\sim p(\boldsymbol{l}_{x})}[\log P(\boldsymbol{l}_{x}\vert \boldsymbol{x})] \\
&=-\mathbb{E}_{\boldsymbol{y}\sim p(\boldsymbol{y}), \boldsymbol{l}_{y}\sim p(\boldsymbol{l}_{y})}[\log P(\boldsymbol{l}_{y}\vert \boldsymbol{y})] \\
&=-\mathbb{E}_{\boldsymbol{x}\sim p(\boldsymbol{x}), \boldsymbol{l}_{y}\sim p(\boldsymbol{l}_{y})}[\log P(\boldsymbol{l}_{y}\vert G_{Y}(E_{X}(\boldsymbol{x}), \boldsymbol{l}_{y}))] \\
&=-\mathbb{E}_{\boldsymbol{y}\sim p(\boldsymbol{y}), \boldsymbol{l}_{x}\sim p(\boldsymbol{l}_{x})}[\log P(\boldsymbol{l}_{x}\vert G_{X}(E_{Y}(\boldsymbol{y}), \boldsymbol{l}_{x}))]
\end{aligned}
\end{equation}

This loss can be used to optimize discriminators $D_{X}$, $D_{Y}$, generators $G_{X}$, $G_{Y}$, and encoders $E_{X}$, $E_{Y}$.

\textbf{Cycle Consistency Loss}
Although the minimization of GAN losses ensures that $G_Y(E_{X}(\boldsymbol{x}), \boldsymbol{l}_{y})$ produce a sample $\hat{\boldsymbol{y}}$ similar to samples drawn from $Y$, the model still can be unstable and prone to failure. To tackle this problem, we further constrain our model with a cycle-consistency loss \cite{8237506}. To achieve this goal,  we want mapping from domain $X$ to domain $Y$ and then back to domain $X$ to reproduce the original sample, i.e., $G_{X}(E_{Y}(G_{Y}(E_{X}(\boldsymbol{x}), \boldsymbol{l}_{y})), \boldsymbol{l}_{x}) \approx \boldsymbol{x}$ and $G_{Y}(E_{X}(G_{X}(E_{Y}(\boldsymbol{y}), \boldsymbol{l}_{x})), \boldsymbol{l}_{y}) \approx \boldsymbol{y}$. Thus, the cycle-consistency loss is:

\begin{equation}
\begin{aligned}
\mathcal{L}_{cyc}&=\mathbb{E}_{\boldsymbol{x}\sim p(\boldsymbol{x})}[\vert \vert G_{X}(E_{Y}(G_{Y}(E_{X}(\boldsymbol{x}), \boldsymbol{l}_{y})), \boldsymbol{l}_{x}) - \boldsymbol{x} \vert \vert] \\
&+ \mathbb{E}_{\boldsymbol{y}\sim p(\boldsymbol{y})}[\vert \vert G_{Y}(E_{X}(G_{X}(E_{Y}(\boldsymbol{y}), \boldsymbol{l}_{x})), \boldsymbol{l}_{y}) - \boldsymbol{y} \vert \vert]
\end{aligned}
\end{equation}

\textbf{Final Objective of CD-GAN}
To sum up, the goal of our approach is to minimize the following objective:
\begin{equation}
\begin{aligned}
\mathcal{L}(E, G, D) & =\mathcal{L}_{GAN}  + \alpha_{0} \mathcal{L}_{rec} + \alpha_{1} \mathcal{L}_{lcl} + \alpha_{2} \mathcal{L}_{c} + \alpha_{3} \mathcal{L}_{cyc}
\end{aligned}
\end{equation}
where $E$, $G$, and $D$ signify encoders $E_X$, $E_Y$, generators $G_X$, $G_Y$, and discriminators $D_X$, $D_Y$, and $\alpha_{0}$, $\alpha_{1}$, $\alpha_{2}$, $\alpha_{3}$ control the relative importance of the losses. Same as solving a regular GAN problem, training the model involves the solving  of a min-max problem, where $E_{X}$,$E_{Y}$, $G_{X}$, and $G_{Y}$ aim to minimize the objective, while $D_{X}$ and $D_{Y}$ aim to maximize it.

\begin{equation}
E^{\ast}, G^{\ast} = arg \min_{E, G} \max_{D} \mathcal{L}(E, G, D)
\end{equation}


\subsection{Cross-Domain Training}

Our proposed model has two encoder-generator pairs, but we have data from $n$ domains. To train the model using samples of all domains equally, we introduce a cross-domain training algorithm. As shown in Fig.~\ref{m:b}, 
there are 4 domains. At each iteration, we randomly select two domains $R$ and $S$, and feed training data of these two domains into the model. At the next iteration, we might take another two domains $P$ and $Q$, and perform the same training. We train the model using all data samples of $4$ domains at every epoch for several iterations. The training algorithm is shown in Algorithm ~\ref{training}. \textit{Cross-domain training} ensures the model to learn a generic feature representation of all domains by training the model equally on independent domains.

\begin{algorithm}
\caption{Joint domain training on CD-GAN using mini-batch stochastic gradient descent}
\begin{algorithmic}
\STATE \textbf{Require:} Training samples from $n$ domains
\STATE \textbf{Initialize} $\boldsymbol{\theta}_E^X$, $\boldsymbol{\theta}_E^Y$,$\boldsymbol{\theta}_G^X$, $\boldsymbol{\theta}_G^Y$,$\boldsymbol{\theta}_D^X$, and $\boldsymbol{\theta}_D^Y$ with the shared network connection weights set to the same values.
\WHILE{Training loss has not converged}
	
	\STATE Randomly draw two domains $X$ and $Y$ from $n$ domains
	\STATE Randomly draw $N$ samples from the two domains, \{$\boldsymbol{x}_1, \boldsymbol{x}_2, \ldots \boldsymbol{x}_N; \boldsymbol{y}_1, \boldsymbol{y}_2, \ldots \boldsymbol{y}_N$\}
	\STATE Get the domain labels of the samples from the two domains, $\{ \boldsymbol{l}_X^i, \boldsymbol{l}_Y^i \}_{i=1}^{N}$
	
	\STATE \textbf{(1) Update} $\boldsymbol{D}_X, \boldsymbol{D}_Y$ \textbf{with fixed} $\boldsymbol{G}_X, \boldsymbol{G}_Y, \boldsymbol{E}_X, \boldsymbol{E}_Y$	
	
	\STATE Generate fake samples using the real ones
	
	\begin{align*}
	\hat{\boldsymbol{x}}_i = G_X(E_Y(\boldsymbol{y}_i), \boldsymbol{l}_x^i),\hat{\boldsymbol{y}}_i = G_Y(E_X(\boldsymbol{x}_i), \boldsymbol{l}_y^i), i=1 \ldots N
	\end{align*}
	
	\STATE Update $\boldsymbol{\theta}_D=(\boldsymbol{\theta}_D^X, \boldsymbol{\theta}_D^Y)$  according to the following gradients
	\begin{align*}
	& \nabla_{\theta_D} \bigg [ \frac{1}{N} \sum_{i=1}^{N}\Big [-\log D_X(\boldsymbol{x}_i) - \log(1-D_X(\hat{\boldsymbol{x}}_i)) -\log D_Y(\boldsymbol{y}_i) \\
	&- \log(1-D_Y(\hat{\boldsymbol{y}}_i)) + \alpha_2 \big[ \log P(\boldsymbol{l}_x\vert \boldsymbol{x}_i)+  \log P(\boldsymbol{l}_y\vert \boldsymbol{y}_i)\big]\Big ] \bigg ]
	\end{align*}
	
	\STATE \textbf{(2) Update} $\boldsymbol{E}_X, \boldsymbol{E}_Y, \boldsymbol{G}_X, \boldsymbol{G}_Y$ \textbf{with fixed} $\boldsymbol{D}_X, \boldsymbol{D}_Y$
	\STATE Update $\boldsymbol{\theta}_{E,G}=(\boldsymbol{\theta}_{E}^{X}, \boldsymbol{\theta}_{E}^{Y}, \boldsymbol{\theta}_{G}^{X}, \boldsymbol{\theta}_{G}^{Y})$  according to the following gradients
	\begin{align*}
	& \nabla_{\theta_{E,G}} \bigg [\frac{1}{N} \sum_{i=1}^{N}\Big[\log(1-D_X(\hat{\boldsymbol{x}}_i)) + \log(1-D_Y(\hat{\boldsymbol{y}}_i)) \\
	&+ \vert \vert \boldsymbol{x}_i - G_X(E_X(\boldsymbol{x}_i), \boldsymbol{l}_{x}^i) \vert \vert_2 + \vert \vert \boldsymbol{y}_i - G_Y(E_Y(\boldsymbol{y}_i), \boldsymbol{l}_{y}^i)\vert \vert_2 \\
	&+  \vert \vert E_X(\boldsymbol{x}_i)-E_Y(\hat{\boldsymbol{y}}_i)  \vert \vert + \vert \vert E_Y(\boldsymbol{y}_i)-E_X(\hat{\boldsymbol{x}}_i)  \vert \vert \\
	&+  \log P(\boldsymbol{l}_x\vert \hat{\boldsymbol{x}}_i)+  \log P(\boldsymbol{l}_y\vert \hat{\boldsymbol{y}}_i) \\
	&+ \alpha \big[\vert \vert \boldsymbol{x}_i-G_X(E_Y(\hat{\boldsymbol{y}}_i),\boldsymbol{l}_{x}^i)  \vert \vert + \vert \vert \boldsymbol{y}_i-G_Y(E_X(\hat{\boldsymbol{x}}_i),\boldsymbol{l}_{y}^i)  \vert \vert\big] \Big] \bigg ]
	\end{align*}

\ENDWHILE
\end{algorithmic}
\label{training}
\end{algorithm}

\section{experiment}
\label{exp}

In this section, we conduct experiments over three datasets to compare our proposed model with reference models in terms of image translation quality and efficiency.

\subsection{Datasets}

To evaluate the scalability and effectiveness of our model, we test it on a variety of multi-domain image-to-image translation tasks using the following datasets:

\textbf{Alps Seasons dataset} \cite{2017arXiv171206909A} is collected from images on Flickr. The images are categorized into four seasons based on the provided timestamp of when it was taken. It consists of four categories: \textit{Spring}, \textit{Summer}, \textit{Fall}, and \textit{Winter}. The training data consists of 6053 images of four seasons, while the test data consists of 400 images.

\textbf{Painters dataset} \cite{8237506} includes painting images of four artists \textit{Monet}, \textit{Van Gogh}, \textit{Cezanne}, and \textit{Ukiyo-e}. We use 2851 images as the training set, and 200 images as the test set.

\textbf{CelebA dataset} \cite{Liu:2015:DLF:2919332.2920139} contains ten thousand identities, each of which has twenty images, i.e., two hundred thousand images in total. Each image in CelebA is annotated with 40 face attributes. We resize the initial $178 \times 218$ size images to $256 \times 256$. We randomly select 4000 images as test set and use all remaining images for training data.

We run all the experiments on a Ubuntu system using an Intel i7-6850K, along with a single NVIDIA GTX 1080Ti GPU.

\subsection{Reference Models}


We compare the performance of our proposed CD-GAN with that of two reference models:

\textbf{CycleGAN} \cite{8237506} This method trains two generators $G: X\rightarrow Y$ and $F: Y\rightarrow X$ in parallel. It not only applies a standard GAN loss respectively for $X$ and $Y$, but applies forward and backward cycle consistency losses which ensure that an image $\boldsymbol{x} $ from domain $X$ be translated to an image of domain $Y$, which can then be translated back to the domain $X$, and vice versa.

\textbf{DualGAN} \cite{8237572} This method uses a dual-GAN mechanism, which consists of a primal GAN and a dual GAN. The primal GAN learns to translate images from domain $X$ to domain $Y$, while the dual-GAN learns to invert the task. Images from either domain can be translated and then reconstructed. Thus a reconstruction loss can be used to train the model.

\rev{\textbf{UNIT} \cite{NIPS2017_6672} This method consists of two VAE-GANs with a fully shared latent space. To complete the task of image-to-image translation between $n$ domains, it needs to be trained $\frac{n\times{(n-1)}}{2}$ times.}

\rev{\textbf{DB} \cite{2017arXiv171202050H} This method addresses the multi-domain image-to-image translation problem by introducing $n$ domain-specific encoders/decoders to learn an universal shared-latent space.}

\subsection{Evaluation Metrics}

There is a challenge to evaluate the quality of synthesized images~\cite{NIPS2016_6125}. Recent works have tried using pre-trained semantic classifiers to measure the realism and discriminability of the generated images. The idea is that if the generated images look to be more close to real ones,
 classifiers trained on the real images will be able to classify the synthesized images correctly as well. Following \cite{2016arXiv160308511Z, Isola2017ImagetoImageTW, 2016arXiv160305631W}, to evaluate the  performance of the models in classifying generated images quantitatively,  we apply the metric \textit{classification accuracy}. For each experiment, we generate enough number of images of different domains, then we use a pre-trained classifier which is trained on the training dataset to classify them to different domains and calculate the classification accuracy.

\subsection{Network Architecture and Implementation}

\begin{figure}[!t]
\centering
\includegraphics[width=0.9\textwidth]{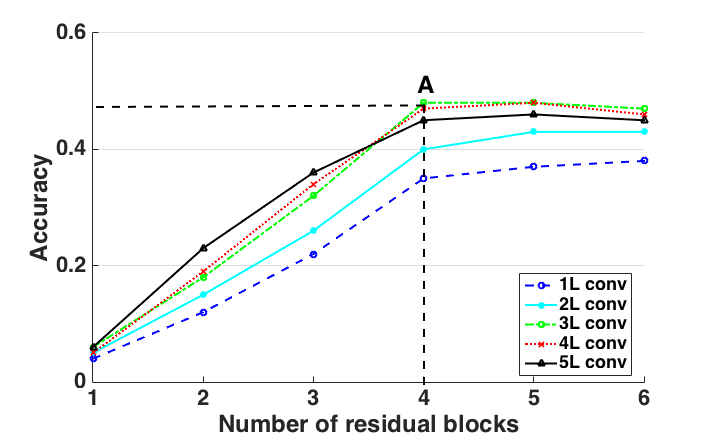}
\caption{The accuracy on varying number of residual blocks and number of convolutional layers.}
\label{res_conv}
\end{figure}

The design of the architecture is always a difficult task \cite{2015arXiv151106434R}. To get a proper model architecture, we adopt the architecture of the discriminator from \cite{Isola2017ImagetoImageTW} which has been proven to be proficient in most image-to-image generation tasks. It has 6 convolutional layers. We keep the discriminator architecture fixed and vary the architectures of the encoders and generators. Following the design of the architectures of the generators in \cite{Isola2017ImagetoImageTW}, we use two types of layers, the regular convolutional layers and the basic residual blocks \cite{DBLP:conf/cvpr/HeZRS16}. Since the encoding process is the inverse of the decoding process, we use the same layers for them but put the layers in the inverse orders. The only difference is the first layer of the encoder and the last layer of the generator. We apply $64$ channels (corresponding to different filters) for the first layer of the encoders, but $3$ channels for the last layer of the generators since the output images have only $3$ RGB channels.
 We gradually change the number of convolutional layers and the number of residual blocks until we get a satisfying architecture. We don't apply \textit{weight sharing} initially. The performance of different architectures is evaluated on the \textit{Painters} dataset and shown in Fig.~\ref{res_conv}. We can see that when the model has 3 regular convolutional layers and 4 basic residual blocks, the model has the best performance. We keep this architecture fixed for other datasets.

We then vary the number of weight-sharing layers in the encoders and the generators. We change the number of weight-sharing layers from 1 to 4. Sharing 1 layer means sharing the highest layer and the lowest layers in the encoder pair. Sharing 2 layers means sharing the highest and lowest two layers. The same sharing method applies for the generator pair (not including the output layer). 
The results are shown in table \ref{tab:share}. We found that sharing 1 layer is enough to have a good performance.

\begin{table}[!t]
\caption{Classification accuracy on number of shared layers in encoders and generators.}
\centering
\begin{tabular}{ccl}
\toprule
\textbf{\# of shared layers} & \textbf{acc. \% (Painters)} & \textbf{acc. \% (Alps Seasons)} \\
\midrule
0 & 49.75 & 29.95  \\
1 & 52.54 & 33.78 \\
2 & 52.81 & 33.54 \\
3 & 51.13 & 33.06 \\
\bottomrule
\end{tabular}
\label{tab:share}
\end{table}

In summary, for the testbed evaluation, we use two encoders each consisting of 3 convolutional layers and 4 basic residual blocks. The generators are composed with 4 basic residual blocks and 3 fractional-strided convolutional layers. The discriminators consist of a stack of 6 convolutional layers. We use LeakyReLU for nonlinearity. The two encoders share the same parameters on their layers 1 and 7, while the two generators share the same parameters on layers 1 and 6,  which is the lowest-level layer before the output layer. The details of the networks are given in table \ref{table:architecture}. We evaluate various network architectures in the evaluation parts. We fix the network architecture as in Table~\ref{table:architecture}.

\begin{table}[!t]
\caption{Network architecture for the multi-modal unsupervised image-to-image translation experiments. $cxkysz$ denote a Convolution-InstanceNorm-ReLU layer with $x$ filters, kernel size $y$, and stride $z$. $Rm$ denotes a residual block that contains two $3 \times 3$ convolutional layers with the same number of filters on both layers. $un$ denotes a $3 \times 3$ fractional-strided-Convolution-InstanceNorm-ReLU layer with $n$ filters, and stride $\frac{1}{2}$. $n_d$ denotes number of domains. $Y$ and $N$ denote whether the layer is shared or not.}
\small
\centering
\begin{tabular}{cccl}
\toprule
\textbf{Layer} & \textbf{Encoders} & \textbf{Generators} & \textbf{Discriminators}  \\
\midrule
1   &  $c64k7s1 (Y)$ & $R256 (Y)$ & $c64k3s2 (N)$
\\
2 & $c128k3s2(N)$ & $R256 (N)$ & $c128k3s2 (N)$
\\
3  & $c256k3s2 (N)$ & $R256 (N)$ & $c256k3s2 (N)$
\\
4 & $R256 (N)$ & $R256 (N)$ &  $c512k3s2 (N)$
\\
5 & $R256 (N)$ & $u256 (N)$ &  $c1024k3s2 (N)$
\\
6 & $R256 (N)$ & $u128 (Y)$& $c(1+n_d)k2s1 (N)$
\\
7 & $R256 (Y)$& $u3 (N)$ &
\\
\bottomrule
\end{tabular}

\label{table:architecture}
\end{table}

We use ADAM \cite{2014arXiv1412.6980K} for training, where the training rate is set to 0.0001 and momentums are set to 0.5 and 0.999. Each mini-batch consists of one image from domain $X$ and one image from domain $Y$. Our model has several hyper-parameters. The default values are $\alpha_{0}=10$, $\alpha_1=0.1$, ${\alpha}_2=0.1$, and ${\alpha}_3=10$. The hyper-parameters of the baselines are set to the suggested values by the authors. 

\subsection{Quantitative Results}
We evaluate our model on different datasets and compare it with baseline models.
\subsubsection{Comparison on Painters Dataset}

\begin{figure}[!t]
\centering
\includegraphics[width=0.9\textwidth]{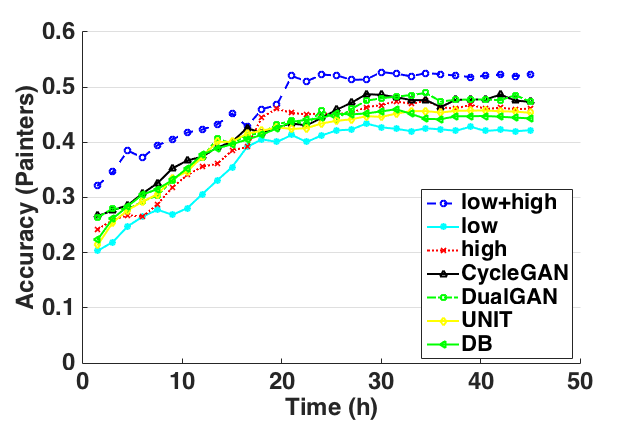}
\caption{The classification accuracy on \textit{Painters} dataset. The 7 models are the proposed model with \textit{the lowest and the highest layer sharing}, \textit{the lowest layer sharing only}, \textit{the highest layer sharing only}, CycleGAN, DualGAN, UNIT, and DB.}
\label{acc_p}
\end{figure}

To compare the proposed model with baseline models \textit{Painters} dataset, we first train the state-of-the-art VGG-11 model \cite{2014arXiv1409.1556S} on training data and get a classifier of accuracy 94.5\%. We then score synthesized images by the classification accuracy against the domain labels these photos were synthesized from. We generate around 4000 images for every 5 hours and the classification accuracies are shown in Fig.~\ref{acc_p}.

We can see that our model achieves the highest classification accuracy of 52.5\% when using both the highest layer and lowest layer sharing, with the training time less than the other reference models in reaching the peak.

\subsubsection{Comparison on Alps Seasons Dataset}
\begin{figure}[!t]
\centering
\includegraphics[width=0.9\textwidth]{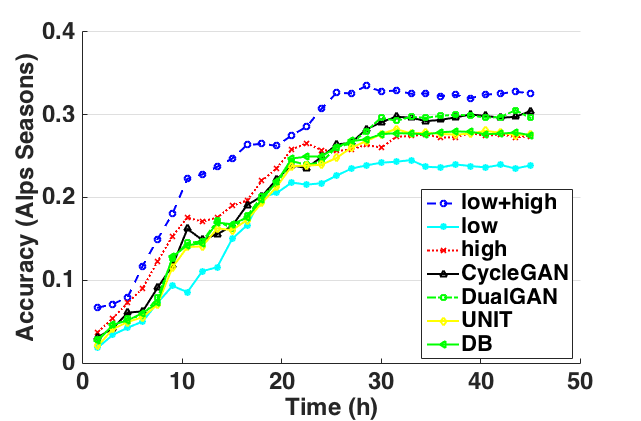}
\caption{The classification accuracy on \textit{Alps Seasons} dataset. The 7 models are the prosed model with \textit{lowest and highest level layers sharing}, \textit{lowest level layers sharing}, \textit{highest level layers sharing}, CycleGAN, DualGAN, UNIT, and DB.}
\label{acc_a}
\end{figure}

We train VGG-11 model on training data of \textit{Alps Seasons} dataset and get a classifier of accuracy 85.5\% trained on the training data. We then classify the generated images by our model and the classification accuracies are shown in Fig.~\ref{acc_a}.

Similar to Fig.~\ref{acc_p}, our model achieves the highest classification accuracy of 33.8\% with the training time less than the baseline models in reaching the peak.




\subsection{Analysis of the loss function}

We compare the ablations of our full loss. As GAN loss and cycle consistency loss are critical for the training of unsupervised image-to-image translation, we keep these two losses as the baseline model and do the ablation experiments to see the importance of other losses.

\begin{table}
  \caption{Ablation study: classification accuracy of \textit{Painters} and \textit{Alps Seasons} datasets for different losses. The following abbreviations are used: R:reconstruction loss, LCL: latent consistency loss, C: classification loss.}
  \label{tab:abl}
  \begin{tabular}{ccl}
    \toprule
    \textbf{Loss}&\textbf{acc.\% (Painters)}&\textbf{acc. \% (Alps Seasons)}\\
    \midrule
    Baseline & 35.23& 20.81\\
    Baseline + R & 36.86& 21.59\\
    Baseline + LCL & 44.42  & 25.05\\
    Baseline + C & 43.63 & 24.01\\
    Baseline + R + LCL & 45.79 & 27.19 \\
    Baseline + R + C & 44.82 & 26.63 \\
    Baseline + LCL + C & 50.74 & 32.51 \\
    Baseline + R + LCL + C & 52.54 & 33.78 \\
  \bottomrule
\end{tabular}
\end{table}

As shown in Tabel~\ref{tab:abl}, the reconstruction loss $R$  is least important with accuracy improvement of about 4.6\% on \textit{Painters} dataset and 3.7\% on \textit{Alps Seasons} dataset. The latent consistency loss $LCL$ brings the model an accuracy improvement of 26.1\% on \textit{Painters} dataset and 20.4\% on \textit{Alps Seasons} dataset. The accuracy is improved by 23.8\% on \textit{Painters} dataset and 15.4\% on \textit{Alps Seasons} dataset by the classification loss $C$.

\subsection{Qualitative Results}
We demonstrate our model on three unsupervised multi-domain image-to-image translation tasks.

\begin{figure}
\centering
\includegraphics[height=5in, width=5in]{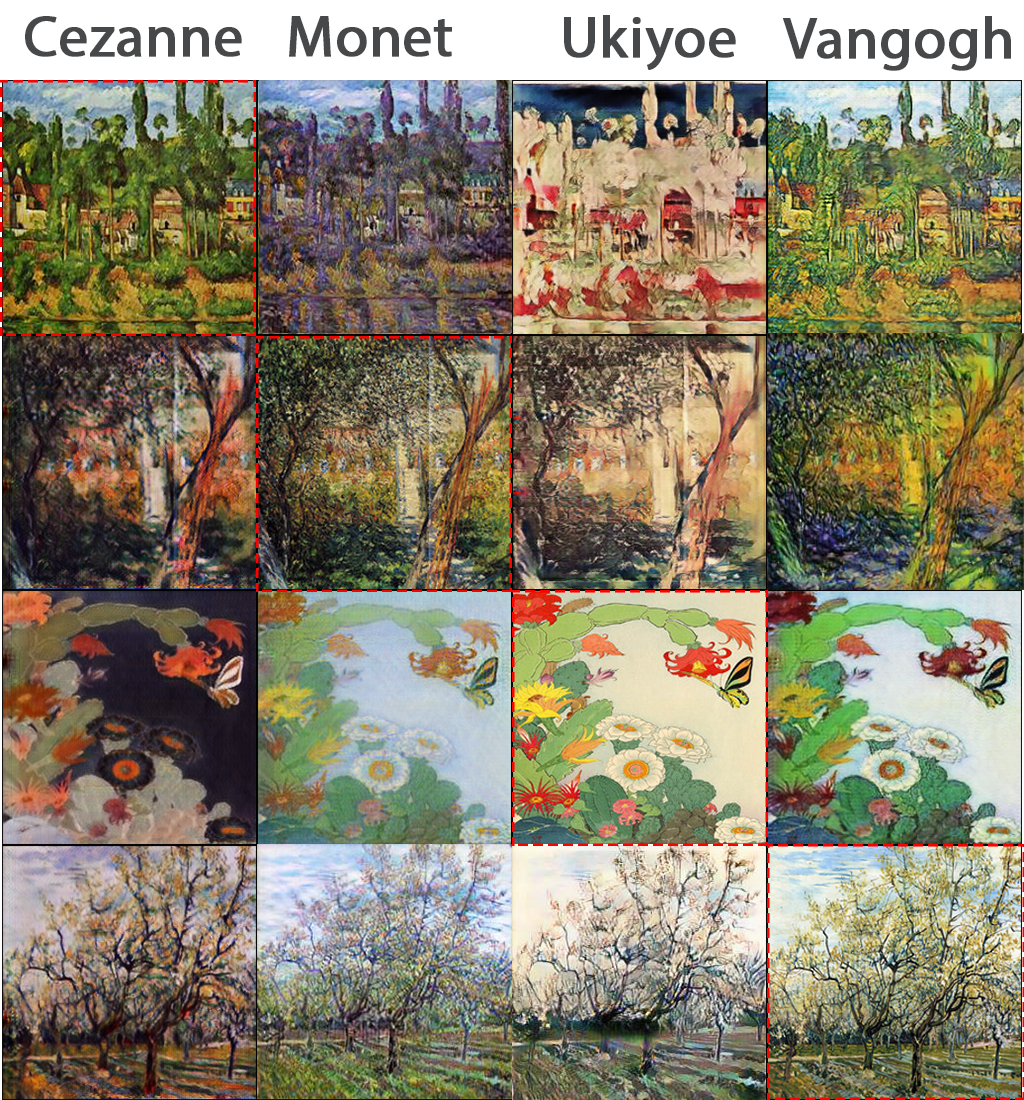}
\caption{\textit{Painters} translation results. The original images are displayed with a dashed square around. The other images are generated according to different painters.}
\label{pr}
\end{figure}

\begin{figure}
\centering
\includegraphics[height=5in, width=5in]{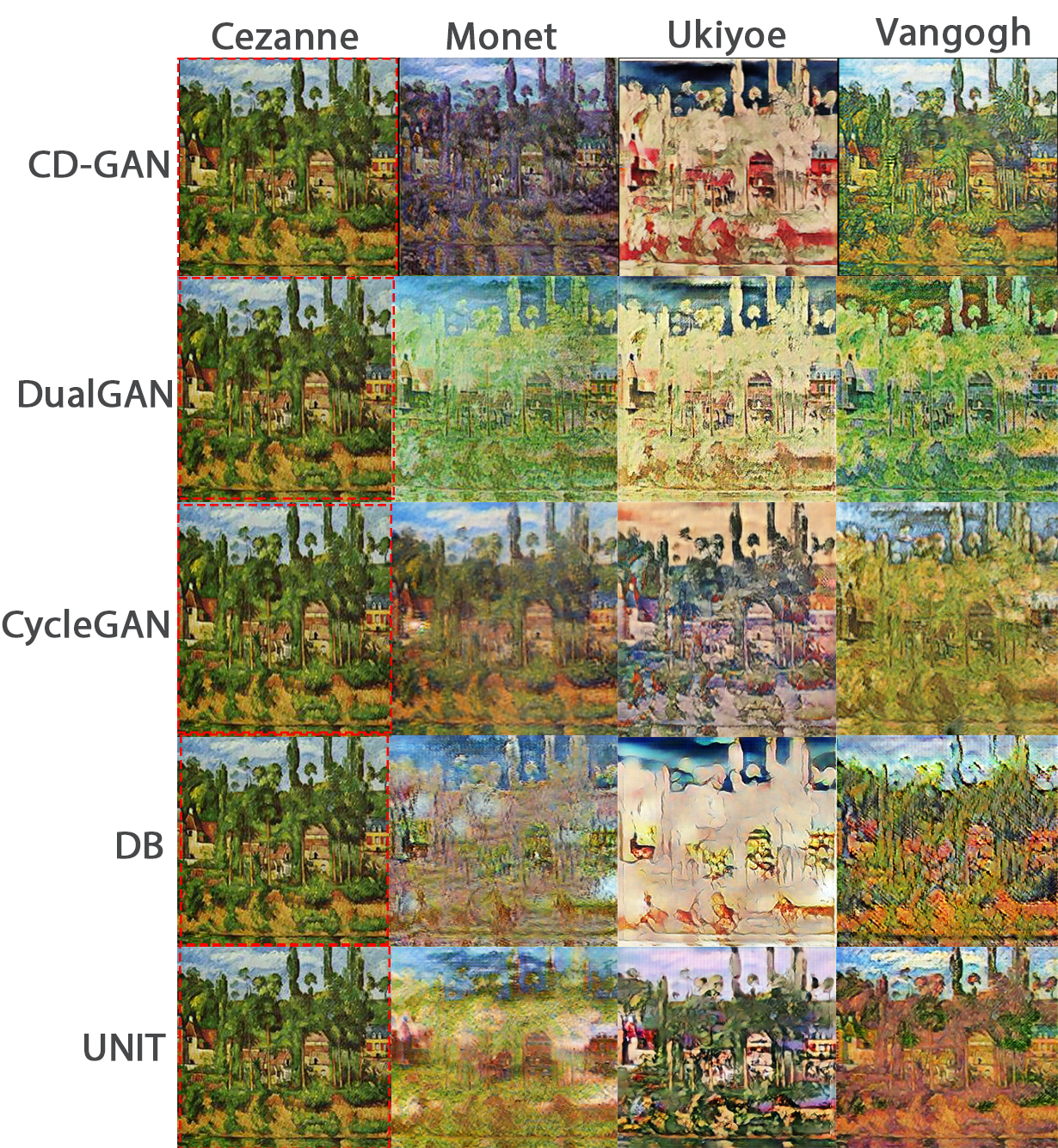}
\caption{\textit{Painters} translation results. The original images are displayed with a dashed square around. The other images are generated according to different painters.}
\label{pr_r}
\end{figure}

\textbf{Painting style transfer (Fig.~\ref{pr})} We train our model on \textit{Painters} dataset and use it to generate images of size $256\times 256$. The model can transfer the painting style of a specific painter to the other painters, e.g., transferring the images of \textit{Cezanne} to images of other three painters \textit{Monet, Ukiyoe} and \textit{Vangogh}. \rev{In Fig.~\ref{pr_r}, we also compare our model with other reference models when given the same test image.}

\begin{figure}
\centering
\includegraphics[height=5in, width=5in]{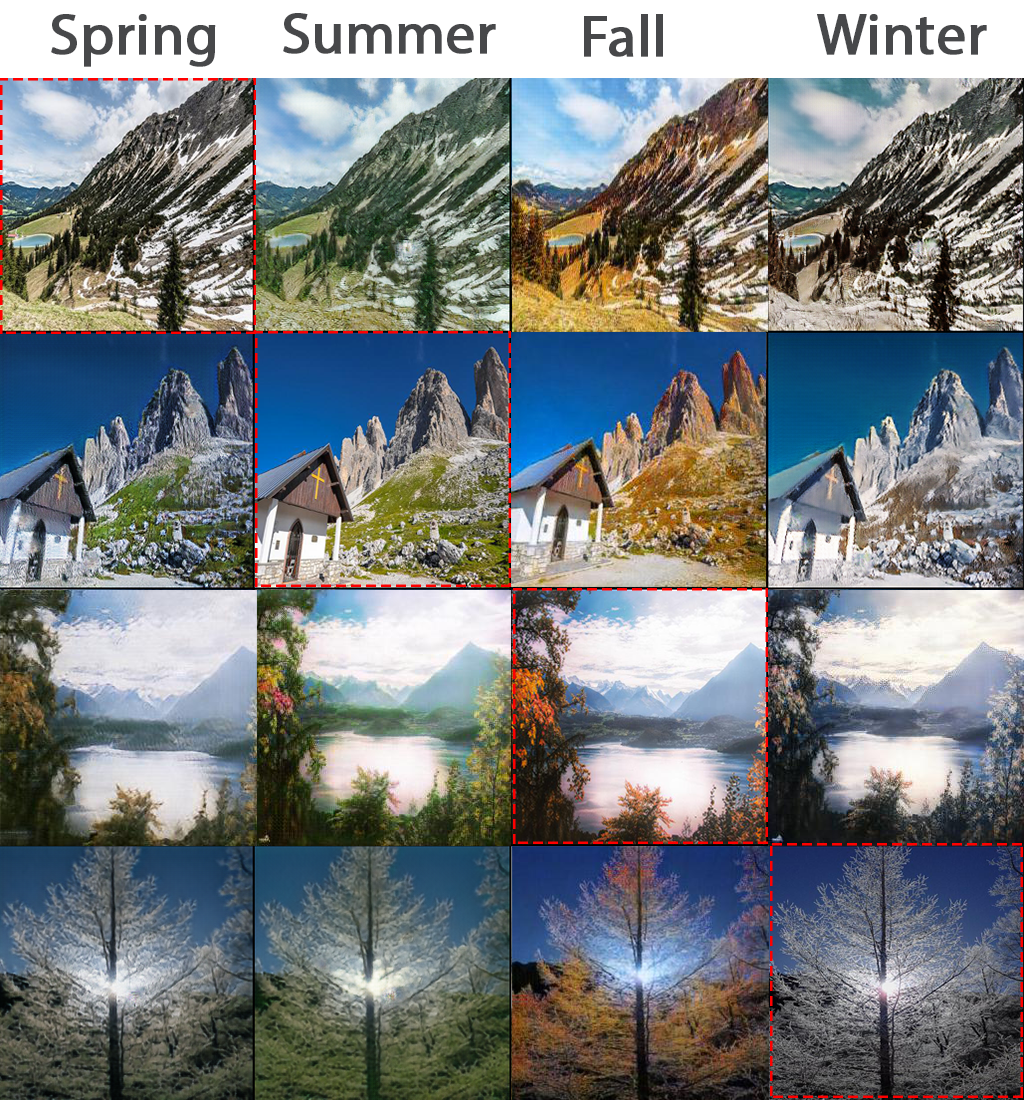}
\caption{\textit{Alps Seasons} translation results. The original images are displayed with a dashed square around. The other images are generated according to different seasons.}
\label{ar}
\end{figure}

\begin{figure}
\centering
\includegraphics[height=5in, width=5in]{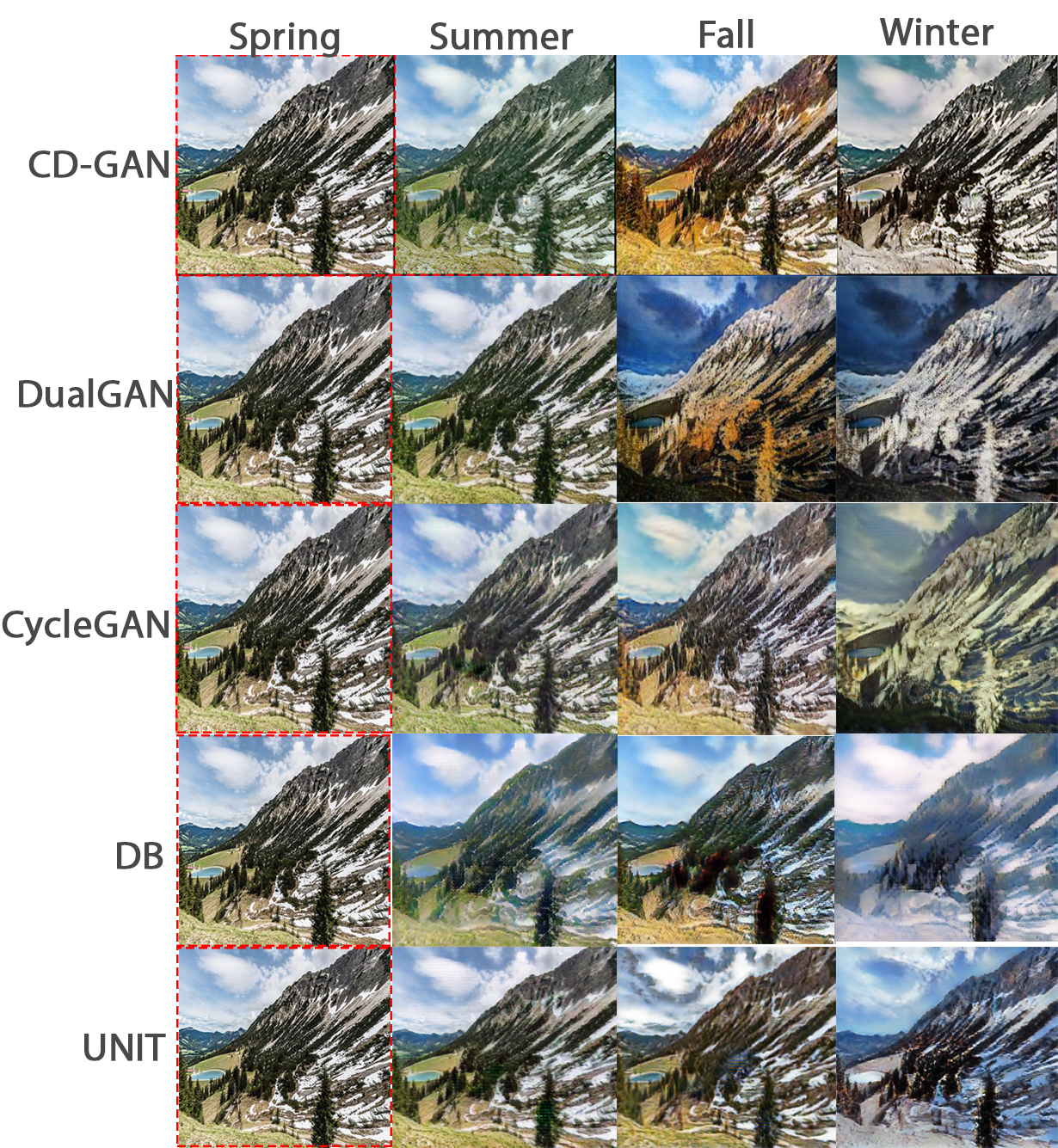}
\caption{\textit{Alps Seasons} translation results. The original images are displayed with a dashed square around. The other images are generated according to different seasons.}
\label{al_r}
\end{figure}

\textbf{Season transfer (Fig.~\ref{ar})}
The model is trained on the \textit{Alps Seasons} dataset. We use the trained model to generate images of different seasons. For example, we generate an image of summer from an image of spring and vice versa. \rev{In Fig.~\ref{al_r}, we also compare our model with other reference models when given the same test image.}

\begin{figure}
\centering
\includegraphics[height=5in, width=5in]{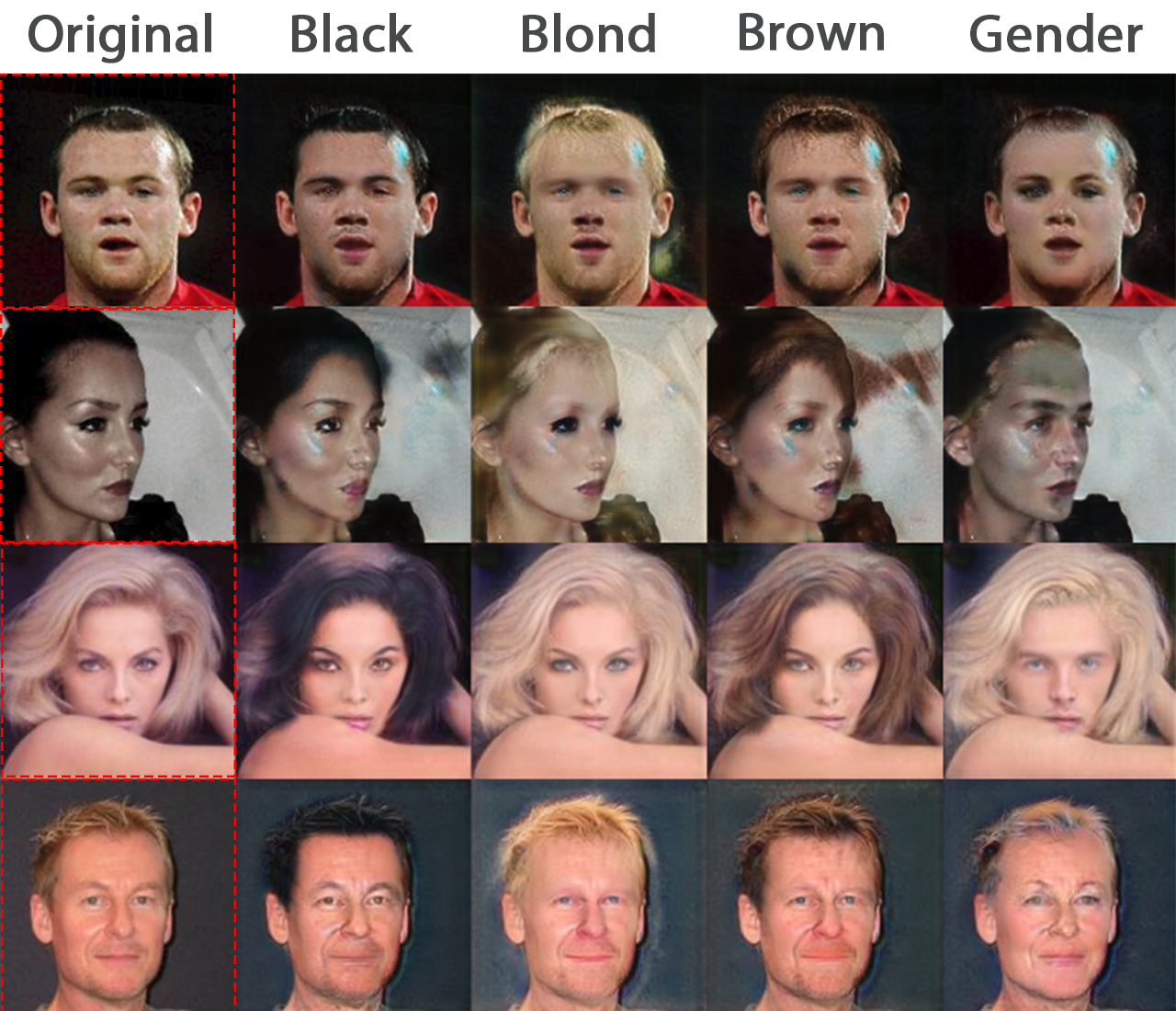}
\caption{Attribute-base face translation results. The original images are displayed with a dashed square around. The other images are generated according to different face attributes.}
\label{fr}
\end{figure}

\textbf{Attribute-base face translation (Fig.~\ref{fr})}
We train the model on \textit{CelebA} dataset for attribute-based face translation tasks. We choose 4 attributes, \textit{black hair}, \textit{blond hair}, \textit{brown hair}, and \textit{gender}. We then use our model to generate images with these attributes. For example, we transfer an image with a man wearing black hair to a man with blond hair, or transfer a man to a woman.

\section{Conclusion}
\label{conclusion}

In this chapter, we propose a Cross-Domain Generative Adversarial Networks (CD-GAN), a novel and scalable model to conduct unsupervised multi-domain image-to-image translation. We show its capability of translating images from one domain to many other domain using several datasets. It still has some limitations. First, training could be unstable due to the training problem of GAN model. Second, the diversity of the generated images are constrained by the cycle consistency loss. We plan to address these two problems in the future work. 
\chapter{Neural Machine Translation}

\section{Introduction}

Syntactic information has been shown to improve the translation quality in NMT models. On the source side, syntax can be incorporated in multiple ways --- either directly during encoding~\cite{Chen18Syntax, SennrichLinguistic16, eriguchi2016tree}, or indirectly via multi-task learning to produce syntax informed representations~\cite{eriguchi2017learning, Baniata18, Niehues17, Zaremoodi18}. On the target side, however, incorporating the syntax is more challenging due to the additional complexity in inference when decoding over latent states. To avoid this, existing methods resort to approximate inference over the latent states using a two-step decoding process~\cite{gu2018top, wang2018tree, wu2017sequence, aharoni2017}. Typically, the first stage decoder produces a beam of latent states, which serve as conditions to feed into the second stage decoder to obtain the target words. Thus, training and inference in these models can only explore a limited sub-space of the latent states.  

In this work, we introduce \namelasyn, a new target side syntax model that allows exhaustive exploration of the latent states to ensure a better translation quality. Similar to prior work, \namelasyn approximates the co-dependence between syntax and semantics of the target sentences by modeling the joint conditional probability of the target words and the syntactic information at each position. However, unlike prior work, \namelasyn eliminates the sequential dependence between the latent variables and simply infers the syntactic information at a given position based on the source text and the partial translation context. This allows \namelasyn to search over a much larger set of latent state sequences. In terms of time complexity, unlike typical latent sequential models, \namelasyn only introduces an additional term that is linear in the size of latent variable vocabulary and the length of the sentence.


We implement \namelasyn by modifying a transformer-based encoder-decoder model. The implementation uses a hybrid decoder that predicts two posterior distributions: the probability of syntactic choices at each position $P(\mathbf{z}_n \vert \mathbf{x}, \mathbf{y}_{<n})$, and the probability of the word choices at each position conditioned on each of the possible values for the latent states $P(\mathbf{y}_n \vert \mathbf{z}_n, \mathbf{x}, \mathbf{y}_{<n})$. 
The model cannot be trained by directly optimizing the data log-likelihood because of its non-convex property. We devise a neural expectation maximization (NEM) algorithm, 
whose E-step computes the posterior distribution of latent states under current model parameters, and M-step updates model parameters using gradients from back-propagation. Given some supervision signal for the latent variables, we can modify this EM algorithm to obtain a regularized training procedure. We use parts-of-speech (POS) tag sequences, automatically generated by an existing tagger, as the source of supervision. 

Because the decoder is exposed to more latent states during training, it is more likely to generate diverse translation candidates. To obtain diverse sequences, we can decode the most likely translations for different POS tag sequences. This is a more explicit and effective way of performing diverse translation than other methods based on diverse or re-ranking beam search~\cite{Vijayakumar18, LiJ16mutual}, or coarse codes planning~\cite{Shu18}. 

We evaluate \namelasyn on four translation tasks. Evaluations show that \namelasyn outperforms models that only use partial exploration of the latent states for incorporating target side syntax. 
A diversity based evaluation also shows that when using different POS tag sequences during inference, \namelasyn produces more diverse and meaningful translations compared to existing models. 
\section{Related Work}
Attention-based Neural Machine Translation (NMT) models have shown promising results in various large scale translation tasks~\cite{BahdanauCB14, luong2015, SennrichHB16, Vaswani17} using an \texttt{Seq2Seq} structure. Many Statistical Machine Translation (SMT) approaches have leveraged benefits from modeling syntactic information~\cite{Chiang2009, Huang2006, Shen2008}. Recent efforts have demonstrated that incorporating syntax can also be useful in neural methods as well.

One branch uses features on the source side to help improve the translation performance~\cite{SennrichLinguistic16, Morishita18, eriguchi2016tree}. Sennrich \textit{et al.}~\cite{SennrichLinguistic16} explored linguistic features like lemmas, morphological features, POS tags and dependency labels and concatenate their embeddings with sentence features to improve the translation quality. In a similar vein, Morishita \textit{et al.}~\cite{Morishita18} and Eriguchi \textit{et al.}~\cite{eriguchi2016tree}, incorporated hierarchical subword features and phrase structure into the source side representations. Despite the promising improvements, these approaches are limited in that the trained translation model requires the availability of external tools during inference -- the source text needs to be processed first to extract syntactic structure~\cite{eriguchi2017learning}.

Another branch uses multitask learning, where the encoder of the NMT model is trained to produce multiple tasks such as POS tagging, named-entity recognition, syntactic parsing or semantic parsing~\cite{eriguchi2017learning, Baniata18, Niehues17, Zaremoodi18}. These can be seen as models that implicitly generate syntax informed representations during encoding. With careful model selection, these methods have demonstrate some benefits in NMT.

The third branch directly models the syntax of the target sentence during decoding~\cite{gu2018top, wang2018tree, wu2017sequence, aharoni2017, bastings2017graph, li2018target}. 
Aharoni \textit{et al.}~\cite{aharoni2017} treated constituency trees as sequential strings and trained a \texttt{Seq2Seq} model to translate source sentences into these tree sequences. Wang \textit{et al.}~\cite{wang2018tree} and Wu \textit{et al.}~\cite{wu2017sequence} proposed to use two RNNs, a Rule RNN and a Word RNN, to generate a target sentence and its corresponding tree structure. Gu \textit{et al.}~\cite{gu2018top} proposed a model to translate and parse at the same time. 
However, apart from the complex tree structure to model, they all have a similar architecture as shown in Figure~\ref{fig:Bayes2}, which limits them to only exploring a small portion of the syntactic space during inference.

\namelasyn uses simpler parts-of-speech information in a latent syntax model, avoiding the typical exponential search complexity in the latent space with a linear search complexity and is optimized by EM. This allows for better translation quality as well as diversity. Similar to our work, \cite{Shankar2018Surprisingly} and \cite{shankar2018posterior} proposed a latent attention mechanism to further reduce the complexity of model implementation by taking a top-K approximation instead of the EM algorithm as in \namelasyn.

\section{A Latent Syntax Model for Decoding}

In a standard sequence-to-sequence model, the decoder directly predicts the target sequence $\mathbf{y}$ conditioned on the source input $\mathbf{x}$. The translation probability $P(\mathbf{y}\vert \mathbf{x})$ is modeled directly using the probability of each target word $\mathbf{y}_n$ at time step $n$ conditioned on the source sequence $\mathbf{x}$ and the current partial target sequence $\mathbf{y}_{<n}$ as follows:
\begin{equation}
P(\mathbf{y} \vert \mathbf{x}; \boldsymbol{\theta}) = \prod_{n=1}^{N} P(\mathbf{y}_n\vert \mathbf{x}, \mathbf{y}_{<n};\boldsymbol{\theta})
\end{equation}
where, $\boldsymbol{\theta}$ denotes the parameters of both the encoder and the decoder.
 
In this work, we model syntactic information of target tokens using an additional sequence of variables, which captures the syntactic choices\footnote{The variables can be used to model any linguistic information that can be expressed as choices for each word position (e.g., morphological choices).} at each time step. There are multiple ways of incorporating this additional information in a sequence-to-sequence model.

\begin{figure}
\centering
\begin{subfigure}[b]{0.45\textwidth}
   \includegraphics[width=1\textwidth]{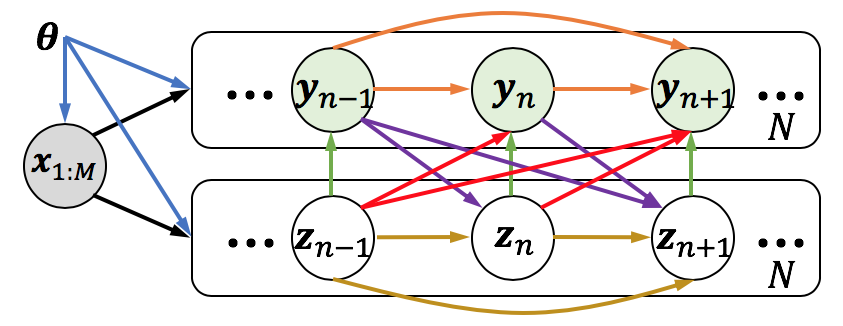}
   \caption{Full co-dependence model.}
   \label{fig:Bayes1} 
\end{subfigure}
\begin{subfigure}[b]{0.45\textwidth}
   \includegraphics[width=1\textwidth]{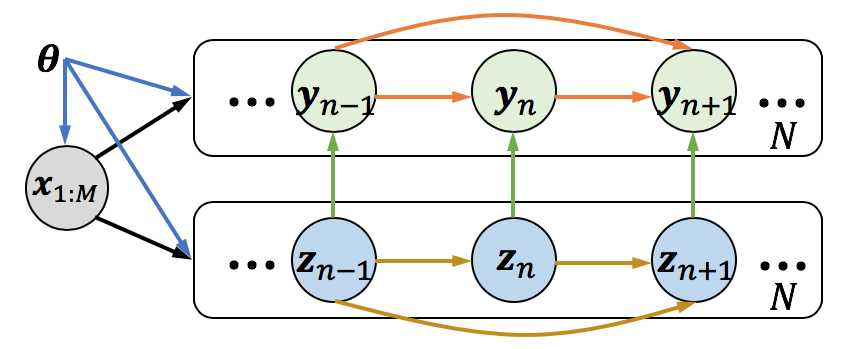}
   \caption{Two-step decoding model.}
   \label{fig:Bayes2}
\end{subfigure}
\begin{subfigure}[b]{0.45\textwidth}
   \includegraphics[width=1\textwidth]{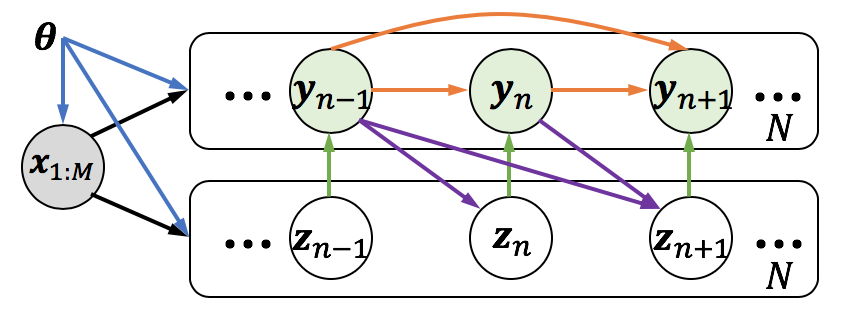}
   \caption{\namelasyn: Our Latent syntax model}
   \label{fig:Bayes3}
\end{subfigure}
\caption[]{Target-side Syntax Models: (a) An ideal solution that captures full co-dependence between syntax and semantics. (b) A widely-used two-step decoding model~\cite{wang2018tree, wu2017sequence, aharoni2017}. 
(c) \namelasyn, our latent syntax model that uses non-sequential latent variables for exhaustive search of latent states.}
\end{figure}
An ideal solution should capture the co-dependence between syntax and semantics. In a sequential translation process, the word choices at each time step depend on both the semantics and the syntax of the words generated at the previous time steps. The same dependence also holds for the syntactic choices at each time step.
Figure~\ref{fig:Bayes1} shows a graphical model that captures this {\em full} co-dependence between the syntactic variable sequence $\mathbf{z}_1, \dots, \mathbf{z}_N$ and the output word sequence $\mathbf{y}_1, \dots, \mathbf{y}_N$. 
Such a model can be implemented using two decoders, one to decode syntax and the other to decode output words. The main difficulty, however, is that inference in this model is intractable since it involves marginalizing over the latent $\mathbf{z}$ sequences.

To keep inference tractable, existing approaches treat syntactic choices $\mathbf{z}$ as observed sequential variables~\cite{gu2018top, wang2018tree, wu2017sequence, aharoni2017}, as shown in Figure~\ref{fig:Bayes2}. 
These models use a two-stage decoding process, where for each time step they first produce most likely latent state $\mathbf{z}_n$ and then use this as input to a second stage that decodes words.
However, this process is unsatisfactory in two respects. First, the inference of syntactic choices is still approximate as it does not explore the full space of $\mathbf{z}$.
Second, these models are not well-suited for controllable or diverse translations. Using such a model to decode from an arbitrary $\mathbf{z}$ sequence is a divergence from its training, where it only learns to decode from a limited space of $\mathbf{z}$ sequences.

\subsection{Model Description}
Our goal is to design a model that allows for exhaustive search over syntactic choices.
We introduce \namelasyn, a new latent model shown in Figure~\ref{fig:Bayes3}. The syntactic choices are modeled as {\em true latent variables} i.e., unobserved variables. Compared to the ideal model in Figure~\ref{fig:Bayes1}, \namelasyn includes two simplifications for tractability: (i) The dependence between successive syntactic choices is modeled indirectly, via the word choices made in the previous time steps. (ii) The word choice at each position depends only on the syntactic choice at the current position and the previous predicted words. Dependence on previous syntactic choices is modeled indirectly.

Under this model, the joint conditional probability of the target word $\mathbf{y}_n$ together with its corresponding latent syntactic choice $\mathbf{z}_n$\footnote{Note that $\mathbf{z}_n \in V_{\mathbf{z}}$, where $V_{\mathbf{z}}$ is the vocabulary of latent syntax for the target, which differs from language to language. 
} is given by:

\begin{equation}
\begin{aligned}
P(\mathbf{y}_n, \mathbf{z}_n \vert \mathbf{x}, \mathbf{y}_{<n}) = P(\mathbf{y}_{n} \vert \mathbf{z}_n, \mathbf{x}, \mathbf{y}_{<n})
 \times P(\mathbf{z}_n\vert \mathbf{x}, \mathbf{y}_{<n})
\end{aligned}
\label{eq:joint}
\end{equation}

\begin{figure}
\centering
\includegraphics[width=0.9\textwidth]{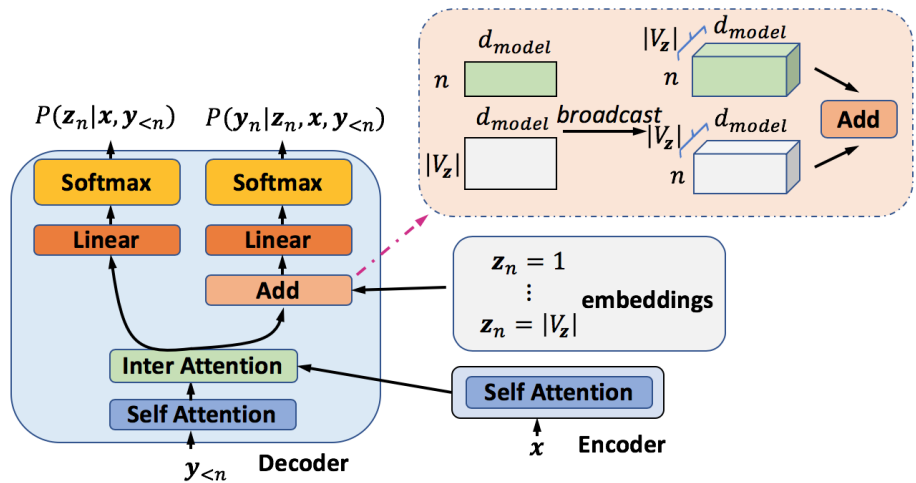}
\caption{The architecture of \namelasyn.}
\label{fig:architecture}
\end{figure}

We implement \namelasyn by modifying the Transformer-based encoder-decoder architecture~\cite{Vaswani17}. As shown in Figure~\ref{fig:architecture}, \namelasyn consists of a shared encoder for encoding source sentence $\mathbf{x}$ and a hybrid decoder that manages the decoding of the latent sequence $\mathbf{z}$ (left branch) and the target sentence $\mathbf{y}$ (right branch) separately. 

The encoder consists of the standard self-attention layer, which generates representations of each token in the source sentence $\mathbf{x}$.
The hybrid decoder consists of a self-attention layer that encodes the output generated thus far (i.e., the partial translation), followed by a inter-attention layer which computes the attention across the encoder and decoder representations. 

The decoder's left branch predicts the latent variable distribution $P(\mathbf{z}_n\vert \mathbf{x}, \mathbf{y}_{<n})$ by applying a simple linear transformation and softmax on the inter-attention output, which contains information about the encoded input $\mathbf{x}$ and the partial translation $\mathbf{y}_{<n}$. 

The right branch predicts the target word distribution $P(\mathbf{y}_{n} \vert \mathbf{z}_n, \mathbf{x}, \mathbf{y}_{<n})$ using the inter-attention output and the embeddings of all the available choices for $\mathbf{z}_n$. The choices for $\mathbf{z}_n$ are represented as embeddings that can be learned during training. We then combine the inter-attention output and the latent choice embeddings through an \texttt{Add} operation, which is a simple composition function that captures all combinations of additive interactions between the two. The dimension of the inter-attention is $n\times d_{model}$ and that of the latent embeddings is $\vert V_{\mathbf{z}} \vert \times d_{model}$, where $\vert V_{\mathbf{z}} \vert$ is the total number of choices for $\mathbf{z}_n$ or the size of the latent variable vocabulary. We broadcast them to the same dimension $n \times \vert V_{\mathbf{z}} \vert \times d_{model}$ and then simply add them together point-wise as shown in Figure~\ref{fig:architecture}. 
This is then fed to a linear transform and softmax over the target word vocabulary.

\subsection{Inference with Exhaustive Search for Latent States}

When using additional variables to model target side syntax, exact inference requires marginalizing over these additional variables.
\begin{equation}
   P(\mathbf{y} \vert \mathbf{x}) = \sum_{\mathbf{z}\in F(\mathbf{z})} P(\mathbf{y} \vert \mathbf{z},\mathbf{x})P(\mathbf{z} \vert \mathbf{x}) 
\end{equation}

To avoid this exponential complexity, prior works use a two-step decoding process with models similar to the one shown in Figure~\ref{fig:Bayes2}. They use greedy or beam search to explore a subset $B(\mathbf{z})$ of the latent space to compute the posterior distribution as follows:
\begin{equation}
   P(\mathbf{y} \vert \mathbf{x}) \simeq \sum_{\mathbf{z}\in B(\mathbf{z})} P(\mathbf{y} \vert \mathbf{z},\mathbf{x})P(\mathbf{z} \vert \mathbf{x}) 
\end{equation}

Finding the most likely translation using \namelasyn also requires marginalizing over the latent states. However, because the latent states in \namelasyn don't directly depend on each other, we can exhaustively search over the latent states. 
In particular, we can show that when $\mathbf{y}$ is fixed (observed),  the $\{\mathbf{z}_n\}_{n=1}^{N}$ variables are d-separated~\cite{bishop2006} i.e., are mutually independent. As a result, the time complexity for searching latent sequence $\mathbf{z}$ drops from ${|V_{\mathbf{z}}|}^N$ to $N|V_{\mathbf{z}}|$.

The posterior distribution for the translation probability at a time step $n$ can be computed as follows:
\begin{equation}
\begin{aligned}
    P&(\mathbf{y}_n \vert \mathbf{x}, \mathbf{y}_{<n}) 
    &= \sum_{\mathbf{z}_n\in F(\mathbf{z}_n)} P(\mathbf{y}_n, \mathbf{z}_n \vert \mathbf{x}, \mathbf{y}_{<n})\\
    &= \sum_{\mathbf{z}_n\in F(\mathbf{z}_n)} P(\mathbf{y}_n \vert \mathbf{z}_n,\mathbf{x}, \mathbf{y}_{<n}) \times P(\mathbf{z}_n \vert \mathbf{x}, \mathbf{y}_{<n})
\end{aligned}
\end{equation}
where, $F(\mathbf{z}_n)$ is the full search space of latent states $\mathbf{z}_n$ and the joint probability is factorized as specified in Equation~\ref{eq:joint}.

For decoding words, we use standard beam search\footnote{Note our primary goal is to perform exhaustive search in the latent space. Search in the target vocabulary space remains exponential in our model.} with $P(\mathbf{y}_n, \mathbf{z}_{n} \vert \mathbf{x}, \mathbf{y}_{<n})$ as the beam cost. With this inference scheme, we can easily control decoding for diversity, by feeding different $\mathbf{z}$ sequences to the right branch of the decoder and decode diverse $\mathbf{y}_n$ by directly using $P(\mathbf{y}_n \vert \mathbf{z}_{n}, \mathbf{x}, \mathbf{y}_{<n})$ as the beam cost. Unlike the two-step decoding models which only evaluate a small number of $\mathbf{z}_n$ values at each time step (constrained by beam size), \namelasyn evaluates all possible values for $\mathbf{z}_n$ at each time step, while avoiding the evaluation of all possible sequences.

\subsection{Training with Neural Expectation Maximization}
The log-likelihood of \namelasyn's parameters $\boldsymbol{\theta}$\footnote{This includes the trainable parameters of the encoder, decoder, and the latent state embeddings.} computed on one training pair $\langle \mathbf{x}, \mathbf{y} \rangle \in D$ is given by:
\begin{equation}
\label{eq:loglikelihood}
\begin{aligned}
\mathcal{L}(\mathbf{\boldsymbol{\theta}}) &= \log P(\mathbf{y} \vert \mathbf{x}; \mathbf{\boldsymbol{\theta}}) \\
&= \sum_{n=1}^{N}\log P(\mathbf{y}_n \vert \mathbf{x}, \mathbf{y}_{<n}; \mathbf{\boldsymbol{\theta}}) \\
&= \sum_{n=1}^{N}\log \sum_{\mathbf{z}_n \in V_{\mathbf{z}}} P(\mathbf{y}_n, \mathbf{z}_n\vert \mathbf{x}, \mathbf{y}_{<n})
\end{aligned}
\end{equation}

Directly optimizing the log-likelihood function (equation~\ref{eq:loglikelihood}) with respect to model parameter $\boldsymbol{\theta}$ is challenging because of the highly non-convex function $P(\mathbf{y}_n, \mathbf{z}_n\vert \mathbf{x}, \mathbf{y}_{<n})$ and the marginalization over $\mathbf{z}_n$.\footnote{Note that marginalization is an issue during training, unlike in inference. As $P(y_n, z_n)$ is already an non-convex function with respect to $\theta$, summing $P(y_n, z_n)$ over different values of $z_n$ makes the function more complicated. Besides, we also need to compute gradients to update the parameters and computing the gradient of a log-of-sum function is costly and unstable. During the translation, we only need to compute the value of $\mathcal{L}(\theta)$ as score for beam searching. Therefore, the marginalization is not an issue.} Alternatively, we optimize the system parameters by Expectation Maximization (EM).

Using Jensen's inequality, equation~\eqref{eq:loglikelihood} can be re-written as:
\begin{equation}
\label{eq:lower}
\begin{aligned}
\mathcal{L}(\mathbf{\boldsymbol{\theta}}) &= \sum_{n=1}^N\log \sum_{\mathbf{z}_n \in V_{\mathbf{z}}} Q(\mathbf{z}_n) \frac{P(\mathbf{y}_n, \mathbf{z}_n\vert \mathbf{x}, \mathbf{y}_{<n})}{Q(\mathbf{z}_n)} \\
\geq & \sum_{n=1}^N \sum_{\mathbf{z}_n \in V_{\mathbf{z}}} Q(\mathbf{z}_n) \log \frac{P(\mathbf{y}_n, \mathbf{z}_n\vert \mathbf{x}, \mathbf{y}_{<n})}{Q(\mathbf{z}_n)} \\
=& \mathcal{L}_{lower}(Q,\boldsymbol{\theta})
\end{aligned}
\end{equation}
where $\mathcal{L}_{lower}(Q,\boldsymbol{\theta})$ is the lower bound of the log-likelihood and $Q$ is any auxiliary probability distribution defined on $\mathbf{z}_n$. $\boldsymbol{\theta}$ is omitted from the expression for simplicity.

We set $Q(\mathbf{z}_n)=P(\mathbf{z}_n \vert \mathbf{x}, \mathbf{y}_{\leq n}; \boldsymbol{\theta}^{old})$, the probability of the latent state computed by the decoder (shown in the left branch in Figure~\ref{fig:architecture}). Substituting this in equation~\eqref{eq:lower}, we obtain the lower bound as

\begin{equation}
\begin{aligned}
&\mathcal{L}_{lower}(Q,\boldsymbol{\theta})=\\ &\sum_{n=1}^N \sum_{\mathbf{z}_n \in V_{\mathbf{z}}} P(\mathbf{z}_n \vert \mathbf{x}, \mathbf{y}_{\leq n}; \boldsymbol{\theta}^{old})
\times \log P(\mathbf{y}_n, \mathbf{z}_n\vert \mathbf{x}, \mathbf{y}_{<n}; \boldsymbol{\theta}) \\
&- P(\mathbf{z}_n \vert \mathbf{x}, \mathbf{y}_{\leq n}; \boldsymbol{\theta}^{old}) \times \log P(\mathbf{z}_n \vert \mathbf{x}, \mathbf{y}_{\leq n}; \boldsymbol{\theta}^{old}) \\
&= \mathcal{Q}(\boldsymbol{\theta} , \boldsymbol{\theta}^{old}) + C
\end{aligned}
\end{equation}
where 

\begin{equation}
\begin{aligned}
\mathcal{Q}(\boldsymbol{\theta}, \boldsymbol{\theta}^{old}) =& \sum_{n=1}^N \sum_{\mathbf{z}_n \in V_{\mathbf{z}}} P(\mathbf{z}_n \vert \mathbf{x}, \mathbf{y}_{\leq n}; \boldsymbol{\theta}^{old}) \\
& \times \log P(\mathbf{y}_n, \mathbf{z}_n\vert \mathbf{x}, \mathbf{y}_{<n}; \boldsymbol{\theta})
\end{aligned}
\label{eq:expectation}
\end{equation}

EM algorithm for optimizing $\mathcal{Q}(\boldsymbol{\theta}, \boldsymbol{\theta}^{old})$ consists of two major steps.
In the E-step, we compute the posterior distribution of $\mathbf{z}_n$ with respect to $\boldsymbol{\theta}^{old}$ by

\begin{equation}
\begin{aligned}
\gamma&(\mathbf{z}_n = i) = P(\mathbf{z}_n = i \vert \mathbf{x}, \mathbf{y}_{\leq n}) \\
&= \frac{P(\mathbf{y}_n, \mathbf{z}_n = i\vert \mathbf{x}, \mathbf{y}_{<n})}{\sum_{\mathbf{z}_n = j}P(\mathbf{y}_{n}, \mathbf{z}_n = j\vert \mathbf{x}, \mathbf{y}_{<n})}
\end{aligned}
\end{equation}

where $\gamma(\mathbf{z}_n = i)$ is the responsibility of $\mathbf{z}_n = i$ given $\mathbf{y}_{n}$, and can be calculated by equation~\eqref{eq:joint}.

In the M-step, we aim to find the configuration of $\boldsymbol{\theta}$ that would maximize the expected log-likelihood using the posteriors computed in the E-step. In conventional EM algorithm for shallow probabilistic graphical model, the M-step is generally supposed to have closed-form solution. However, we model the probabilistic dependencies by deep neural networks, where $\mathcal{Q}(\boldsymbol{\theta}, \boldsymbol{\theta}^{old})$ is highly non-convex and non-linear with respect to network parameters $\boldsymbol{\theta}$. Therefore, there exists no analytical solution to maximize it. However, since deep neural network is differentiable, we can update $\boldsymbol{\theta}$ by taking a gradient ascent step:

\begin{equation}
\begin{aligned}
	\boldsymbol{\theta}^{new}=\boldsymbol{\theta}^{old} + \eta \frac{\partial \mathcal{Q}( \boldsymbol{\theta},\boldsymbol{\theta}^{old})}{\partial \boldsymbol{\theta}},
\end{aligned}
\label{eq:theta}
\end{equation}

The resulting algorithm belongs to the class of \textit{generalized EM algorithms} and is guaranteed (for a sufficiently small learning rate $\eta$) to converge to a (local) optimum of the data log likelihood \cite{wu83}.

\subsection{Regularized EM training}
The EM training we derived does not assume any supervision for the latent variables $\mathbf{z}$. This can be seen as inferring the latent syntax of the target sentences by clustering the target side tokens into $|V_{\mathbf{z}}|$ different categories. Given some token-level syntactic information, we can modify the training procedure to regularize the generation of latent sequence $P(\mathbf{z}_n \vert \mathbf{x}, \mathbf{y}_{< n})$ such that true latent sequences have higher probabilities under the model. In this work, we consider parts-of-speech sequences of the target sentences for regularization. 

The regularized EM training objective is thus redefined as
\begin{align}
	\mathcal{L}_{total}(\mathbf{\boldsymbol{\theta}}) = \mathcal{L}_{lower}(\boldsymbol{\theta}) + \lambda \mathcal{L}_{\mathbf{z}}(\boldsymbol{\theta}),
\end{align} 
where $\mathcal{L}_{lower}(\boldsymbol{\theta})$ is the EM lower bound in equation~\eqref{eq:lower} and $\mathcal{L}_{\mathbf{z}}(\boldsymbol{\theta})$ denotes cross entropy loss between $P(\mathbf{z}_n \vert \mathbf{x}, \mathbf{y}_{< n})$ and the true POS tag sequences and $\lambda$ is a hyper-parameter that controls the impact of the regularization.

This regularized training algorithm is shown in Algorithm~\ref{algo:training}.

\begin{algorithm}
\small
\caption{Training NMT with latent POS tag sequences through Regularized Neural EM}
\begin{algorithmic}
\STATE \textbf{Objective:} Maximize the log likelihood function $\mathcal{Q}( \boldsymbol{\theta},\boldsymbol{\theta}^{old})$ with respect to $\mathbf{\boldsymbol{\theta}}$ over observed variable $\mathbf{y}$ and latent variable $\mathbf{z}$, governed by parameters $\boldsymbol{\theta}$.
\STATE \textbf{Input:} Parallel corpus training data $\langle \mathbf{x}, \mathbf{y} \rangle$; POS tag sequence $\mathbf{z}$ of target sentence; the number of EM update steps per batch $K$.
\STATE \textbf{Initialize:} Initialize random values for the parameters $\boldsymbol{\theta}^{old}$.
\WHILE{Training loss has not converged}
	\STATE Select $\langle \mathbf{x}, \mathbf{y} \rangle \in D$, parse $\mathbf{z}$ of $\mathbf{y}$.
	\FOR{$k \gets 1$ to $K$}
		\STATE \textbf{1. E-step:} Evaluate $\gamma(\mathbf{z}_n=i)$ for $i\leq |V_{\mathbf{z}}|$.
		\STATE \textbf{2. M-step:} Evaluate $\boldsymbol{\theta}^{new}$ given by equation~(\ref{eq:theta}).
		\STATE \textbf{3.} Let $\boldsymbol{\theta}^{old} \gets \boldsymbol{\theta}^{new}$.
	\ENDFOR
\ENDWHILE
\end{algorithmic}
\label{algo:training}
\end{algorithm}

\section{Evaluation}
We evaluate \namelasyn on four translation tasks, including three with moderate sized datasets IWSLT 2014~\cite{Cettolo14} German-English (De-En), English-German (En-De), English-French (En-Fr), and one with a relatively larger dataset, the WMT 2014 English-German (En-De). We describe the datasets in more details in the appendix.

We compare against three types of baselines: (i) general Seq2Seq models that use no syntactic information, (ii) models that incorporate source side syntax directly, and multitask learning models which include syntax indirectly, and (iii) models that use syntax on the target side. We also define a \namelasyn \textbf{Empirical Upper Bound (EUB)}, which is our proposed model using true POS tag sequences for inference.

We use BLEU as the evaluation metric~\cite{papineni02} for translation quality. For diverse translation evaluation, we utilize \textit{distinct-1} score~\cite{li2016diversity} as the evaluation metric, which is the number of distinct unigrams divided by total number of generated words.

For all translation tasks, we choose the \textit{base} configuration of Transformer with $d_{model}=512$.
During training, we choose Adam optimizer~\cite{Kingma14} with $\beta_1 = 0.9$, $\beta_2 = 0.98$ with initial learning rate is 0.0002 with 4000 warm-up steps. We describe additional implementation and training details in the Appendix.


\subsection{Results on IWSLT'14 Tasks}

Table~\ref{tab:result-iwslt} compares \namelasyn versions against some of the state-of-the-art models on the IWSLT'14 dataset.
\namelasyn-K rows show results when varying the number of EM update steps per batch ($K$).

On the De-En task, \namelasyn provides a 1.7 points improvements over the Transformer baseline, demonstrating that the \namelasyn's improvements come from incorporating target side syntax effectively. This result is also better than a transformer-based source side syntax model by 1.5 points.
\namelasyn results are also better than the published numbers for LSTM-based models that use multi-task learning for source side and models that uses target side syntax. Note that since the underlying architectures are different, we only present these results to show that the results with \namelasyn are comparable with other models that have incorporated syntax.

On the En-De task, our model achieves 29.2 in terms of BLEU score, with 2.6 points improvement over the Transformer baseline and 2.4 points improvement over Transformer-based Source Side Syntax model. Compared with NPMT~\cite{huang2018towards}, which is a BiLSTM based model, we achieve 3.84 point improvement.

On the En-Fr task, our model set a new state-of-the-art with a BLEU score of 40.6, which is 1.7 points improvement over the second best model which uses Transformer to incorporate source side syntax knowledge. Our model also surpasses the basic Transformer model by about 2.1 points.

We notice that across all tasks, the performance of our model improves with number of EM update steps per batch ($K$). With larger $K$ values, we get better lower bounds $\mathcal{L}_{lower}(\boldsymbol{\theta})$ on each training batch, thus leading to better optimization. For update steps beyond $K>5$, the performance does not improve any more.

Last, the EUB row indicates the performance that can be obtained when feeding in the true POS tags. The large improvement here shows the potential for improvement when modeling target side syntax.

\begin{table*}[!t]
\centering
\small
\begin{tabular}{cccccl}
\toprule
\multirow{2}{*}{\textbf{Method Type}} & \multirow{2}{*}{\textbf{Model}} & \multicolumn{3}{c}{\textbf{BLEU}} \\\cline{3-5}
& & \textbf{De-En} & \textbf{En-De} & \textbf{En-Fr}     \\
\midrule
\multirow{4}{*}{BiLSTM} & BiLSTM~\cite{denkowski2017} & -- & -- & 34.8 \\
& Dual Learning~\cite{Wang18dual} & 32.35 & -- & -- \\
& AST~\cite{Cheng18Towards} & -- & -- & 38.03 \\
& NPMT~\cite{huang2018towards} & -- & 25.36 & -- \\
\midrule
Multi-Task (BiLSTM) & MTL-NMT~\cite{Niehues17} & 27.78 & -- & -- \\
\midrule
Source Side Syn. (Transformer) & Source-NMT~\cite{SennrichLinguistic16} & 33.5 & 26.8 & 38.9 \\
\midrule
\multirow{2}{*}{Target Side Syn. (BiLSTM)} & DSP-NMT~\cite{Shu18} & 29.78 & -- & -- \\
& Tree-decoder~\cite{wang2018tree} & 32.65 & -- & -- \\
\midrule
\multirow{6}{*}{Transformer}& Transformer  & 33.3 & 26.6 & 38.5 \\
& \namelasyn (Unsupervised)  & 30.8 & 25.2 & 34.5 \\
& \namelasyn (K=1)  & 34.63 & 28.1 & 39.7 \\
& \namelasyn (K=3)  & 34.91 & 28.9 & 40.4\\
& \namelasyn (K=5)  & \textbf{35.0} & \textbf{29.2} & \textbf{40.6} \\
& \namelasyn \textbf{EUB} & 51.4 & 47.3 & 54.2\\
\bottomrule
\end{tabular}
\caption{\textbf{IWSLT'14 English-German and English-French results} - shown are the BLEU scores of various models on TED talks translation tasks. We highlight the \textbf{best} model in bold.}
\label{tab:result-iwslt}
\end{table*}

Table~\ref{tab:result-de-en-samples} shows one example where \namelasyn produces correct translations for a long input sentence. The output of \namelasyn is close to the reference and the output of \namelasyn when given the gold POS tag sequence is even better, demonstrating the benefits of modeling syntax. The transformer model however fails to decode the later portions of the long input accurately. 
\begin{table*}[!t]
\centering
\small
\begin{tabular}{cp{9cm}cl}
\toprule
SRC & letztes jahr habe ich diese beiden folien gezeigt , um zu veranschaulichen , dass die arktische eiskappe , die für annähernd drei millionen jahre die grösse der unteren 48 staaten hatte , um 40 prozent geschrumpft ist . \\
REF & last year i showed these two slides so that demonstrate that the arctic ice cap , which for most of the last three million years has been the size of the lower 48 states , has shrunk by 40 percent . \\
\midrule
Transformer & last year , i showed these two slides to illustrate that the arctic ice caps that had the size of the lower 48 million states \color{red}{to 40 percent .} \\
\namelasyn  & last year , i showed these two slides to illustrate that the arctic ice cap , \color{blue}{which for nearly three million years} had the size of the lower 48 states , \color{blue}{was shrunk by 40 percent .}  \\
\namelasyn (groundtrue POS)  & last year i showed these two slides just to illustrate that the arctic ice cap , \color{blue}{which for nearly about the last three million years} has been the size of the lower 48 states , \color{blue}{has shrunk by 40 percent .} \\
\bottomrule
\end{tabular}

\caption{Translation examples on IWSLT'14 De-En dataset from our model and the Transformer baseline. We put correct translation segment in blue and highlight the wrong one in red.}
\label{tab:result-de-en-samples}
\end{table*}

\subsection{Speed}
We compare the speeds of our (un-optimized) implementation of \namelasyn with a vanilla transformer with no latents in its decoder. Table~\ref{tab:result-speed} shows the training time per epoch, and the inference time for the whole test set. computed on the IWSLT'14 De-En task. When $K=1$, \namelasyn takes almost twice as much time as the vanilla transformer for training. Increasing $K$ increases training time further. For inference, \namelasyn takes close to four times as much time compared to the vanilla Transformer. In terms of complexity, \namelasyn only adds a linear term (in POS tag size to the decoding complexity. Specifically, its decoding complexity is $B \times O(m) \times O(\vert V_{\mathbf{z}}\vert \times N)$ where $B$ is beam size, $m$ is a constant proportional to the tag set size and $N$ is output size. 
As the table shows, empirically, our current implementation incurs $m \simeq 4$. We leave further optimizations for future work.

\begin{table}[!t]
\centering
\small
\begin{tabular}{cccl}
\toprule
\textbf{Model} & \textbf{Training Time/Epoch}$\downarrow$ & \textbf{Inference Time}$\downarrow$ \\
\midrule
Transformer & \textbf{3.6} min & \textbf{12.8} s\\
\namelasyn (K=1) & 6.3 min & 56.1 s \\
\namelasyn (K=3) & 18.1 min & 55.6 s \\
\namelasyn (K=5) & 28.0 min & 55.6 s \\
\bottomrule
\end{tabular}
\caption{\textbf{IWSLT'14 De-En training and inference speed evaluation}. $\downarrow$ means the smaller the better. We highlight the \textbf{best} model in bold.}
\label{tab:result-speed}
\end{table}

\subsection{Diversity}

We compare the diversity of translations using \textit{distinct-1} score~\cite{li2016diversity}, which is simply the number of distinct unigrams divided by total number of generated words. We use our model to generate 10 translations for each source sentence of the test dataset. We then compare our results with baseline Transformer. The result is shown in Table~\ref{tab:result-diversity}. Much like translation quality, \namelasyn's diversity increases with number of EM updates and is better than diversity of the transformer and a source side encoder model.

\begin{table}[!t]
\centering
\small
\begin{tabular}{ccccl}
\toprule
\multirow{2}{*}{\textbf{Model}} & \multicolumn{3}{c}{\textbf{distinct-1}} \\\cline{2-4}
& De-En & En-De & En-Fr \\
\midrule
Transformer & 0.231 & 0.242 & 0.258\\
Source-NMT & 0.232 & 0.244 & 0.260 \\
\namelasyn (Unsupervised) & 0.228 & 0.231 & 0.239 \\
\namelasyn (K=1) & 0.237 & 0.251 & 0.265 \\
\namelasyn (K=3) & 0.241 & 0.253 & 0.270 \\
\namelasyn (K=5) & \textbf{0.245} & \textbf{0.255} & \textbf{0.273} \\
\namelasyn \textbf{EUB} & 0.328 & 0.516 & 0.354 \\
\bottomrule
\end{tabular}
\caption{\textbf{IWSLT'14 En-De/De-En/En-Fr diversity translation evaluation}. We highlight the \textbf{best} model in bold.}
\label{tab:result-diversity}
\end{table}

\subsubsection{Controlling Diversity with POS Sequences}
One of the main strengths of \namelasyn is that it can generate translations conditioned on a given POS sequence. First, we present some examples that we generate by decoding over different POS tag sequences. Given a source sentence, we use \namelasyn to provide the most-likely target pos tag sequence. Then, we obtain a random set of valid POS tag sequences that differ from this maximum likely sequence by some edit distance. For each of these randomly sampled POS tag sequences, we let \namelasyn generate a translation that fits the sequence. Table~\ref{tab:result-de-en-diverse} shows some example sentences. \namelasyn is able to generate diverse translations that reflect the sentence structure implied by the input POS tags. 
However, in trying to fit the translation into the specified sequence, it deviates somewhat from the ideal translation. 

\begin{figure}[t!]
\centering
\includegraphics[width=0.8\textwidth]{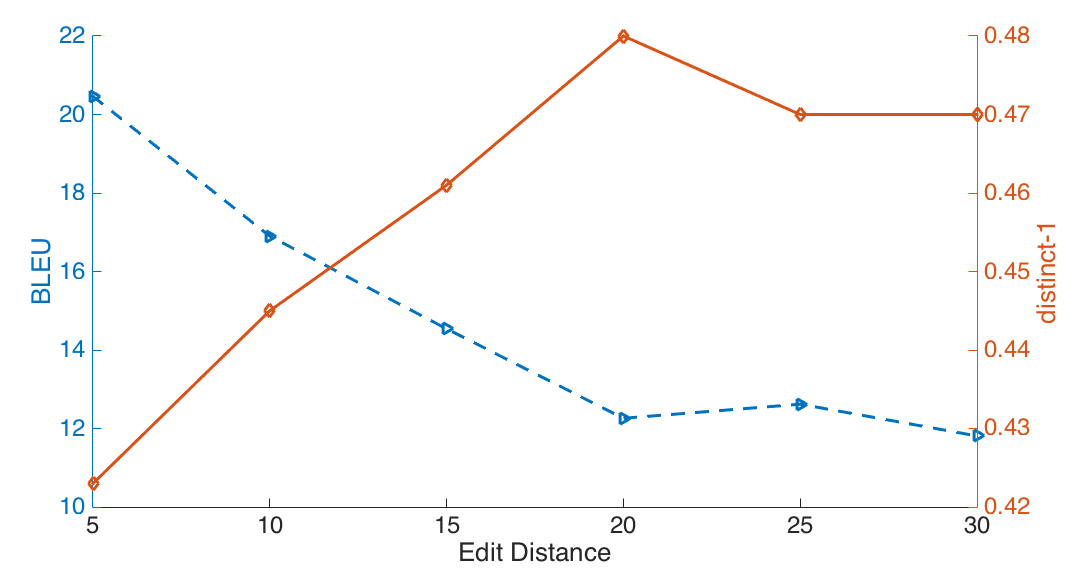}
\caption{Diversity vs. Translation Quality: BLEU and \textit{distinct-1} scores for targets decoded using POS sequences of increasing edit distance.}
\label{fig:edit}
\end{figure}

To understand how diversity plays against translation quality, we also conduct a small scale quantitative evaluation. We pick a subset of the test dataset, and for each source sentence in this subset, we sample $10$ POS tag sequences whose edit distance to their corresponding Top-1 POS tag sequence equal to a specific value, we then use them to decode $W$ translations. We calculate their final BLEU and \textit{distinct-1} scores. 
The results are shown in Figure:~\ref{fig:edit}. As the edit distance increases, diversity increases dramatically but at the cost of translation quality. Since the POS tag sequence acts as a template for generation, as we move further from the most likely template, the model struggles to fit the content accurately. Understanding this trade-off can be useful for re-ranking or other scoring functions. 

\begin{table}[!t]
\centering
\small
\begin{tabular}{cp{9cm}cl}
\toprule
SRC & und natürlich auch , wie nimmt gestaltung einfluss auf die wahrnehmung . \\
REF & and of course how design affects perception . \\
\midrule
$0$ & cc in nn wrb nn vbz vbz nn . \\
 &  and of course how design is affecting perception . \\
$20$ &  ls rb vb in dt nn vbz jj jj jj . \\
 & i also think that the design is affecting perception . \\
 $30$ & rb , dt nn nn dt nn in prp rb . \\
 & also , the way design adds influence in perception too . \\
 $30$ & prp vbz in prp vb vbn rb in nn . \\
 & it turns out it included design impact on perception .\\
\bottomrule
\end{tabular}
\caption{Examples of translations decoded from specified POS sequences with different edit distances (shown as values in first column). SRC: source sentence. REF: reference sentence.}
\label{tab:result-de-en-diverse}
\end{table}

\subsection{Results on WMT'14 En-De}
To assess the impact on a larger dataset, we show results on the WMT'14 English-German in table~\ref{tab:result-wmt-en-de}. Compared to the previously reported systems, we see that our transformer implementation is a strong baseline.
\namelasyn produces small gains, with the best gain at K=5 -- a BLEU score improvement of 0.6. This demonstrates that syntactic information can contribute more to the increase of translation quality on a smaller dataset. 



\begin{table}[!t]
\centering
\begin{tabular}{ccl}
\toprule
\textbf{Model} & \textbf{BLEU} \\
\midrule
BiRNN+GCN~\cite{bastings2017graph} & 23.9 \\
ConvS2S~\cite{Gehring17} & 25.16  \\
MoE~\cite{Shazeer17} & 26.03 \\
Transformer (base) & 27.3 \\
\namelasyn (K=1) (base) & 27.6 \\
\namelasyn (K=3) (base) & 27.8 \\
\namelasyn (K=5) (base) & \textbf{27.9} \\
\bottomrule
\end{tabular}

\caption{\textbf{WMT'14 English-German results} - shown are the BLEU scores of various models on TED talks translation tasks. We highlight the \textbf{best} model in bold.}
\label{tab:result-wmt-en-de}
\end{table}

\section{Conclusion}
Modeling target-side syntax through true latent variables is difficult because of the additional inference complexity. In this work, we presented \namelasyn, a latent syntax model that allows for efficient exploration of a large space of latent sequences. This yields significant gains on four translation tasks, IWSLT'14 English-German, German-English, English-French and WMT'14 English-German. The model also allows for better decoding of diverse translation candidates. This work only explored parts-of-speech sequences for syntax. Further extensions are needed to tackle tree-structured syntax information. 


\chapter{Generic Image Captioning}

\section{Introduction}
Research on image captioning to generate textual descriptions of images has made a great progress in recent years thanks to the introduction of encoder-decoder architectures~\cite{Anderson2018,Aneja18,Johnson16,Karpathy2017,Lu2018NeuralBT,Venugopalan2017,Kelvin2015,Yang2020Fashion}.
Existing models are generally trained and evaluated on datasets created for image captioning  like COCO~\cite{ChenCOCO15,Lin14} and Flickr~\cite{Hodosh13} that only contain generic object categories but not pair-wise relations of the objects in the image.

To equip the captioning model with relation information, some more recent studies resort to the scene graph generation~\cite{xu2017scenegraph,zellers2018scenegraphs} to provide the graph representations of real-world images with the semantic summaries of objects and their pair-wise relationships. 
For example, the graph in Figure~\ref{fig:scene-graph} encodes the  key objects in the image such as people (`man'), their possessions (`hair' and `shirt', both possessed by the man), and their activities (the man is `holding' a `racket'). 
The graph representation has been applied to improve the image related tasks that involve natural language~\cite{Teney17,yin2017obj2text}.
When it comes to the task of image captioning, recent studies~\cite{wang2019role,yang2019,Yao2018} propose to first use a scene graph generation model well-trained on Visual Genome~\cite{Krishna17} dataset to predict the pair-wise relationships existing in the COCO image  and then use a Graph Convolutional Net (GCN) to encode the relation information. 
Typically, the object region features and the relation representations are then merged together via concatenation or convolution to feed into a decoder for generating a sentence using the Maximum Likelihood Estimation (MLE).
These methods typically suffer from at least one of three main weaknesses:
(i) There are mis-alignments between the image objects and the relation labels, because the regions containing the objects do not correspond to those used to predict the relations;
(ii) Given that the goal of using a GCN is to extract the relation information,  \rev{the training of model for GCN is less effective by only using the objective to optimize the captioning} without considering the object relationship;
(iii) The encoder itself cannot extract the relations between objects but relying on other pre-trained models to do it, which makes the captioning less explainable.
As another observation, recent studies~\cite{agrawal2016analyzing,caglayan2019,devlin2015exploring,Goyal2019,shekhar2019} have pointed out that good metric scores can be achieved with a strong decoder, without the need of underlying encoder to truly understand the visual content.
Thus, it becomes less likely to determine if  the models are really learning some important relationships through the encoder or they just follow some language rules by the decoder.
To be more concrete, for a generated sentence like `a man is riding a bike', can the model really tell the difference among `riding', `rolling' or `on' or it just follows some language expression rules (\textit{i.e.}, `riding' is more commonly used than `rolling' and `on')?

Regularizing the encoder with a relation-centric objective is essential since it can not only \rev{guide} the encoder to learn representations with relation information embedded, but also explicitly express the pair-wise relationships and explain the generation of some relational words.
In this paper, we propose \textit{\name}: the RElational transFORMER that learns a scene graph to express the object relationships in the process of decoding a sentence description. 
\name incorporates both image captioning and scene graph generation components via a novel transformer encoder.
Different from conventional Transformer~\cite{Vaswani2017} that only uses the image captioning as the final objective to train both the encoder and the decoder, \name uses a scene graph generation objective to \rev{guide} the encoder to learn better relational representations.
Since the image captioning and scene graph generation are two distinct tasks, directly using the Multi-Task Learning paradigm is non-trivial.
We propose a \textit{sequential training} algorithm that \rev{guides} the \name to learn both tasks step by step.

Our work has three main contributions.
(i) We propose to generate scene graphs as a way to enrich the captions that they together can better describe the images;
(ii) We design a novel relational Transformer (\name) that can better learn the image features for captioning with the relationships embedded via an auxiliary scene graph generation task;
(iii) We propose a \textit{sequential training} algorithm that \rev{guides} the \name to accomplish both tasks in three consecutive steps.
Experimental results show that \name can achieve better performance than state-of-the-art methods on both image caption generation and scene graph generation.
We will release the source code.

\section{Background and Related Work}
In this section, we first introduce the background knowledge of scene graph generation, and then discuss the related work on image captioning and the application of scene graphs in image captioning.
\subsection{Scene Graph Generation}
\label{sec:sgg}
\begin{figure}
    \centering
    \small
    \includegraphics[width=0.8\textwidth]{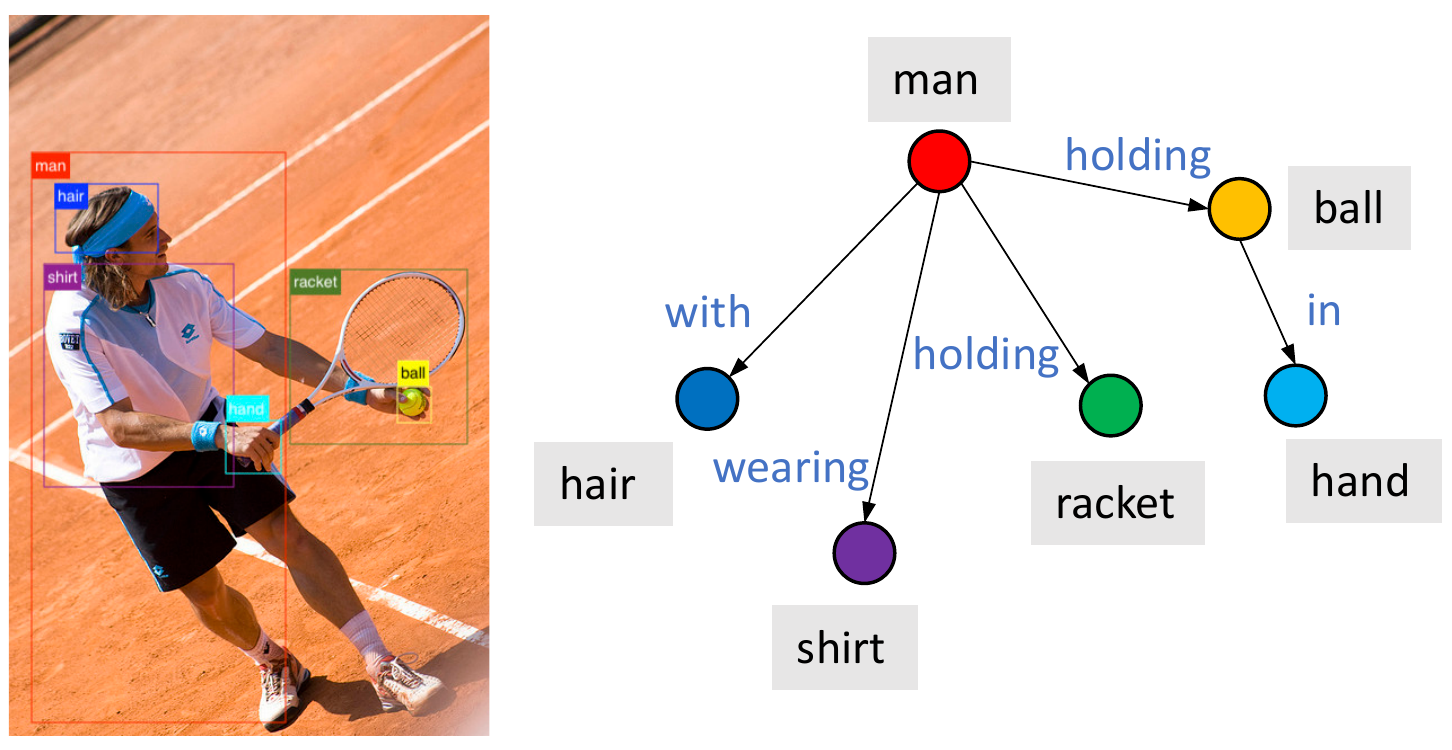}
    \caption{A scene graph containing entities, such as \texttt{man}, \texttt{hair} or \texttt{racket}, that are localized in the image with bounding boxes and the relationships between those entities, such as \texttt{with}, \texttt{wearing} and \texttt{holding}.}
    \label{fig:scene-graph}
\end{figure}
A \textit{scene graph}, $G$, as shown in Figure~\ref{fig:scene-graph}, is a structural representation of the semantic contents in an image~\cite{Krishna17}.
It consists of:
\begin{itemize}
    \item a set of \textit{bounding boxes} $B=\{b_1,\ldots,b_m \}$, $b_i=(x_{i1},y_{i1},x_{i2},y_{i2})$\footnote{$x_{i1},y_{i1}$ are the top-left coordinates of $b_i$, while $x_{i2},y_{i2}$ are the bottom-right coordinates.}.
    \item a corresponding set of \textit{objects} $O=\{o_1,\ldots,o_m \}$, where $o_i\in \mathcal{C}$ is the class label assigned to the bounding box $b_i$ and $\mathcal{C}$ is the set containing all label categories.
    \item a set of pair-wise relationships $R=\{\ldots, r_{i\rightarrow j},\ldots \}$, with $r_{i\rightarrow j}\in \mathcal{R}$ representing the relationship between a start node $(b_i,o_i)$ and an end node $(b_j,o_j)$. $\mathcal{R}$ is the set of relation types, including the `background' predicate, which indicates that there is no edge between the specified objects.
\end{itemize}

Scene graph~\cite{tang2020unbiased,tang2018learning,zellers2018scenegraphs} is often generated with a few procedures: object detection (detecting $b_i$), classification (classifying $o_i$) and predicate (relation label) prediction to determine $r_{i\rightarrow j}$ given $(b_i,o_i)$ and $(b_j,o_j)$.
Most of the methods on scene graph generation have been developed on the Visual Genome~\cite{Krishna17} dataset, which provides annotated scene graphs for $100$K images, consisting of over $1$M instances of objects and $600$K relations.
Since only a small portion of images in this data set also exist in the COCO captioning dataset~\cite{ChenCOCO15}, directly using a multi-task learning scheme on both tasks is challenging.

\subsection{Image Captioning}
State-of-the-art approaches~\cite{Anderson2018,He2020image,Johnson16,Sammani_2020_CVPR,wang2020unique,Kelvin2015,Yang2020Fashion} 
mainly use encoder-decoder frameworks with attention to generate captions for images.
Xu \textit{et al.}~\cite{Kelvin2015} developed soft and hard attention mechanisms to focus on different regions in the image when generating different words.
Similarly, Anderson \textit{et al.}~\cite{Anderson2018} used a Faster R-CNN~\cite{Ren15} to extract regions of interest that can be attended to.
Yang \textit{et al.}~\cite{Yang2020Fashion} used self-critical sequence training for image captioning. 

Various Transformer-based~\cite{Vaswani2017} models have achieved promising success on the
image captioning task~\cite{cornia2020m2,He2020image,Simao2019,Li_2019_ICCV}.
Cornia \textit{et al.}~\cite{cornia2020m2} proposed a meshed-memory transformer that learns a multi-level representation of the image regions, and uses a mesh-like connectivity at decoding stage to exploit low- and high-level features.
Li \textit{et al.}~\cite{Li_2019_ICCV} introduced the entangled attention that enables the Transformer to exploit semantic and visual information simultaneously.
He \textit{et al.}~\cite{He2020image} introduced the image transformer, which consists of a modified encoding transformer and an implicit decoding transformer to adapt to the structure of images.
Herdade \textit{et al.}~\cite{Simao2019} introduced the object transformer, that explicitly incorporates information
about the spatial relationship between detected objects through geometric
attention.

Some research studies have been using scene graphs for image captioning~\cite{guo2019vsua,wang2019role,Yao2018,zhong2020comprehensive}.
Zhong \textit{et al.}~\cite{zhong2020comprehensive} proposes a scene graph decomposition method that decomposes a scene graph into a set of sub-graphs, with each sub-graph capturing a semantic component of the input image.
By selecting important sub-graphs, different target sentences are decoded.
Wang \textit{et al.}~\cite{wang2019role} uses two encoders, one is a ResNet image encoder, the other is a GCN encoder for the relation labels of the objects.
The two encoders are then attached with an LSTM decoder and combined together using attention.
Yao \textit{et al.}~\cite{Yao2018} proposes to use two GCNs to encode the spatial and semantic relations in an image.
Then the features from the two encoders are merged together via attention to get the final feature.
Guo \textit{et al.}~\cite{guo2019vsua}
proposed to explicitly model the object interactions in semantics and geometry based on Graph Convolutional Networks (GCNs). 
Yang \textit{et al.}~\cite{yang2019} proposed the scene graph auto-encoding technique to learn a dictionary that helps to encode the desired language prior, which guides the encoding-decoding pipeline.

\name differs from the previous methods in 
two aspects: (i) \name can not only generate a caption but also a scene graph to capture the relationships between the objects without using an external scene graph generator.
(ii) \name integrates the scene graph generation with image captioning using a \textit{sequential training} algorithm to better learn the relational image features step by step.

\section{Design of \name}
In this section, we first discuss the problems of existing schemes and propose the  basic architecture to construct the \name that combines the scene graph generation with image captioning to learn relational features of images.
We then present the scene graph generation task which is used to first pre-train the encoder of \name and later used as an auxiliary objective to generate captions. 
Finally, we describe the image captioning task as well as the training algorithm to train on both tasks.

\subsection{A Relational Encoding Learning Idea}
\label{sec:intuition}
In a standard  image captioning model based on encoder-decoder structure, the decoder directly predicts the target sequence $\mathbf{y}$ conditioned on the source input $\mathbf{x}$. 
The captioning probability $P(\mathbf{y}\vert \mathbf{x})$ is modeled directly using the probability of each target word $\mathbf{y}_i$ at the time step $i$ conditioned on the source input sequence $\mathbf{x}$ and the current partial target sequence $\mathbf{y}_{1:i-1}$ as follows:
\begin{equation}
\small
P(\mathbf{y} \vert \mathbf{x}; \boldsymbol{\theta}) = \prod_{i=1}^{N} P(\mathbf{y}_i\vert \mathbf{x}, \mathbf{y}_{1:i-1};\boldsymbol{\theta})
\label{eq:img_cap}
\end{equation}
where $\boldsymbol{\theta}$ denotes the parameters of the model\footnote{Through-out this paper, we omit $\boldsymbol{\theta}$ for simplicity.}.

\begin{figure}
\centering
\small
\begin{subfigure}[b]{0.8\textwidth}
   \includegraphics[width=1\linewidth]{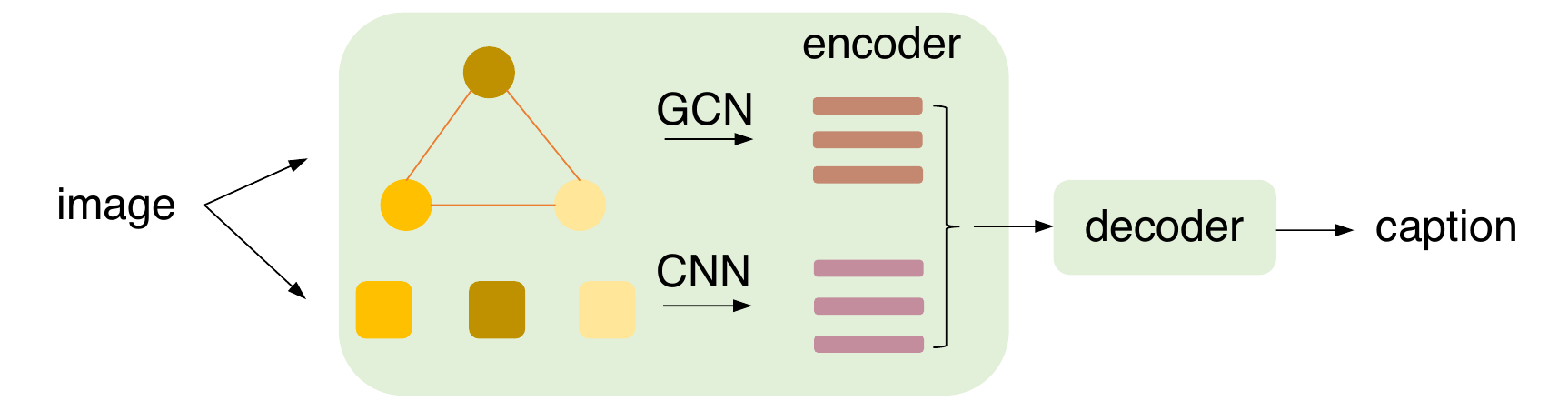}
   \caption{Two-encoder captioning model.}
   \label{fig:gcn} 
\end{subfigure}

\begin{subfigure}[b]{0.8\textwidth}
   \includegraphics[width=1\linewidth]{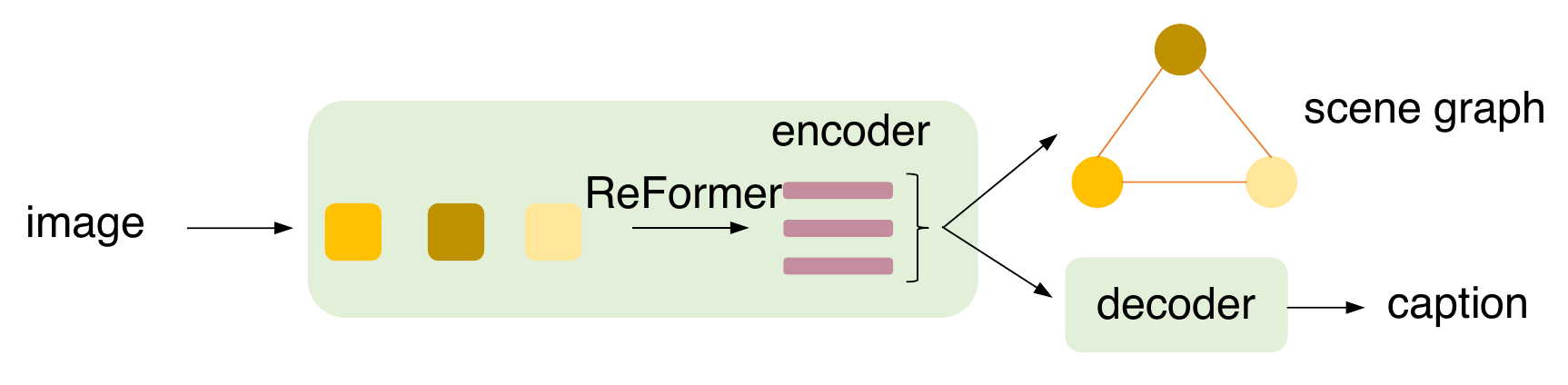}
   \caption{\name: Our Relational Transformer model.}
   \label{fig:reformer}
\end{subfigure}

\caption[]{\small{Models trying to encode relation information: (a) A widely-used two-encoder captioning model that one encoder is a GCN used to encode the scene graphs into relational features and the other is a pre-trained CNN (\textit{i.e.} ResNet) for images.
(b) \name, our Relational Transformer that generates a scene graph and a corresponding caption.}}
\end{figure}

In general, $\mathbf{x}$ are image features that can be obtained by feeding an image $X$ to a pre-trained CNN (\textit{e.g.} ResNet) encoder $\mathbf{x}=CNN(X)$.
To integrate relational information into the image features, some state-of-the-art methods first use a well-trained scene graph generation model to extract a graph $g$ from the same image and then use a Graph Convolutional Net (GCN) to encode $g$ to vectors, shown in Figure~\ref{fig:gcn}.
Then the image features and the relational features are merged together $\mathbf{x}=[CNN(X);GCN(g)]$, where $[\cdot;\cdot]$ can be concatenation, attention or convolution.
This straightforward way of integrating relational information may suffer from a few problems.
First, training a GCN with the objective of getting the image caption (Maximum Likelihood Estimation, \textit{i.e.}, MLE) might be less effective.
The goal of using a GCN is to capture the relational information existing in the image,
while the likelihood function used to estimate the probability distribution is not directly relevant to the relation of objects in an image.
Second, if the model no longer has access to the pre-trained scene graph generator (\textit{e.g.}, using a different dataset), the caption generation might not be feasible any more.
Third, recent studies show that good metric scores can be obtained with a strong decoder, without the underlying encoder to truly understand the visual content. This indicates that the caption objective alone cannot effectively guide the training of the encoder to accurately extract the relationships between the objects. 

The above analysis motivates us to design a model that can generate scene graphs together with the learning of image caption using the same dataset  without the need of an external scene graph generator, but with the objective of producing the good scene graph. 
In this way, we can also better train the encoder to well understand the visual contents without being misled by the good results from a stronger decoder.
We propose a second objective $P(g \vert X)$ for good  scene graph generation and apply it to the encoder learning.
We use a Transformer model where the encoder is used to generate scene graphs and the decoder is applied to generate captions.
By incorporating this `RElational' objective, our transFORMER can also better learn the image features with relation information embedded. 
The simplified model structure of \name is shown in Figure~\ref{fig:reformer}.


\begin{figure*}
    \centering
    \includegraphics[width=0.95\textwidth]{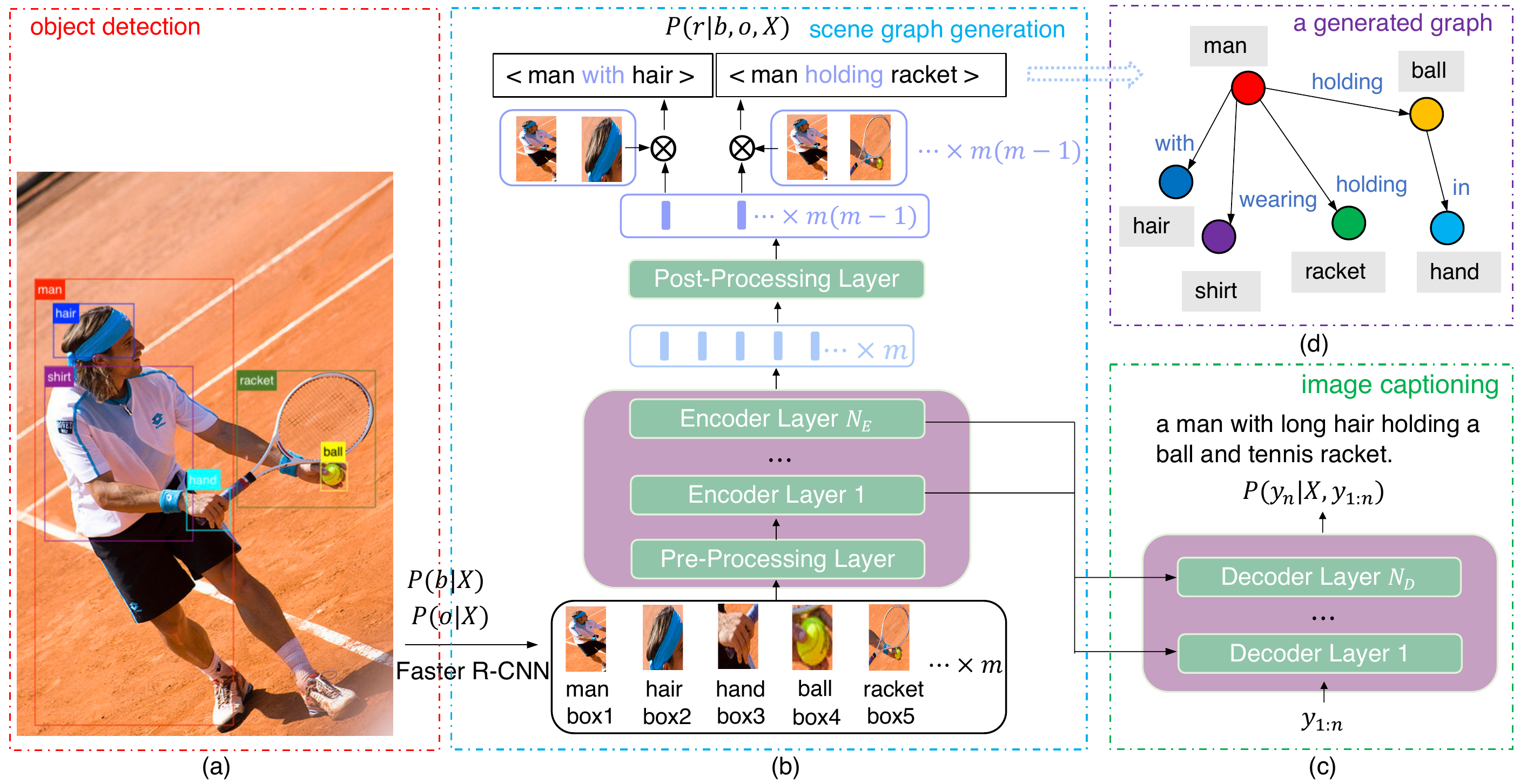}
    \caption{\small{The architecture of \name. \name consists of a (a) Faster R-CNN object detection model to provide $m$ bounding boxes, object labels and object region features, a (b) Transformer encoder to generate $m\times (m-1)$ pair-wise relations for the objects in the image and a (c) Transformer decoder for generating captions. The generated scene graph is shown in (d).}}
    \label{fig:model}
\end{figure*}

\subsection{Scene Graph Generation}
\label{sec:sggpt}

We first lay out the mathematical formulation of the scene graph generation problem.
As introduced in Section~\ref{sec:sgg}, for an image of $m$ objects, its visually grounded scene graph consists of tuples $\big(r_{i\rightarrow j}, (b_i, o_i), (b_j, o_j)\big)_{i,j=1,i\neq j}^m$\footnote{$i\in {1\ldots m}$, $j\in {1\ldots m}$, but $i\neq j$.}, with $b_i$ being the bounding box of the object $i$, $o_i$ the label and $r_{i\rightarrow j}$ the predicate (relation label) between objects $i$ and $j$.
Thus, the scene graph generation objective can be derived as follows:
\begin{equation}
\footnotesize
\begin{aligned}
    P(g \vert X)
    =& \prod_{(i,j)}^{m} P(r_{i\rightarrow j}\vert (b_i,o_i),(b_j,o_j),X) \\
    &\times \prod_{k}^{m}P(o_k\vert b_k,X)P(b_k\vert X)
\end{aligned}
\label{eq:sgg_obj}
\end{equation}
To simplify the learning process and improve the effectiveness of training, we conveniently use Faster-RCNN to model $P(o_k\vert b_k, X)$ and $P(b_k\vert X)$ together.
The negative log-likelihood of the model's parameters computed on the training data is then given by:
\begin{equation}
\footnotesize
\begin{aligned}
    \mathcal{L}_g&=-\log P(g \vert X) 
    = -\sum_{(i,j)}^{m} \log P(r_{i\rightarrow j}\vert (b_i,o_i),(b_j,o_j),X)\\ 
    &- \sum_{k}^{m}[\log P(b_k\vert X) + \log P(o_k\vert b_k, X)]
\end{aligned}
\label{eq:sgg_llh}
\end{equation}
We model $P(b_k\vert X)$ and $P(o_k\vert b_k, X)$ with object detector Faster R-CNN~\cite{Ren15} to automatically generate a set of bounding boxes $b_k$ and the corresponding object labels $o_k$ from an image $X$ (Figure~\ref{fig:model}(a)).
In practice, training a Faster R-CNN is highly non-trivial. To simplify the training of our proposed \name, we first train  a Faster R-CNN till its convergence. Then with its parameters learnt, we train the model $P(r_{i\rightarrow j}\vert (b_i,o_i), (b_j, o_j),X)$ (Figure~\ref{fig:model}(b)) with $(b_i, o_i)$, $(b_j,o_j)$ and $X$ as the input.
In theory, if the groundtruth $b_k$ and $o_k$ are provided, we can eliminate Faster R-CNN. 
To feed in $b_k$, $o_k$ and $X$, we use a Pre-Processing Layer (Section~\ref{sec:ppl}) which converts the three inputs into vectors and then concatenate them together to get the final input features.



Modeling $P(r_{i\rightarrow j}\vert(b_i,o_i),(b_j,o_j),X)$ is actually a predicate classification task.
We use the \name encoder (Section~\ref{sec:el}) to first encode the input into high-level features and then apply a Post-Processing Layer (Section~\ref{sec:popl}) to get the final output for classification.
The features right before the Post-Procesing layer contain useful relational information and are used as the input to the decoder for caption generation (Section~\ref{sec:ic}).
We talk about more details about the model architecture in Section~\ref{sec:re}.

\subsection{Encoder Architecture}
\label{sec:re}

The encoder takes as input a set of $m$ image region tuples,
$(b_i, o_i, \mathbf{v}_i)_{i=1}^m$, where $\mathbf{v}_i$ is defined as the mean-pooled convolutional feature from region $i$ with dimension $2048$ and the number of region varies for different images.
It consists of three main components which process the input consecutively:
(i) a Pre-Processing Layer that takes the bounding boxes, object labels and image region features as input and linearizes them to form the input vector; (ii) the Encoder Layers further process the features  with Multi-Head Attention to create a contextualized representation of each object; and (iii) the Post-Processing Layer where the object features are paired to make predictions of their relationships.

\subsubsection{Pre-Processing Layer}
\label{sec:ppl}
Different from conventional image captioning model where only image region features $\mathbf{v}_i$ are used as the input, we also use object labels $o_i$ as well as bounding boxes $b_i$ as the input to encode both object information and spatial relation information.

Given the bounding box $b=(x_1, y_1, x_2, y_2)$, to better represent its location as well as size in the image, we normalize it with the size of the image and convert it into a $9$-dimensional vector $(\frac{c_x}{W}, \frac{c_y}{H}, \frac{w}{W}, \frac{h}{H}, \frac{x_1}{W}, \frac{y_1}{H}, \frac{x_2}{W}, \frac{y_2}{H}, \frac{wh}{WH})$, where$(c_x, c_y)$ is the coordinate of the bounding box center, $(w, h)$ and $(W, H)$ are the width and height of the bounding box and the image respectively. 
We use a two-layer feed-forward net to encode it into vector $\mathbf{v}_b$ of dimension $d_b$.

As object labels (\textit{e.g.} `man') are meaningful words from natural languages, we use Glove~\cite{pennington2014glove} features, which contain pre-trained features to capture fine-grained semantic and syntactic regularities in natural languages.
The dimension of the object label feature $\mathbf{v}_l$ is denoted as $d_l$.

Thus, the final image feature is represented as $\mathbf{v}=f([\mathbf{v}_i; \mathbf{v}_b; \mathbf{v}_l])$, where $[\cdot;\cdot;\cdot]$ is the concatenation operation, and $f(\cdot)$ is a feed-forward network to map the feature to dimension $d_h$.

\subsubsection{Transformer Encoder Layer}
\label{sec:el}
Given a set of image features $\mathbf{v}=\{\mathbf{v}_1,\ldots, \mathbf{v}_m\}$ extracted using the Pre-Processing Layer, we use
a stack of $N_E$ Transformer~\cite{Vaswani2017} encoder layers to obtain a permutation invariant encoding.
Each encoder layer consists of a multi-head self-attention layer followed by a small feed-forward network. 
The multi-head self-attention layer itself consists of $h$ identical heads. 
Each attention head first calculates the queries $\mathbf{Q}$, keys $\mathbf{K}$ and values $\mathbf{V}$ for the $m$ input features as follows:
\begin{equation}
\small
    \mathbf{Q}=\mathbf{X}\mathbf{W}_Q,\mathbf{K}=\mathbf{X}\mathbf{W}_K,\mathbf{V}=\mathbf{X}\mathbf{W}_V,
    \label{eq:qkv}
\end{equation}
where $\mathbf{X}$ contains all the input vectors $\mathbf{x}_1,\ldots,\mathbf{x}_m$ stacked into a matrix and $\mathbf{W}_Q$, $\mathbf{W}_K$, $\mathbf{W}_V$ are learned projection matrices.
The image features are used as the input to the first self-attentive layer.
For the following layers, we use the output of the previous encoder layer as the input to the current layer.
The output of each head are then computed via scaled dot-product attention without using any recurrence:
\begin{equation}
\small
    head(\mathbf{X})=Attention(\mathbf{Q},\mathbf{K},\mathbf{V})=softmax(\frac{\mathbf{Q}\mathbf{K}^{T}}{\sqrt{d}})\mathbf{V}
    \label{eq:att}
\end{equation}
where $d$ is a scaling factor.
Eq.~\ref{eq:qkv} and \ref{eq:att} are calculated for every head independently. 
The output of all $h$ heads are then concatenated to one output vector and multiplied with a learned projection matrix $\mathbf{W}_O$:
\begin{equation}
\small
    MultiHead(\mathbf{X})=Concat(head_1,\ldots,head_h)\mathbf{W}_O
    \label{eq:mha}
\end{equation}
The multihead attention is then fed into a point-wise feed-forward network (FFN), which is
applied to each output of the attention layer:
\begin{equation}
\small
    \mathbf{h}=FFN(\mathbf{x})=\max(0, \mathbf{x}\mathbf{W}_1+\mathbf{b}_1)\mathbf{W}_2+\mathbf{b}_2
\end{equation}
where $\mathbf{W}_1$, $\mathbf{b}_1$ and $\mathbf{W}_2$, $\mathbf{b}_2$ are the weights and biases of two fully connected layers. 
In addition, skip-connections and layer-norm are applied to the outputs of the self-attention and the feed-forward
layer.


Other choices like LSTM or convolutional layers are also feasible.
In this paper, we choose self-attention layers because self-attention operation can be seen as a way of encoding pair-wise relationships between input features given that the self attention weights depend on the pair-wise similarities between the input features.
This is coherent to the objective of the paper: to model the relationships from the input objects.

\subsubsection{Post-Processing Layer}
\label{sec:popl}
For an image of $m$ objects, there are $m(m-1)$ possible relationships.
For each possible relationship between object $i$ and $j$, we compute the probability of the relationship of label $r_{i\rightarrow j}$.
Since the output of the last Transformer encoder layer $\mathbf{h}$ contains $m$ features $\mathbf{h}_1,\ldots,\mathbf{h}_m$, directly predicting the relationships using these features is impossible.
We thus use a Post-Processing Layer that maps $m$ object features into $m(m-1)$ pair-wise features.
We use a Linear layer to map $\mathbf{h}_i$ of dimension $d_h$ into $\mathbf{r}_i$ of dimension $2d_h$.
By doubling the dimension of $\mathbf{h}_i$, we can then equally split $\mathbf{r}_i$ into two parts, head $\mathbf{r}_i^h$ and tail $\mathbf{r}_i^t$, with head standing for the relationship starts from this node while tail stands for the relationship ends at this node.
Thus, for each pair-wise relationship representation, we can have:
\begin{equation}
\small
    \mathbf{r}_{i\rightarrow j}=(\mathbf{W}_r[\mathbf{r}_i^h;\mathbf{r}_j^t])\odot[\mathbf{v}_i;\mathbf{v}_j]
\end{equation}
where $\odot$ is point-wise multiplication operation, $[\cdot ; \cdot]$ is concatenation and $\mathbf{W}_r$ is a trainable parameter.
The distribution is:
\begin{equation}
\small
    P(r_{i\rightarrow j}\vert (b_i,o_i), (b_j,o_j),X)=softmax(\mathbf{r}_{i\rightarrow j})
\end{equation}

\subsection{Weighted Decoder for Image Captioning}
\label{sec:ic}
As discussed in Section~\ref{sec:intuition},
the objective of our model for caption generation is to minimize the negative log-likelihood of the correct caption using the maximum likelihood estimation: 
\begin{equation}
\small
    \mathcal{L}_{c} = -\sum_{i=1}^{n}\log p(\mathbf{y}_i\vert \mathbf{y}_{1:n-1}, \mathbf{x})
\label{eq:loglikelihood}
\end{equation}
The decoder that is used to model Eq.~\ref{eq:loglikelihood} consists of a stack of $N_D$ decoder layers (Figure~\ref{fig:model}(c)).
For each layer, to calculate the distribution for the word at the time step $i$, it takes as input: the embeddings of all previously generated words $\mathbf{y}_{0:i-1}$ and the context embeddings $\mathbf{x}$ from the encoder.

For conventional Transformer~\cite{Vaswani2017}, $\mathbf{x}=\mathbf{h}^{N_E}_{1:m}$, which means the decoder layers only take the output of the final layer (\textit{i.e.} the $N_E$-th) as input.
This might omit some of the useful features from the lower encoder layers.
In this paper, given the outputs from all the encoder layers, $\{\mathbf{h}^{1:N_E}_{1:m} \}$, we take a weighted sum across all layers to obtain
the final image feature as:
\begin{equation}
\small
    \mathbf{x}_i=\alpha_l\sum_{l=1}^{N_E}\mathbf{h}^l_i
\end{equation}
where $\alpha_l$ are weights obtained using a softmax layer.

\subsection{Sequential Training with Inferred Labels}
\label{sec:training}

Training \name involves three steps: (i) training the Faster R-CNN object detector on Visual Genome dataset;
(ii) the trained Faster R-CNN is applied to train the encoder with Eq.~\ref{eq:sgg_llh} on Visual Genome dataset; (iii) when the encoder is well trained, it is further trained with the joint objective of scene graph generation and image captioning following Eq.~\ref{eq:total} on COCO dataset.
As the relation labels are not available on COCO dataset, we infer the labels for objects in COCO dataset using the encoder trained in step (ii).

The overall loss function in step (iii) is:
\begin{equation}
\small
    \mathcal{L} = \mathcal{L}_c + \lambda\mathcal{L}_r
\label{eq:total}
\end{equation}
with $\lambda$ being a hyper-parameter.

One possible variant of this training algorithm is to only use $\mathcal{L}_c$ without $\mathcal{L}_r$ as the training loss, as done in the literature work.
However, our Ablation studies in Tab.~\ref{tab:ablation} show that this variant cannot achieve as good results as those using the proposed training algorithm.
The main purpose of incorporating $\mathcal{L}_r$ is to ensure that the relational features learned do not vary much in step (iii) while being used to generate an accurate scene graph.

\section{Experiments}
\subsection{Datasets}

We evaluate \name on two large-scale publicly available datasets: MS COCO~\cite{Lin14} which is an image captioning dataset and Visual Genome~\cite{Krishna17} which is a scene graph generation dataset.

\textbf{COCO.} The dataset is the most popular benchmark for image captioning, which contains $82,783$ training images and $40,504$ validation images.
There are $5$ human annotated descriptions per image.
As the annotations of the official test set are not publicly available, we follow the widely used splits provided by~\cite{Karpathy2017}, where $5,000$ images are used for validation, $5,000$ for testing and the rest for training.
We convert all the descriptions in the training set to lower case and discard
rare words which occur less than 5 times, resulting in the final vocabulary with
$10,201$ unique words in the COCO dataset.
We also evaluate our model on the COCO online test server.
The results of the online evaluation are shown in the Supplementary.

\textbf{Visual Genome.} The dataset is a large-scale image dataset for modeling the relationships between objects, which contains $108$K images with densely
annotated objects, attributes, and relations. 
To pre-train the Faster R-CNN object detector, we take $98$K for training, $5$K for validation and $5$K for testing. 
As part of images (about $50$K) in the Visual Genome are also found in COCO, the split of the Visual Genome is carefully selected to avoid contamination of the COCO validation
and test sets. 
We perform extensive cleaning and filtering of
training data, and train Faster R-CNN over the selected $1,600$ object classes. 
To pre-train the encoder of \name on the scene graph generation task, we adopt the same data split for training the object detector. 
Moreover, we select the top-$50$ frequent predicates in training data. 
The semantic relation detection model is thus trained over the $50$ relation classes plus a non-relation class.

\subsection{Methods \& Metrics}

We compare against three types of baselines.
(i) The CNN-LSTM~\cite{hochreiter1997long} based models: Up-Down~\cite{Anderson2018} which uses attention over regions of interest, NBT~\cite{Lu2018NeuralBT} that first generates a sentence `template' and then fill in by visual concepts identified by object detectors, Att2all~\cite{Rennie2017} that uses self-critical sequence training for image captioning, and AoA~\cite{huang2019attention} which uses attention on attention for encoding image regions and an LSTM language model;
(ii) Transformer-based models:
$\mathcal{M}^2$-T~\cite{cornia2020m2} which uses a mesh-like connectivity to learn prior knowledge, 
Image-T~\cite{He2020image}, an image transformer,
Object-T~\cite{Simao2019} that models the spatial relationship between objects, and
ETA~\cite{Li_2019_ICCV} which proposes the entangled attention mechanism;
(iii) The GCN-LSTM based models: VSUA~\cite{guo2019vsua} that uses GCNs to model the semantic and geometric interactions of the objects, GCN~\cite{Yao2018} which exploits pairwise relationships between image regions through a GCN, and SGAE~\cite{yang2019} which instead uses auto-encoding scene graphs.

For the caption generation evaluation, we follow the other baselines and use the BLEU-1 and BLEU-4 \cite{papineni2002}, ROUGE \cite{lin2004rouge}, METEOR \cite{denkowski2014meteor}, CIDEr \cite{Vedantam15cider} and SPICE~\cite{spice2016} metrics.

To evaluate the Scene Graph Generation, we divide it into three sub-tasks~\cite{zellers2018scenegraphs}: 
(i) Predicate Classification (PredCls), to predict $r_{i\rightarrow j}$ with $(b_i, o_i)$ and $(b_j, o_j)$ given;
(ii) Scene Graph Classification (SGCls), to predict the object labels $o_i$ and $o_j$ and the relationship $r_{i\rightarrow j}$ with $b_i$ and $b_j$ given;
(iii) Scene Graph Detection (SGDet), to directly predict $(b_i, o_i)$, $(b_j, o_j)$ and $r_{i\rightarrow j}$ from an image without the groundtruth bounding boxes or object labels.

The evaluation metrics we report are recall @$x$, where $x=20,50,100$.
Recall @$x$ computes the fraction of times the correct relationship is predicted in the top $x$ confident relationship predictions. 
It was first proposed in \cite{lu2016visual} and then widely adopted in other papers \cite{zellers2018scenegraphs,tang2018learning,tang2018learning}.
We notice that mean average precision (mAP) is another widely used metric. 
However, mAP is a pessimistic evaluation metric because we can not exhaustively annotate all possible relationships in an image. 
We thus do not report the results using this metric.
\subsubsection{Implementation Details}


\begin{table}[!t]
\footnotesize
\centering
\begin{tabular}{ccccccl}
\toprule
Method & B-1  & B-4 & M & R & C & S  \\
\midrule
Up-Down~\cite{Anderson2018} & 79.8 & 36.3 & 27.7 & 56.9 & 120.1 & 21.4 \\
Att2all~\cite{Rennie2017} & -- --  & 34.2 & 26.7 & 55.7 & 114.0 & -- --  \\
NBT~\cite{Lu2018NeuralBT}  & 75.5 & 34.7 & 27.1 & 54.7 & 107.2 & 20.1  \\
AoA~\cite{huang2019attention} & 80.2 & 38.9 & 29.2 & 58.8 & 129.8 & 22.4 \\
\midrule
ETA~\cite{Li_2019_ICCV} & 81.5 & 39.3 & 28.8 & 58.9 & 126.6 & 22.7 \\
Object-T~\cite{Simao2019} & 80.5 & 38.6 & 28.7 & 58.4 & 128.3 & 22.6 \\
Image-T~\cite{He2020image} & 80.8 & 39.5 & 29.1 & 59.0 & 130.8 & 22.8 \\
$\mathcal{M}^2$-T~\cite{cornia2020m2} & 80.8 & 39.1 & 29.2 & 58.6 & 131.2 & 22.6 \\
\midrule
GCN~\cite{Yao2018} & 80.5 & 38.2 & 28.5 & 58.3 & 127.6 & 22.0  \\
SGAE~\cite{yang2019} & 80.8 & 38.4 & 28.4 & 58.6 & 127.8 & 22.1  \\
VSUA~\cite{guo2019vsua} & -- -- & 38.4 & 28.5 & 58.4 & 128.6 & 22.0 \\
\midrule
\name & \textbf{82.3} & \textbf{39.8} & \textbf{29.7} & \textbf{59.8} & \textbf{131.9} & \textbf{23.0}  \\
\bottomrule
\end{tabular}
\caption{\small{Results on COCO dataset. We only report the single model results on the `Karpathy' test split. We highlight the \textbf{best} model in bold.}}
\label{tab:coco}
\end{table}

To represent image regions, we use Faster R-CNN with ResNet-101 finetuned on the Visual Genome dataset, thus obtaining a $2048$-dimensional feature vector for each region. 
To represent words, we use one-hot vectors and linearly project them to the input dimensionality of $512$. 
The dimension of the encoded bounding box, object label and the final image feature are $d_b=100$, $d_l=300$, and $d_h=512$ respectively.
We set the dimension of each layer to $d=512$ and the number of heads to $h=8$. 
We use the same number of encoders and decoders, thus having $N_E=N_D=3$.
We employ dropout with a probability $0.9$ after each attention and feed-forward layer. 
Training with the overall objective function (Eq.~\ref{eq:total}) is done following the learning rate scheduling strategy of~\cite{Vaswani2017} with a warmup equal to $10$K iterations. 
Then, during CIDEr optimization, we use a fixed learning rate of $5\times 10^{-6}$. 
We train all models using the Adam~\cite{kingma2015} optimizer.
We use Glove~\cite{pennington2014glove} embedding to initialize word embedding layer.
The total number of objects in one image varies from $10$ to $50$, depending on the IOU threshold that is set to $0.3$.
After some parameter tuning, we fix $\lambda=0.1$ at which \name provides the best CIDEr score.

\subsection{Evaluation}


\subsubsection{General Caption Generation}

We first evaluate our model with the general caption generation metrics.
We first compare the performances of our \name with those of several recent proposals for image captioning on the COCO `Karpathy' test split.
As shown in Tab.~\ref{tab:coco}, 
in general, the Transformer-based models outperforms other two types of baselines: the models using pre-trained CNN to encode image information and the LSTM decoder with attention to decode a caption and those using GCN to encode scene graph information and the LSTM decoder to generate a sentence.
This proves that Transformer can be used to better learn  high-level image features and is capable of decoding sentences with \rev{a} higher quality.
The GCN-based models outperform the CNN-based ones but \rev{perform worse than the ones using the Transformer}, which indicates that the scene graph information is useful for better learning the image features but still \rev{not well} explored.
Our proposed \name outperforms all other models. 
For example, it provides an improvement of $1.5$, $0.7$, $0.5$, $1.2$, $0.7$ and $0.4$ points over baseline model $\mathcal{M}^2$-T~\cite{cornia2020m2} on $6$ metrics respectively.
This demonstrates the effectiveness of concurrently exploiting the Transformer structure and scene graphs in extracting the relational image features.

\subsubsection{Ablation}
\begin{table}[!t]
\footnotesize
\centering
\begin{tabular}{cccccl}
\toprule
Method & B-4 & M & R & C & S  \\
\midrule
Trans.-2  & 35.7 & 27.4 & 56.4 & 121.3 & 20.5  \\
Trans.-3 & 36.5 & 27.8 & 57.0 & 123.6 & 21.1  \\
Trans.-4  & 36.3 & 27.6 & 56.5 & 121.5 & 20.8  \\
Trans.-5 & 36.1 & 27.5 & 56.8 & 121.9 & 20.7  \\
\midrule
Weighted Trans.  & 37.3 & 28.4 & 57.5 & 125.4 & 21.7  \\
\midrule
\name & \textbf{39.8} & \textbf{29.7} & \textbf{59.8} & \textbf{131.2} & \textbf{23.0}  \\
\midrule
\name $^{\ast}$ & 38.5 & 28.7 & 58.3 & 128.9 & 22.1  \\
\name $ - \mathcal{L}_r$ & 38.9 & 28.8 & 58.7 & 129.4 & 22.3  \\
\midrule
\name $-$ Weighted & 39.3 & 29.2 & 58.9 & 130.3 & 22.5  \\
\bottomrule
\end{tabular}
\caption{\small{Ablation study and comparison of \name variants. Results are reported on the `Karpathy' test split. $^{\ast}$ denotes that we fix the encoder during the caption training. We highlight the \textbf{best} model in bold.} }
\label{tab:ablation}
\end{table}

We first do ablation studies on the number of Transformer layers.
We start from the vanilla Transformer without using any other techniques proposed in the paper.
We vary the number of encoder and decoder layers from 2 to 6.
As shown in Tab.~\ref{tab:ablation}, the Transformer model with $3$ layers achieves the best results.
To evaluate the importance of keeping $\mathcal{L}_c$ in step (iii) of  Section~\ref{sec:training}, we conduct two ablation experiments.
We first keep the parameters of the encoder fixed, which means $\mathcal{L}_r$ is not used to update the parameters of the encoder.
We denote this variant as \name $^{\ast}$.
We find that the CIDEr score drops from $131.2$ to $128.9$.
We then remove $\mathcal{L}_r$ and only use $\mathcal{L}_c$ to update the parameters of the encoder and the decoder.
We denote this case as \name $-\mathcal{L}_r$.
We find that the CIDEr score drops as well.
We can thus conclude that using $\mathcal{L}_r$ helps to improve the quality of caption generation.


\subsubsection{Scene Graph Generation Evaluation}
\begin{table}[!t]
\footnotesize
\centering
\setlength\tabcolsep{1pt}
\begin{tabular}{cccccccccl}
\toprule
\multirow{2}{*}{Method} & \multicolumn{3}{c|}{\centering\textbf{SGGen}} & \multicolumn{3}{c|}{\centering\textbf{SGCls}} & \multicolumn{3}{c}{\centering\textbf{PredCls}} \\
\cmidrule{2-10}
 & R@20 & R@50 & R@100 & R@20 & R@50 & R@100 & R@20 & R@50 & R@100  \\
\midrule
IMP & 14.6 & 20.7 & 24.5 & 31.7 & 34.6 & 35.4 & 52.7 & 59.3 & 61.3  \\
MOTIFS  & 21.4 & 27.2 & 30.3 & 32.9 & 35.8 & 36.5 & 58.5 & 65.2 & 67.1  \\
VCTree & 22.0 & 27.9 & 31.3 & 35.2 & 38.1 & 38.8 & 60.1 & 66.4 & \textbf{68.1} \\
\name & \textbf{25.4} & \textbf{33.0} & \textbf{37.2} & \textbf{36.6} & \textbf{40.1} & \textbf{41.1} & \textbf{60.5}	& \textbf{66.7}	& \textbf{68.1}  \\
\bottomrule
\end{tabular}
\caption{\small{Scene graph generation evaluation results on Visual Genome dataset. 
We highlight the \textbf{best} model in bold.}}
\label{tab:sgg}
\end{table}

Generating scene graphs is essential not only because it provides a way of explaining the relationships between the objects in the image, but also the resource of the scene graphs when there is no other state-of-the-art scene graph generator available.
To evaluate the proposed \name in generating scene graphs, we compare it with other state-of-the-art scene graph generation methods.
IMP~\cite{xu2017scenegraph} solves the scene
graph inference problem using standard RNNs and learns to iteratively improves its predictions via message passing.
MOTIFS~\cite{zellers2018scenegraphs} is a stacked bi-directional LSTM architecture designed to capture higher order motifs in
scene graphs.
VCTree~\cite{tang2018learning} is a dynamic tree structure that places the objects in an image into a visual context which helps to improve the scene graph generation task.
The results are shown in Tab.~\ref{tab:sgg}.
\name achieves better results than all other three baselines on all the three tasks: \textbf{SGGen}, \textbf{SGCls} and \textbf{PredCls}.
This demonstrates the capability of our \name to exploit the relationships between the objects in the image.

\subsubsection{Qualitative Evaluation}

\begin{figure}
\small
    \centering
    \includegraphics[width=0.9\textwidth]{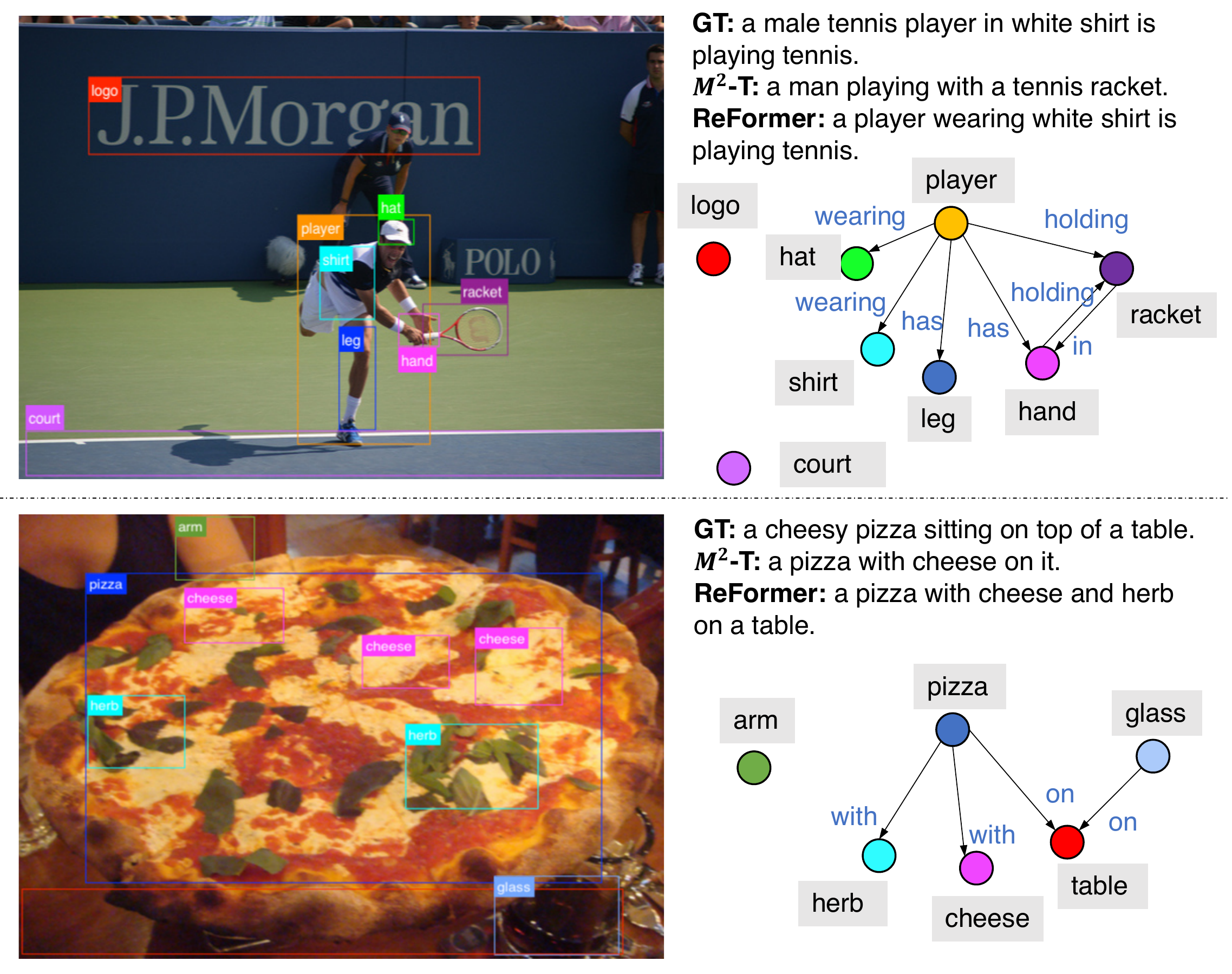}
    \caption{\small{Examples of captions generated by \name and the $\mathcal{M}^2$-T model, as well as the corresponding ground-truths. \name generates a scene graph to show the pair-wise relationships.}}
    \label{fig:sample}
\end{figure}

In Figure~\ref{fig:sample}, we show the image, groundtruth caption and the caption generated by $\mathcal{M}^2$ Transformer and \name.
Our model is able to not only generate a meaningful and more accurate caption than the baseline, but also generate a scene graph showing the relationships between the objects in the image.
With the graphs generated, captions become more expressive and explainable.
Interestingly, the graphs usually contain useful information to describe the image.
For instance, in the first example, the scene graph generated by \name tells us that the player is wearing a shirt and he is holding a racket.
While in the second example, the graph shows that there is a glass on the table and an arm from a person.
These information cannot be inferred from only the caption generated.
Since the average number of words in a sentence in the COCO dataset is $\approx 10$, it becomes very difficult for a captioner to describe details from the image using such short sentences.
Thus, a scene graph can be a good complementing component to the caption.

\section{Conclusion}


Exploring object relationships for image captioning is a challenging task, because it not only requires a strong encoder-decoder model to generate accurate captions but also a scheme to embed the relational information in the encoder. To effectively improve the image caption quality, we propose the use of \name to integrate the extraction of object relationship and caption generation into the same learning framework that the encoder 
can be more accurately trained. 
Our method achieves significant gains over COCO  dataset compared to the state-of-the-art models. However, there is still a room to improve the quality of image captioning. 
For example, to get more accurate scene graphs, one might use the online crowd-sourcing tools like Amazon Mechanical Turk to manually annotate the COCO dataset with relational labels. 
\chapter{Fashion Captioning}

\section{Introduction}
Motivated by the quick global growth of the fashion industry, which is worth trillions of dollars\footnote{https://www.statista.com/topics/965/apparel-market-in-the-us/},
 extensive efforts have been devoted to fashion related research over the last few years. Those research directions include clothing attribute prediction and landmark detection \cite{Wang2018AttentiveFG,liuLQWTcvpr16DeepFashion}, fashion recommendation \cite{Yu2018}, item retrieval \cite{Liu:2012,Wang2017ClothingRW}, clothing parsing \cite{2018arXiv180610787G,He2017RealTimeFC}, and outfit recommendation \cite{han2017learning,Lu_2019_CVPR,Cucurull_2019_CVPR}.

Accurate and enchanting descriptions of clothes on shopping websites can help customers without fashion knowledge to better understand the features (attributes, style, functionality, benefits to buy, etc.) of the items and increase online sales  by enticing more customers~\footnote{https://www.lyfemarketing.com/blog/best-product-description/}.
However, manually writing the descriptions is a non-trivial and highly expensive task. 
Thus, the automatic generation of descriptions is in urgent need.
Since there exist no studies on generating fashion related descriptions, in this chapter, we propose specific schemes on \textit{Fashion Captioning}. Our design is built upon our newly created \textit{FAshion CAptioning Dataset} (FACAD), the first fashion captioning dataset consisting of over $800$K images and $120$K descriptions with massive attributes and categories.
Compared with general image captioning datasets (e.g. MS COCO~\cite{ChenCOCO15}), 
the descriptions of fashion items have three unique features (as can be seen from Fig.~\ref{fig:sample}), which makes the automatic generation of captions a challenging task. 
First, fashion captioning needs to describe the fine-grained attributes of a single item, while image captioning generally narrates the objects and their relations in the image (e.g., a person in a dress). 
Second, the expressions to describe the clothes tend to be long so as to present the rich attributes of fashion items. The average length of captions in FACAD is 23 words while a sentence in the MS COCO caption dataset contains 10.4 words in average.
Third, FACAD has a more enchanting expression style than MS COCO to arouse greater customer interests.
Sentences like ``pearly'', ``so-simple yet so-chic'', ``retro flair'' are more attractive than the plain or ``undecorated'' MS COCO descriptions. 
\begin{figure}
\small
\centering
\includegraphics[width=0.9\textwidth]{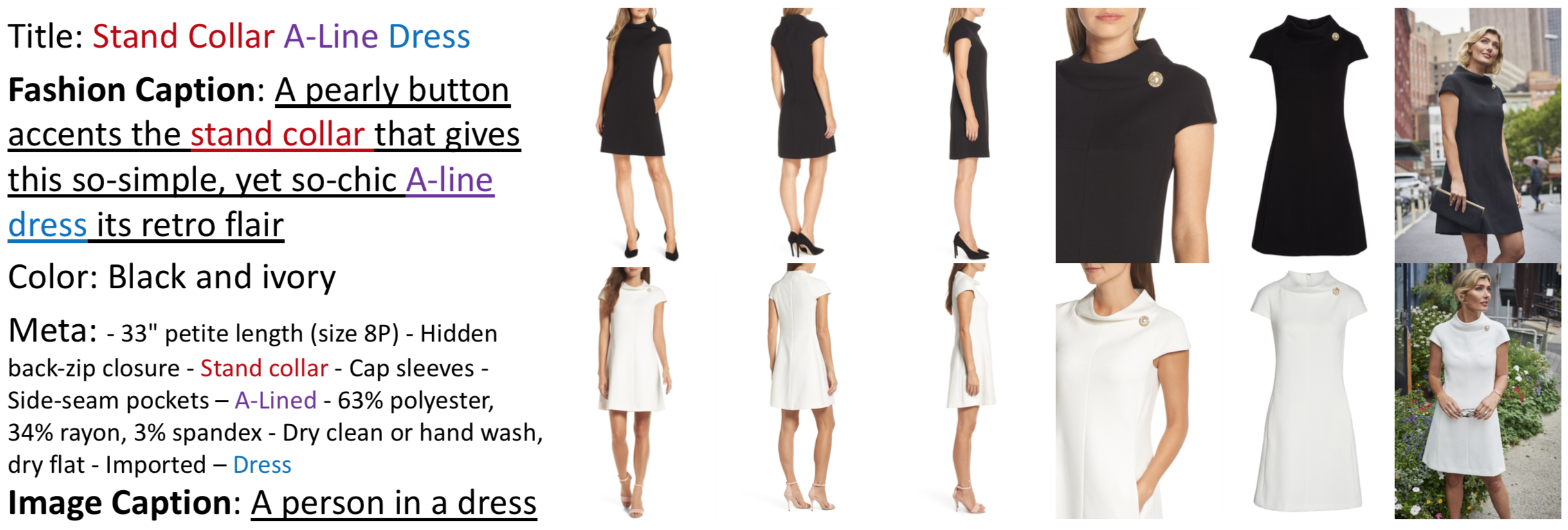}
\caption{An example for Fashion Captioning. The images are of different perspectives, colors and scenarios (shop-street). Other information contained include a title, a description (caption) from a fashion expert, the color info and the meta info. Words in color denotes the attributes used in sentence. Note: Image Caption is generated by an image captioning model trained on MS COCO dataset.
}
\label{fig:sample}
\end{figure}

The image captioning problem has been widely studied and achieved great progress in recent years.
An encoder-decoder paradigm is generally followed with a deep convolutional neural network (CNN)  to encode the input images and a Long Short Term Memory (LSTM) decoder to generate the descriptions~\cite{Kelvin2015,Karpathy2017,Johnson16,Simao2019,Anderson2018}.
The encoder-decoder model is trained via maximum likelihood estimation (MLE), which aims to maximize the likelihood of the next word given the previous words.
However, MLE-based methods will cause the model to generate ``unmatched'' descriptions for the fashion items, where sentences cannot precisely describe the attributes of items. This is due to two reasons.
First, MLE treats the attribute and non-attribute words equally.  Attribute words are not emphasized and directly optimized in the training process, however, they are more important and should be considered as the key parts in the evaluation.
Second, MLE maximizes its objective word-by-word without considering the global semantic meaning of the sentence. This shortcoming may lead to generating a caption that wrongly describes the category of the item. 

To generate better descriptions for fashion items, we propose two semantic rewards as the objective to optimize and train our model using Reinforcement Learning (RL). 
Specifically, we propose an \textit{attribute-level semantic} (ALS) reward with an attribute-matching algorithm to measure the consistency level of attributes between the generated sentences and ground-truth. By incorporating the semantic metric of attributes into our objective, we increase the quality of sentence generation \del{in terms of more accurately describing the attributes of an item}\rev{from the semantic perspective}.
As a second procedure, we propose a \textit{sentence-level semantic} (SLS) reward to capture the semantic meaning of the whole sentence.\del{so that the output description can describe the same category as the ground-truth caption as measured by a pre-trained text classifier.}
\rev{Given a text classifier pretrained on the sentence category classification task, the high level features of the generated description, i.e., the category feature, should stay the same as the ground-truth sentence.
In this chapter, we use the output probability of the generated sentence as the groundtruth category as the SLS reward.}
Since both ALS reward and SLS reward are non-differentiable, we seek RL to optimize them.

In addition, to guarantee that the image features extracted from the CNN encoder are meaningful and correct, we design a visual attribute predictor to make sure that the predicted attributes match the ground-truth ones.
Then attributes extracted are used as the condition in the LSTM decoder to produce the words of description. 

This work has three main contributions. 
\begin{enumerate}
    \item We build a large-scale fashion captioning dataset FACAD of over $800$K images which are comprehensively annotated with categories, attributes and descriptions.  To the best of our knowledge, it is the first fashion captioning dataset available. We expect that this dataset will greatly benefit the research community, in not only developing various fashion related algorithms and applications, but also helping visual language related studies.  
    \item We introduce two novel rewards (ALS and SLS) into the Reinforcement Learning framework to capture the semantics at both the attribute level and the sentence level to largely increase the accuracy of fashion captioning.
    \item We introduce a visual attribute predictor to better capture the attributes of the image.
The generated description seeded on the attribute information can more accurately describe the item.
\end{enumerate}

\section{Related Work}
We provide a review on fashion related studies and image captioning.
\subsubsection{Fashion Studies}
Most of the fashion related studies~\cite{Cucurull_2019_CVPR,Vasileva18FasionCompatibility,Wang2018AttentiveFG,liuLQWTcvpr16DeepFashion,Yu2018,Liu:2012,2018arXiv180610787G} involve images.
For outfit recommendation, Cucurull \textit{et al.}~\cite{Cucurull_2019_CVPR} used a graph convolutional neural network to model the relations between items in a outfit set,  while Vasileva \textit{et al.}~\cite{Vasileva18FasionCompatibility} used a triplet-net to integrate the type information into the recommendation.
Wang \textit{et al.}~\cite{Wang2018AttentiveFG} used an attentive fashion grammar network for landmark detection and clothing category classification.
Yu \textit{et al.}~\cite{Yu2018} introduced the aesthetic information, which is highly relevant with user preference, into clothing recommending systems. Text information has also been exploited. Han \textit{et al.}~\cite{han2017learning} used title features to regularize the image features learned.
Similar techniques were used in ~\cite{Vasileva18FasionCompatibility}.
But no previous studies focus on fashion captioning.
\subsubsection{Image Captioning}
Image captioning helps machine understand visual information and express it in natural language, and has attracted increasingly interests in computer vision. State-of-the-art approaches~\cite{Kelvin2015}\cite{Johnson16}\cite{Simao2019}\cite{Anderson2018} mainly use encoder-decoder frameworks with attention to generate captions for images.
Xu \textit{et al.}~\cite{Kelvin2015} developed soft and hard attention mechanisms to focus on different regions in the image when generating different words.
Johnson \textit{et al.}~\cite{Johnson16} proposed a fully convolutional localization network to generate dense regions of interest and use the generated regions to generate captions. 
Similarly, Anderson \textit{et al.}~\cite{Anderson2018} and Ma \textit{et al.}~\cite{ma2017} used an object detector like Faster R-CNN~\cite{Ren15} or Mask R-CNN~\cite{He2017MaskR} to extract regions of interests over which an attention mechanism is defined. 
Regardless of the methods used, image captioning generally describes the contents based on the relative positions and relations of objects in an image. 
Fashion Captioning, however, needs to describe the implicit attributes of the item which cannot be easily localized by object detectors.

Recently, policy-gradient methods for Reinforcement Learning (RL) have been utilized to train deep end-to-end systems directly on non-differentiable metrics~\cite{Williams92}. 
Commonly the output of the inference is applied to normalize the rewards of RL. Ren \textit{et al.}~\cite{Zhou2017} introduced a decision-making framework utilizing a \textit{policy network} and a \textit{value network} to collaboratively generate captions with reward driven by visual-semantic embedding.
Rennie \textit{et al.}~\cite{Rennie2017} used self-critical sequence training for image captioning.
The reward is provided using CIDEr~\cite{Vedantam15cider} metric.
Gao \textit{et al.}~\cite{Gao_2019_CVPR} extended \cite{Rennie2017} by running a $n$-step self-critical training.
The specific metrics used in RL approach are hard to generalize to other applications, and optimizing specific metrics often impact other metrics severely. However, the semantic rewards we introduce are general and effective in improving the quality of caption generation.

\section{The FAshion CAptioning Dataset}
We introduce a new dataset - FAshion CAptioning Dataset (FACAD) - to study captioning for fashion items. 
In this section, we will describe how FACAD is built and what are its special properties.
\subsection{Data Collection, Labeling and Pre-Processing}
\label{sec:data}
We mainly scrawl data from \textit{Nordstrom}\footnote{https://shop.nordstrom.com/} online shopping website as it provides detailed information, which can be exploited for the fashion captioning task.
Each clothing item has on average $6\sim7$ images of various colors and poses.
The resolution of the images is $1560\times 2392$, much higher than other fashion datasets.

\begin{figure}
\centering
\begin{subfigure}{0.39\textwidth}
\centering
  \includegraphics[width=\textwidth]{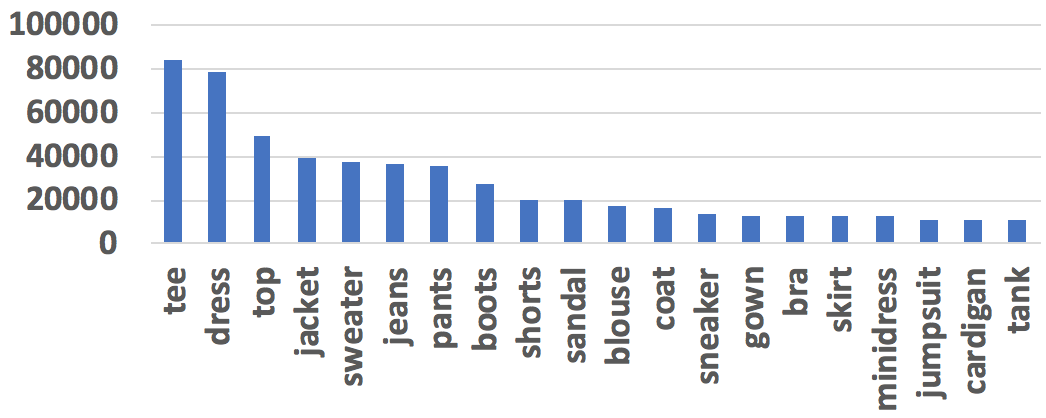}
  \caption{}
  \label{fig:cate} 
\end{subfigure}
\begin{subfigure}{0.60\textwidth}
\centering
  \includegraphics[width=\textwidth]{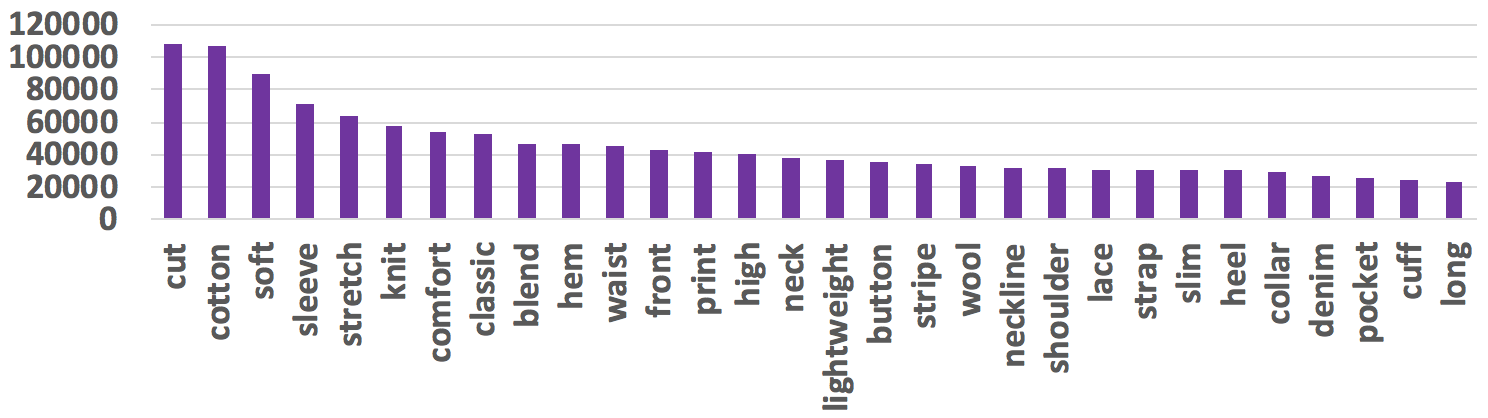}
  \caption{}
  \label{fig:attr}
\end{subfigure}
\caption[]{(a) Number of items in the top-20 categories. (b) Number of items in the top-30 attributes.}
\label{fig:stat}
\end{figure}

In order to better understand fashion items, we label them with rich categories and attributes.
An example category of clothes can be ``dress''  or ``T-shirt'', while an attribute such as ``pink'' or ``lace'' provides some detailed information about a specific item.
The list of the categories is generated by picking the last word of the item titles. 
After manual selection and filtering, there are 273 total valuable categories left. 
We then merge similar categories and only keep ones that contain over 500 items, resulting in 74 unique categories.
Each item belongs to only one category.
The number of items contained by the top-20 categories are shown in Fig.~\ref{fig:cate}.

Since there are a large number of attributes and each image can have several attributes, manual labeling is non-trivial. 
We utilize the title, description and meta data to help label attributes for the items. 
Specifically, we first extract the nouns and adjectives in the title using Stanford Parser~\cite{socher2013parsing}, and then select a noun or adjective as the attribute word if it also appears in the caption and meta data. 
The total number of attributes we extracted is over 3000 and we only keep those that appear in more than 50 items, resulting in a list of 1098 attributes.
Each item owns approximately $7.3$ attributes.
We show the number of items that are associated with the top-30 attributes in Fig.\ref{fig:attr}. 

To have clean captions, we tokenize the descriptions using NLTK tokenizer\footnote{https://www.nltk.org/api/nltk.tokenize.html} and remove the non-alphanumeric words. We lowercase all caption words.

\subsection{Comparison with other datasets}
The statistics of our FACAD is shown in Table~\ref{tab:datasets}. 
Compared with other fashion datasets such as~\cite{Liu16DeepFashion,DeepFashion2,Zheng18,Zou19,guo2019fashion}, FACAD has two outstanding properties. 
First, it is one of the biggest fashion datasets, with over $800$K diverse fashion images of all four seasons, ages (kids and adults), categories (clothing, shoes, bag, accessories, etc.), angles of human body (front, back, side, etc.).
Second, it is the first dataset to tackle captioning problem for fashion items. 
120K descriptions with average length of 23 words was pre-processed for later researches. 

Compared with MS COCO~\cite{ChenCOCO15} image captioning dataset, FACAD is different in three aspects. 
First, FACAD contains the fine-grained descriptions of attributes of fashion-related items, while MS COCO  narrates the objects and their relations in general images. 
Second, FACAD has longer captions (23 words per sentence on average) compared with 10.4 words per sentence of the MS COCO caption dataset, imposing more difficulty for text generation. 
Third, the expression style of FACAD is enchanting, while that of MS COCO is plain without rich expressions. 
As illustrated in Fig.~\ref{fig:sample}, words like “pearly”, “so-simple yet so-chic”, “retro flair” are more attractive than the plain MS COCO descriptions, like ``a person in a dress''. 
This special enchanting style is important in better describing an item and attracting more customers, but also imposes another challenge for building the caption models.

\begin{table}[!t]
\captionsetup{font=small}
\caption{Comparison of different datasets. CAT: category, AT: attribute, CAP: caption, FC: fashion captioning, IC: image captioning, CLS: fashion classification, SEG: segmentation, RET: retrieval.}
\small
\centering
\begin{tabular}{cccccccccl}
\toprule
Datasets & \# img & \# CAT  & \# AT & \# CAP & avg len & style & task \\
\midrule
\textbf{FACAD} & \textbf{800K} & \textbf{74} & \textbf{1098} & \textbf{120K} & \textbf{23}  & \textbf{enchanting} & \textbf{FC} \\
\hline
MS COCO~\cite{ChenCOCO15} & 123K & -- & -- & 616K & 10.4 & plain & IC  \\
VG~\cite{Krishna17} & 108K & -- & -- & 5040K & 5.7 & plain & IC \\
\hline
DFashion~\cite{Liu16DeepFashion}\cite{DeepFashion2} & 800K & 50 & 1000 & -- & -- & -- & CLS \\
Moda~\cite{Zheng18} & 55K & 13 & -- & -- & -- & -- & SEG \\
Fashion AI~\cite{Zou19} & 357K & 6 & 41 & -- & -- & -- & CLS  \\
Fashion IQ~\cite{guo2019fashion} & 77K & 3 & 1000 & -- & -- & -- & RET \\
\bottomrule
\end{tabular}
\label{tab:datasets}
\end{table}

\section{Respecting Semantics for Fashion Captioning}
In this section, we first formulate the basic fashion captioning problem and its general solution using Maximum Likelihood Estimation (MLE).
We then propose a set of strategies to increase the accuracy of fashion captions: 1)  learning specific fashion attributes from the image; 2) establishing attribute-level and sentence-level semantic rewards so that the caption can be generated to be more similar to the ground truth through Reinforcement Learning (RL); 3) alternative training with MLE and RL to optimize the model.

\subsection{Basic Problem Formulation}
We define a dataset of image-sentence pairs as $\mathcal{D}=\{ (X,Y) \}$.
Given an item image $X$, the objective of Fashion Captioning is to generate a description $Y=\{ y_1,\ldots ,y_T \}$ with a sequence of $T$ words, $y_i \in V^K$ being the $i$-th word, $V^K$ being the vocabulary of $K$ words. The beginning of each sentence is marked with a special $<$BOS$>$ token, and the end with an $<$EOS$>$ token.
We denote $\mathbf{y}_i$ as the embedding for word $y_i$. 
To generate a caption, the objective of our model is to minimize the negative log-likelihood of the correct caption using maximum likelihood estimation (MLE): 
\begin{equation}
    \mathcal{L}_{MLE} = -\sum_{t=1}^{T}\log p(y_t\vert y_{1:t-1}, X)
\label{eq:loglikelihood}.
\end{equation}

As shown in Fig.~\ref{fig:model}, we use an encoder-decoder architecture to achieve this objective. The encoder is a pre-trained CNN, which takes an image as the input and extracts $B$ image features, $\mathbf{X}=\{ \mathbf{x}_1,\ldots , \mathbf{x}_B \}$.
We dynamically re-weight the input image features $\mathbf{X}$ with an attention matrix $\mathbf{\gamma}$ to focus on specific regions of the image at each time step $t$~\cite{Kelvin2015}, which results in a weighted image feature $\mathbf{x}_t=\sum_{i=1}^B\gamma_t^i\mathbf{x}_i$.
The weighted image feature is then fed into a decoder which is a Long Short-Term Memory (LSTM) network for sentence generation. 
The decoder predicts one word at a time and controls the fluency of the generated sentence. 
More specifically, when predicting the word at the $t$-th step, the decoder takes as input the embedding of the generated word ${y}_{t-1}$, the weighted image feature $\mathbf{x}_t$ and the previous hidden state $\mathbf{h}_{t-1}$. 
The initial memory state and hidden state of the LSTM are initialized by an average of the image features fed through two feed-forward networks $f_c$ and $f_h$ which are trained together with the whole model: $\mathbf{c}_0=f_c(\frac{1}{B}\sum_{i=1}^B\mathbf{x}_i)$, $\mathbf{h}_0=f_h(\frac{1}{B}\sum_{i=1}^B\mathbf{x}_i)$. 
The decoder then outputs a hidden state $\mathbf{h}_{t}$ (Eq.~\ref{eq:hid}) and applies a linear layer $f$ and a \textit{softmax} layer to get the probability of the next word (Eq.~\ref{eq:y}):
\begin{equation}
    \mathbf{h}_{t} = LSTM([\mathbf{y}_{t-1}; \mathbf{x}_t], \mathbf{h}_{t-1})
    \label{eq:hid}
\end{equation}
\begin{equation}
    p_{\theta}(y_t\vert y_{1:t-1}, \mathbf{x}_t)=softmax(f(\mathbf{h}_{t}))
    \label{eq:y}
\end{equation}
where $[;]$ denotes vector concatenation.

\begin{figure}
\centering
\includegraphics[width=0.9\textwidth]{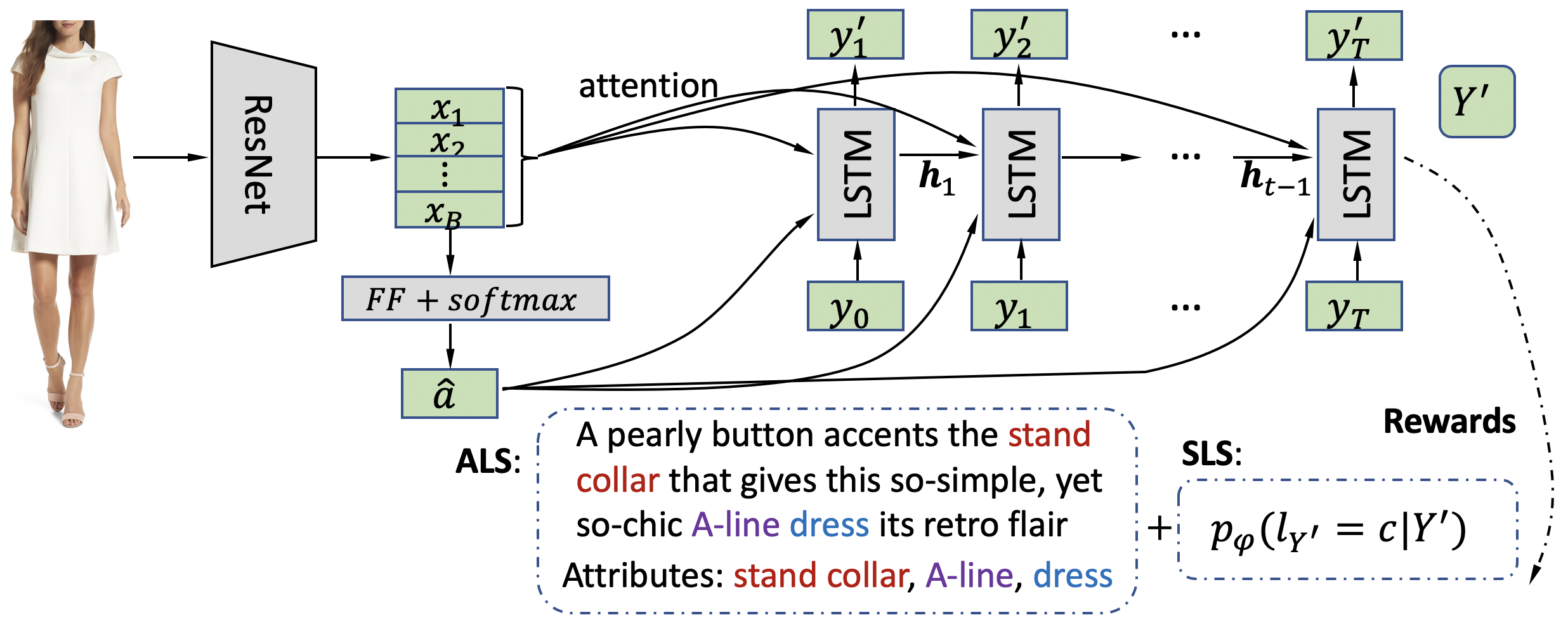}
\caption{The proposed model architecture and rewards.}
\label{fig:model}
\end{figure}

\subsection{Attribute Embedding}
To make sure that the caption correctly describes the item attributes, we introduce an attribute feature $\mathbf{z}$ into the model, which modifies Eq.~\ref{eq:loglikelihood} into:
\begin{equation}
    \mathcal{L}_{MLE} = -\sum_{t=1}^{T}\log p(y_t\vert y_{1:t-1}, \mathbf{z}, X)
\label{eq:loglikelihood_z}.
\end{equation}
This objective aims at seeding sentence generation with the attribute feature of the image. To regularize the encoder to output attribute-correct features, we add a visual attribute predictor to the encoder-decoder model.
As each item in the FACAD has its attributes shown in the captions, the predictor can be trained by solving the problem of multi-label classification. 
The trained model can be applied to extract the attributes of an image to produce the caption. 

Fig.~\ref{fig:model} illustrates the attribute prediction network. 
We attach a feed-forward (FF) network to the CNN feature extractor, and its output is fed into an $A$-way softmax to produce a probability vector of size $A$, with $A$ being the total number of attributes possible.
For a given training sample, let $\mathbf{a}=[ a_1,\ldots , a_A ]$ be its attribute label vector, where $a_j=1$ if the image is annotated with $j$-th attribute, and $a_j=0$ otherwise. Let $\hat{\mathbf{a}}=[ \hat{a}_1,\ldots , \hat{a}_A ]$ be the predicted probability vector from the model. Then the element-wise logistic loss function of the attribute predictor can be defined as:
\begin{equation}
    \mathcal{L}_{a}=\sum_{j=1}^A\log (1+exp(-a_j\hat{a}_j))
    \label{eq:loss_a}
\end{equation}
We can then modify Eq.~\ref{eq:hid} and Eq.~\ref{eq:y} to include the attribute embedding as:
\begin{equation}
    \mathbf{h}_{t} = LSTM([\mathbf{y}_{t-1}; \mathbf{x}_t; \mathbf{z}],  \mathbf{h}_{t-1})
    \label{eq:hid_z}
\end{equation}
\begin{equation}
    p_{\theta}(y_t\vert y_{1:t-1}, \mathbf{x}_t, \mathbf{z})=softmax(f_h(\mathbf{h}_{t}))
    \label{eq:y_z}
\end{equation}
where $\mathbf{z}=\mathbf{W}_a\hat{\mathbf{a}}$ and $\mathbf{W}_a$ is a trainable weight matrix for attributes, $[;]$ denotes vector concatenation.

\subsection{Increasing the Accuracy of Captioning with Semantic Rewards}
Simply training with MLE can force the model to generate most likely words in the vocabulary, but not help decode the attributes that are crucial to the fashion captioning.
To solve this issue, we propose to exploit two semantic metrics to increase the accuracy of fashion captioning:
an attribute-level semantic reward to encourage our model to generate a sentence with more attributes in the image, and a sentence-level semantic reward to encourage the generated sentence to more accurately describe the category of a fashion item.
Because optimizing the two rewards is a non-differentiable process, during the MLE training, we supplement fashion captioning with a Reinforcement Learning (RL) process. 

In the RL process, our encoder-decoder network with attribute predictor can be viewed as an \textit{agent} that interacts with an external \textit{environment} (words and image features) and takes the \textit{action} to predict the next word. 
After each action, the agent updates its internal \textit{state} (cells and hidden states of the LSTM, attention weights, etc).
Upon generating the \textit{end-of-sequence} ($<$EOS$>$) token, the agent observes a \textit{reward} $r$ as a judgement of how good the overall decision is. We have designed two levels of rewards, as defined below:
\subsubsection{Attribute-Level Semantic (ALS) Reward}
We propose the use of {\em attribute-level semantic} (ALS) reward to encourage our model to locally generate as many correct attributes as possible in a caption.
First, we need to represent an attribute with a phrase. We denote a contiguous sequence of $n$ words as an $n$-gram, and we only consider $n=1, 2$ since nearly all the attributes contain 1 or 2 words. 
We call an $n$-gram that contains a correct attribute a tuple $t_n$. That is, a tuple $t_n$ in the generated sentence contains the attribute in the groundtruth sentence and results in an attribute``Match''.
We define the proportion of ``Matching'' for attributes of $n$ words in a generated sentence as:  
\begin{equation}
    P(n)=\frac{Match(n)}{H(n)},
\end{equation}
where $H(n)$ is the total number of $n$-grams contained by a sentence generated. An $n$-gram may or may not contain an attribute. 
For a generated sentence with $M$ words, $H(n)=M+1-n$. The total number of ``Matches'' is defined as:
\begin{equation}
    Match(n)=\sum_{t_n}\min(C_g(t_n), C_r(t_n))
\end{equation}
where $C_g(t_n)$ is the number of times a tuple $t_n$ occurs in the generated sentence, and 
$C_r(t_n)$ is the number of times the same tuple $t_n$ occurs in the groundtruth caption. We use $\min()$ to make sure that the generated sentence does not contain more repeated attributes than the groundtruth.
We then define the ALS reward as:
\begin{equation}
    r_{ALS}=\beta\{ \prod_{n=1}^2P(n) \}^{\frac{1}{n}}
\end{equation}
where $\beta$ is used to penalize short sentences which is defined as:
\begin{equation}
    \beta = \exp\{ \min(0, \frac{l-L}{l}) \}
\end{equation}
where $L$ is the length of the groundtruth and $l$ is the length of the generated sentence. 
When the generated sentence is much shorter than the groundtruth, although the model can decode the correct attributes with a high reward, the sentence may not be expressive with an enchanting style. We thus leverage a penalization factor to discourage this.

\subsubsection{Sentence-Level Semantic (SLS) Reward}
The use of attribute-level semantic score can help generate a sentence with more correct attributes, which thus increases the similarity of the generated sentence with the groundtruth one at the local level. To further increase the similarity between the generated sentence and groundtruth caption at the global level, we consider enforcing a generated sentence to describe an item with the correct category. This design principle is derived based on our observation that items of the same category share many attributes, while those of different categories often have totally different sets of attributes. Thus, a sentence generally contains more correct attributes if it can describe an item with a correct category. 

To achieve the goal, we pretrain a text category classifier $p_{\phi}$, which is a $3$-layer text CNN, using captions as data and their categories  as labels
($\phi$ denotes the parameters of the classifier).
Taking the generated sentence $Y^{\prime}=\{ y_1^{\prime},\ldots , y_T^{\prime} \}$ as inputs, the text category classifier will output a probability distribution $p_{\phi}(l_{Y^{\prime}}\vert Y^{\prime})$, where $l_{Y^{\prime}}$ is the category label for $Y^{\prime}$. 
The sentence-level semantic reward is defined as:
\begin{equation}
    r_{SLS} = p_{\phi}(l_{Y^{\prime}}=c\vert Y^{\prime})
\end{equation}
where $c$ is the target category of the sentence.

\subsubsection{Overall Semantic Rewards}
To encourage our model to improve both the ALS reward and the SLS reward, we use an overall semantic reward which is a weighted sum of the two:
\begin{equation}
    r = \alpha_1r_{ALS} + \alpha_2r_{SLS}
\end{equation}
where $\alpha_1$ and $\alpha_1$ are two hyper-parameters.

\subsubsection{Computing Gradient with REINFORCE}
The goal of RL training is to minimize the negative expected reward:
\begin{equation}
    \mathcal{L}_{r}=-\mathbb{E}_{Y^{\prime}\sim p_{\theta}}[r(Y^{\prime})]
\end{equation}

To compute the gradient $\nabla_{\theta}\mathcal{L}_{r}(\theta)$, we use the REINFORCE algorithm~\cite{Williams92} to calculate the expected gradient of a non-differentiable reward function:
\begin{equation}
\begin{aligned}
    \nabla_{\theta}\mathcal{L}_{r}(\theta)
    =-\mathbb{E}_{Y^{\prime} \sim p_{\theta}}[r(Y^{\prime})\nabla_{\theta}\log p_{\theta}(Y^{\prime})]
\end{aligned}
\end{equation}

To reduce the variance of the expected rewards, the gradient can be generalized by incorporating a \textit{baseline} $b$:
\begin{equation}
    \nabla_{\theta}\mathcal{L}_{r}(\theta)=-\mathbb{E}_{Y^{\prime} \sim p_{\theta}}[(r(Y^{\prime})-b)\nabla_{\theta}\log p_{\theta}(Y^{\prime})]
\end{equation}

In our experiments, the expected gradient is approximated using $H$ samples from $p_{\theta}$ and the baseline is the average reward of all the $H$ sampled sentences:
\begin{equation}
\begin{aligned}
    \nabla_{\theta}\mathcal{L}_{r}(\theta) \simeq -\frac{1}{H}\sum_{j=1}^H[(r_j(Y_j^{\prime})-b)\nabla_{\theta}\log p_{\theta}(Y_j^{\prime})]
\end{aligned}
\end{equation}
where $b=\frac{1}{H}\sum_{j=1}^Hr(Y^{\prime}_j)$, 
$Y^{\prime}_j\sim p_{\theta}$ is the $j$-th sampled sentence from model $p_{\theta}$ and $r_j(Y_j^{\prime})$ is its corresponding reward.

\subsection{Joint Training of MLE and RL}
In practice, rather than starting RL training from a random policy model, we warm-up our model using MLE and attribute embedding objective till converge.
We then integrate the pre-trained MLE, attribute embedding, and RL into one model to retrain until it converges again, following the  overall loss function:
\begin{equation}
    \mathcal{L} = \mathcal{L}_{MLE} + \lambda_1\mathcal{L}_r + \lambda_2\mathcal{L}_a
\end{equation}
with $\lambda_1$ and $\lambda_2$ being two hyper-parameters.
\section{Experiments}
We evaluate the performance of our proposed model through extensive experimental studies. As we are not aware of any existing studies on fashion captioning, we make comparisons with a number of literature methods originally designed for general image captioning, which are applied to describe fashion items.

\subsection{Basic Setting}

\noindent\textbf{Dataset and Metrics}
We run all methods over FACAD. It contains 800K images and 120K descriptions, 
and we split the whole dataset, with approximately 640K image-description pairs for training, 80K for validation, and the remaining 80K for test. 
Images for the same item share the same description.
The number of images associated with one item varies, ranging from 2 to 12. 
As several images in FACAD (e.g., clothes shown in different angles)  share the same description, instead of randomly splitting the dataset, we ensure that the  images with the same caption are contained in the same data split.
We convert all sentences to lowercase and discard non-alphanumeric characters. For words in the training set, we only keep the ones that appear at least 5 times, resulting in a vocabulary of 14163 words.

For fair and thorough performance measure, we report results under the commonly used metrics for image captioning, including BLEU~\cite{papineni2002}, METEOR~\cite{denkowski2014meteor}, ROUGEL~\cite{lin2004rouge}, CIDEr~\cite{Vedantam15cider}, SPICE~\cite{spice2016}.
In addition, we compare the attributes in the generated captions with those in the test set as ground truth to find the average precision rate for each attribute using mean average precision (mAP). To evaluate whether the generated captions belong to the correct category, we report the category prediction accuracy (ACC).
We pre-train a $3$-layer text CNN~\cite{kim2014} as the category classifier $p_{\phi}$, achieving a classification accuracy of $90\%$ on test set.

\noindent\textbf{Network Architecture}
As shown in Fig.~\ref{fig:model}, we use a ResNet-101~\cite{He2015DeepRL}, pretrained on ImageNet to encode each image feature. 
Since there is a large domain shift from ImageNet to FACAD, we fine tune the conv4\_x and the conv5\_x layers to get better image features. The features output from the final convolutional layer are used to further train over FACAD.
We use LSTM~\cite{hochreiter1997long} as our decoder. 
The input node dimension and the hidden state dimension of LSTM are both set to $512$. 
The word embeddings of size $512$ are uniformly initialized within $[-0.1, 0.1]$.
After testing with several combinations of the hyper-parameters, we set the $\alpha_1=\alpha_2=1$ to assign equal weights to both rewards, and $\lambda_1=\lambda_2=1$ to balance MLE, attribute prediction and RL objectives during training.

\noindent\textbf{Training Details}
All the models are trained according to the following procedure, unless otherwise specified.
We initialize all models by training using MLE objective with cross entropy loss with ADAM~\cite{kingma2015} optimizer at an initial learning rate of $1\times 10^{-4}$.
We anneal the learning rate by a factor of 0.9 every two epochs.
After the model training converges on the MLE objective, if RL training is further needed in a method, we switch to MLE + RL training till another converge. The overall process takes about 4 days on two NVIDIA 1080 Ti GPUs.

\noindent\textbf{Baseline Methods}
To make fair comparisons, we take image captioning models based both on MLE training and training with MLE$+$RL. For all the baselines, we use their published codes to run the model, performing a hyperparameter search based on the original author's guidelines. We follow their own training schemes to train the models.

\textit{MLE-based Methods.}
\textbf{CNN-C}\cite{Aneja18} is a CNN-based image captioning model which uses a masked convolutional decoder for sentence generation. \textbf{SAT}~\cite{Kelvin2015} applies CNN-LSTM with attention, and we use its hard attention method. \textbf{BUTD}~\cite{Anderson2018} combines the bottom-up and the top-down attention, with the bottom-up part containing a set of salient image regions, each is represented by a pooled convolutional feature vector. \textbf{LBPF}~\cite{Qin2019CVPR} uses a look back (LB) approach to introduce attention value from the previous time step into the current attention generation and a predict forward (PF) approach to predict the next two words in one time step. \textbf{TRANS}~\cite{Simao2019} proposes the use of geometric attention for image objects based on Transformer~\cite{Vaswani2017}.

\textit{MLE + RL based Methods.}
\textbf{AC}~\cite{Zhang2017ac} uses actor-critic Reinforcement Learning algorithm to directly optimize on CIDEr metric. \textbf{Embed-RL}~\cite{Zhou2017} utilizes a ``policy'' and a ``value'' network to jointly determine the next best word. The policy network serves as a local guidance while the value network serves as a global guidance. \textbf{SCST}~\cite{Rennie2017} is a self-critical sequence training algorithm. \textbf{SCNST}~\cite{Gao_2019_CVPR} is a $n$-step self-critical training algorithm extended from~\cite{Rennie2017}. We use 1-2-2-step-maxpro variant which achieved best performance in the work.

\subsection{Performance Evaluations} 
\noindent\textbf{Results on Fashion Captioning}
Our Semantic Rewards guided Fashion Captioning (SRFC) model achieves the highest scores on all seven metrics.
Specifically, it provides $2.7$, $2.4$, $4.5$, $9.4$, $2.2$, $0.074$ and $0.062$ points of improvement over the best baseline SCNST on BLEU4, METEOR, ROUGEL, CIDEr, SPICE, mAP and ACC respectively, demonstrating the effectiveness of our proposed model in providing fashion captions. 
The improvement mainly comes from 3 parts, attribute embedding training, ALS reward and SLS reward. 
To evaluate how much contribution each part provides to the final results, we remove different components from SRFC and see how the performance degrades.
For SRFC without attribute embedding, our model experiences the performance drops of $0.8$, $0.6$, $1.0$, $3.0$, $0.3$, $0.011$ and $0.021$ points.
After removing ALS, the performance of SRFC drops $1.3$, $0.8$, $1.5$, $4.6$ and $0.6$ points on the first five metrics.
For the same five metrics, the removing of SLS results in higher performance degradation, which indicates that the global semantic reward plays a more important role in ensuring accurate description generation. 
More interestingly, removing ALS produces a larger drop in mAP, while removing SLS impacts more on ACC. This means that ALS focuses more on producing correct attributes locally, while SLS helps ensure the global semantic accuracy of the generated sentence.
Removing both ALS and SLS leads to a large decrease of the performance on all metrics, which suggests that most of the improvement is gained by the proposed two semantic rewards.
Finally, with the removal of all three components, the performance of our model is similar to that of the baselines without using any proposed techniques. This demonstrates that all three components are necessary to have a good performance on fashion captioning.

\begin{table*}[!t]
\centering
\small
\caption{\small \textbf{Fashion captioning results -} scores of different baseline models as well as different variants of our proposed method. 
\textbf{A}: attribute embedding learning. 
We highlight the \textbf{best} model in bold.}
\begin{tabular}{ccccccccl}
\toprule
 \textbf{Model} & \textbf{BLEU4} & \textbf{METEOR} & \textbf{ROUGEL} & \textbf{CIDEr} & \textbf{SPICE} & \textbf{mAP} & \textbf{ACC} \\
\midrule
CNN-C~\cite{Aneja18}  & 20.7 & 20.3 & 39.8 & 99.5 & 18.9 & 0.153 & 0.450  \\
SAT~\cite{Kelvin2015}  & 21.1 & 20.5 & 40.6 & 100.4 & 19.0 & 0.164 & 0.453  \\
BUTD~\cite{Anderson2018}  & 21.9 & 21.7 & 41.7 & 102.1 & 19.7 & 0.182 & 0.459  \\
LBPF~\cite{Qin2019CVPR}  & 24.2 & 23.3 & 45.2 & 107.3 & 22.6 & 0.193 & 0.491  \\
TRANS~\cite{Simao2019}  & 23.2 & 22.8 & 44.3 & 106.5 & 21.8 & 0.187 & 0.475  \\
\midrule
AC~\cite{Zhang2017ac}  & 23.5 & 22.1 & 44.8 & 108.1 & 21.9 & 0.186 & 0.463  \\
Embed-RL~\cite{Zhou2017}  & 22.9 & 22.4 & 44.1 & 106.7 & 21.0 & 0.190 & 0.479  \\
SCST~\cite{Rennie2017}  & 24.0 & 23.2 & 44.9 & 108.2 & 22.5 & 0.204 & 0.487  \\
SCNST~\cite{Gao_2019_CVPR} & 24.5 & 23.8 & 45.7 & 109.4 & 22.7 & 0.206 & 0.490  \\
\midrule
SRFC  & \textbf{27.2} & \textbf{26.2} & \textbf{50.2} & \textbf{118.8} & \textbf{24.9} & \textbf{0.280} & \textbf{0.552}  \\
SRFC$-$A  & 26.4 & 25.6 & 49.2 & 115.8 & 24.6 & 0.269 & 0.531   \\
SRFC$-$ALS  & 25.9 & 25.4 & 48.7 & 114.2 & 24.3 & 0.233 & 0.527   \\
SRFC$-$SLS  & 25.6 & 25.2 & 48.3 & 113.7 & 24.1 & 0.264 & 0.503   \\
SRFC$-$ALS$-$SLS   & 23.2 & 22.9 & 44.5 & 106.1 & 21.1 & 0.208 & 0.488 \\
SRFC$-$A$-$ALS$-$SLS  & 21.9 & 21.7 & 41.2 & 101.5 & 20.1 & 0.166 & 0.454  \\
\bottomrule
\end{tabular}
\label{tab:quantitative}
\end{table*}

\noindent\textbf{Results with Subjective Evaluation}
As fashion captioning is used for online shopping systems, attracting customers is a very important goal.
Automatically evaluating the ability to attract customers is infeasible.
Thus, we perform human evaluation on the attraction of generated captions from different models.
5 human judges of different genders and age groups are presented with 200 samples each. 
Each sample contains an image, 10 generated captions from all 10 models, with the sequence randomly shuffled. 
Then they are asked to choose the most attractive caption for each sample.
The results in Table~\ref{tab:human} show that our model produces the most attractive captioning.

\begin{table*}[!t]
\centering
\small
\caption{\small \textbf{Human evaluation on captioning attraction.} We highlight the \textbf{best} model in bold.}
\begin{tabular}{ccccccccccl}
\toprule
 \textbf{Model} & CNN-C & SAT & BUTD & LBPF & TRANS & AC & Embed-RL & SCST & SCNST & SRFC \\
\midrule
\% best & 7.7 & 7.9 & 8.1 & 10.0 & 8.8 & 8.4 & 8.5 & 10.2 & 10.7 & \textbf{19.7} \\
\bottomrule
\end{tabular}
\label{tab:human}
\end{table*}

\noindent\textbf{Qualitative Results and Analysis} Fig.~\ref{fig:qualitative} shows two qualitative results of our model against SCNST and ground truth. In general, our model can generate more reasonable descriptions compared with SCNST for the target image in the middle column. In the first example, we can see that our model generates a description with more details than SCNST, which only correctly predicted the category and some attributes of the target item.

By providing two other items of the same category and their corresponding captions, we have two interesting observations.
First, our model generates descriptions in two steps, it starts learning valuable expressions from similar items (in the same category) based on attributes extracted, and then applies these expressions to describe the target one.
Taking the first item (top row of Fig. \ref{fig:qualitative}) as an example, our model first gets the correct attributes of the image, i.e., \textit{italian sport coat}, \textit{wool}, \textit{silk}.
Then it tries to complete a diverse description by learning from the captions of those items with similar attributes. Specifically, it uses \textit{a richly textured blend} and \textit{handsome} from the first item (left column) and \textit{framed with smart notched lapel} (right column) from the second item to make a new description for the target image. 
The second observation is that our model can enrich description generation by focusing on the attributes identified even if they are not presented in the groundtrue caption.
Even though the \textit{notched lapel} is not described by the ground-truth caption, our model correctly discovers this attribute and generates \textit{framed with smart notched lapel} for it. 
This is because that \textit{notched lapel} is a frequently referred attribute for items of the category \textit{coat}, and this attribute appears in $11.4\%$ descriptions.
Similar phenomena can be found for the second result.

The capability of extracting the correct attributes owes to the \textit{Attribute Embedding Learning} and \textit{ALS} modules.
The \textit{SLS} can help our model generate diverse captions by referring to those from other items with the same category and similar attributes.

\begin{figure}
\captionsetup{font=footnotesize}
\centering
\includegraphics[width=0.95\textwidth]{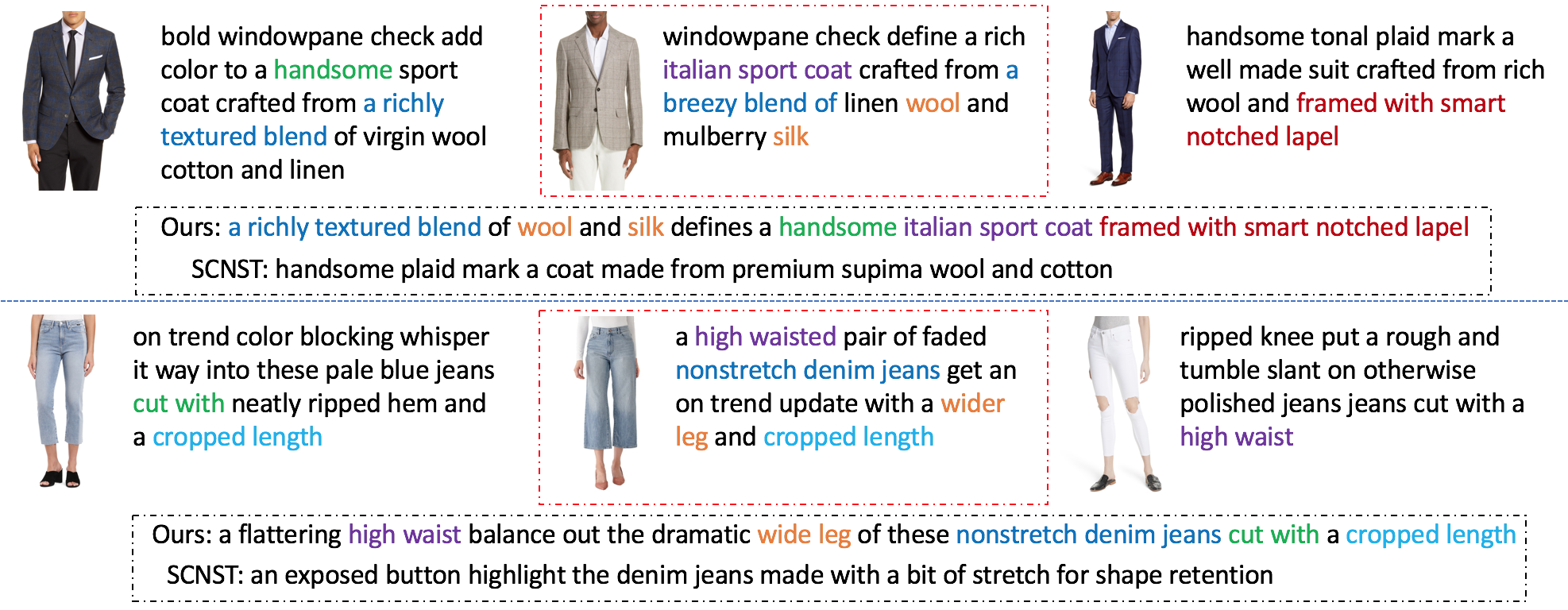}
\caption{\small Two qualitative results of SRFC compared with the groundtruth and SCNST. 
Two target items and their corresponding groundtruth are shown in the red dash-dotted boxes in the middle column. 
The black dash-dotted boxes contain the captions generated by our model and SCNST.
Our model diversely learns different expressions from the other items (on the first and third columns) to describe the target item.
}
\label{fig:qualitative}
\end{figure}

\section{Conclusion}
Generating accurate descriptions for fashion items is of vital importance for both customers and online shopping.
In this work, we propose a novel learning framework for \textit{fashion captioning} and create the first fashion captioning dataset FACAD.
In light of describing fashion items in a correct and expressive manner, we define two novel metrics ALS and SLS, based on which we concurrently train our model with  MLE, attribute embedding and RL training. Our performance results demonstrate that our design can achieve significant gain over other image captioning models when run on FACAD.
Our proposed model can automatically learn different expressions of items from the same category to make a new description.
Since this is the first work on fashion captioning, we apply the evaluation metrics commonly used in the general image captioning.
Further research is needed to develop better evaluation metrics.
\chapter{News Image Captioning}

\section{Introduction}
Research on generating textual descriptions of images has made great progress in recent years with the introduction of encoder-decoder architectures~\cite{Kelvin2015,Johnson16,Venugopalan2017,Karpathy2017,Anderson2018,Lu2018NeuralBT,Aneja18}.
Those models are generally trained and evaluated on image captioning datasets like COCO~\cite{Lin14,ChenCOCO15} and Flickr~\cite{Hodosh13} that only contain generic object categories but no details such as names, locations, or dates.    
The captions generated by these methods are thus generic descriptions of the images.

\begin{figure}[t]
    \centering
    \includegraphics[width=\columnwidth]{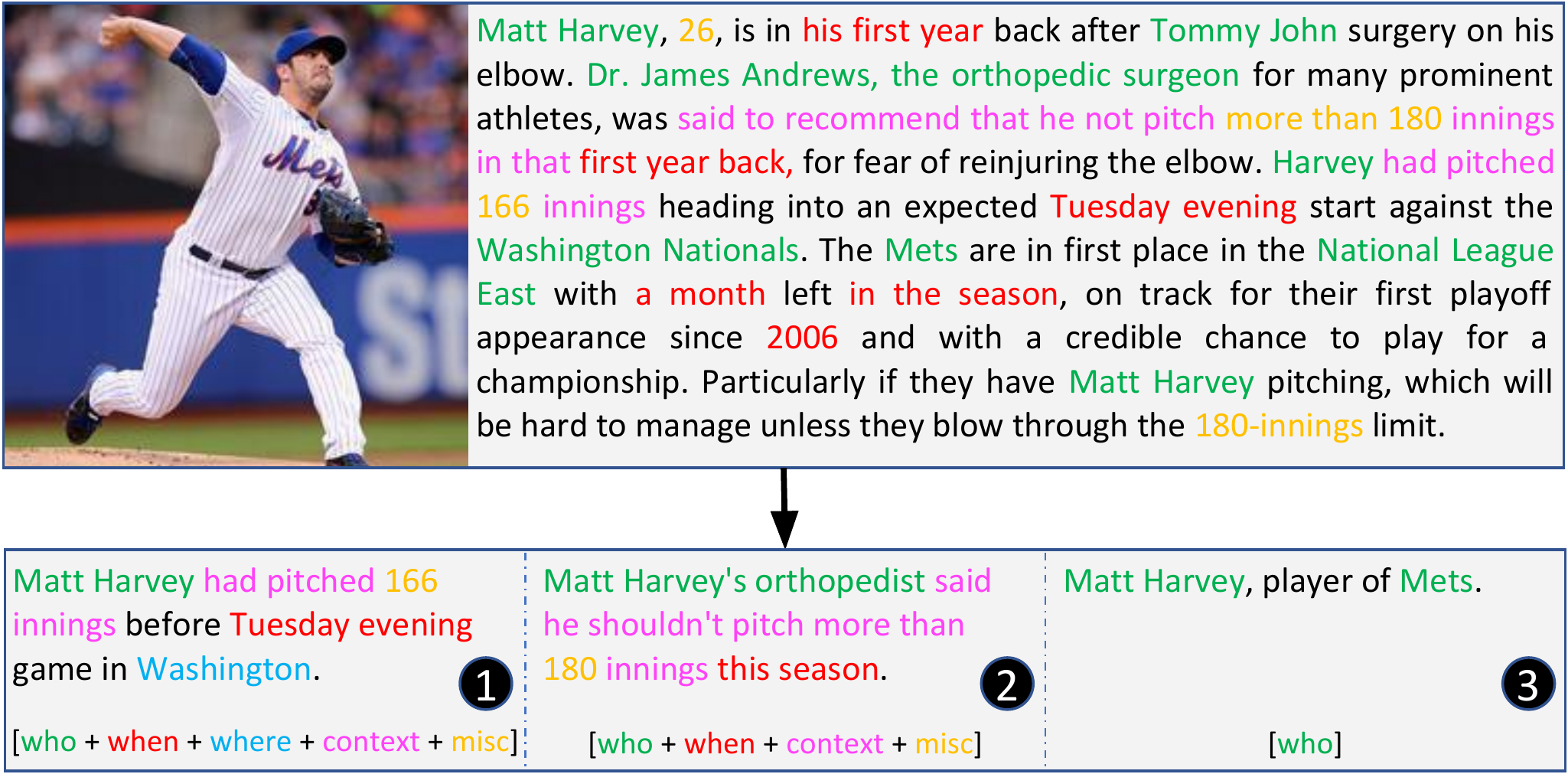}
    \caption{Three possible captions (bottom) for one image-article pair input (top).
    These three captions follow different `templates' composed of 
    \textit{who} ({in \color{ForestGreen}green}),
    \textit{when} ({in \color{red}red}), 
    \textit{where} ({in \color{Cerulean}blue}),
    \textit{context} ({in \color{VioletRed}purple}) and \textit{misc} ({in \color{YellowOrange} orange}) components.}
    \label{fig:captioning}
\end{figure}

The news image captioning
problem~\cite{Feng13,Ramisa18,Biten_2019_CVPR,Tran2020Tell} can be seen as a multi-modal extension of the image captioning task with
additional context provided in the form of a news article.
Specifically, given image-article pairs as input, the news captioning task aims to generate an informative caption that describes the image with proper named entities and context extracted from the article.
The development of automatic news image caption generation methods can ease the process of adding images to articles and produce more engaging content.
According to \textit{The News Manual}\footnote{\url{https://www.thenewsmanual.net/Manuals\%20Volume\%202/volume2_47.htm}} and \textit{International Journalists' Network}\footnote{\url{https://ijnet.org/en/resource/writing-photo-captions}}, a caption should
help news readers understand six main components (\textit{who}, \textit{when}, \textit{where}, \textit{what}, \textit{why}, \textit{how}) related to the image and 
article.
As shown in Fig.~\ref{fig:captioning}, different journalists can write captions to cover different components for the same image and article pair.
Previous news image captioning work~\cite{Biten_2019_CVPR,Tran2020Tell} has not directly addressed the challenge of generating a  caption that follows those journalistic principles.


In this work, we tackle the news image captioning problem by introducing these guidelines in our modeling through a new
concept called a `caption template', which is composed of $5$ key components, detailed in Section~\ref{sec:template}.
We propose a \fullname (\name) model that, given an image-article pair, aims to predict the most likely active template components and, using component-specific decoding block, 
 produces a caption following the provided template guidance.
 \name thus  
 models the underlying structure of the captions, which helps to improve the generation quality. 



Captions for images that accompany news articles often include named entities \textit{and} rely heavily on context found throughout the article (making the text encoding process especially challenging). We propose two techniques to address these issues: (i) integration of features specifically to extract relevant named entities, 
and (ii) a multi-span text reading (MSTR) method, which first splits long articles into multiple text spans and then merges the extracted features of all spans together.

Our work has two main contributions: (i) the definition of the template components of a news caption based on journalistic guidelines, and their explicit integration in the caption generation process of our 
\name model; 
(ii) the design of encoding mechanisms to extract relevant information for the news image captioning task throughout the article, specifically a dedicated named entity representation and the ability to process longer article. Experimental results show better performance than state of the art on news image caption generation. We will release the source code of our method.

\section{Related Work}



\subsection{Generic Image Captioning}
State-of-the-art approaches~\cite{Johnson16,wang2020unique,He2020image,Sammani_2020_CVPR}
mainly use encoder-decoder frameworks with attention to generate captions for images.
\cite{Kelvin2015} developed soft and hard attention mechanisms to focus on different regions in the image when generating different words.
Similarly, \cite{Anderson2018} used a Faster R-CNN~\cite{Ren15} to extract regions of interest that can be attended to.
\cite{Yang2020Fashion} used self-critical sequence training for image captioning.

Our work differs from generic image captioning in 
three 
aspects: (i) our model's input consists of image-article pairs; 
(ii) our caption generation is a guided process following news image captioning journalistic guidelines;
(iii) news captions contain named entities and additional context extracted from the article, making them more complex. 


\subsection{News Article Image Captioning}

\begin{table}[t]
    \centering
    \small
    \begin{adjustbox}{max width=\columnwidth}
    \begin{tabular}{ccc}
    \toprule
    Type & Description & Component \\
    \midrule
    PERSON & People, including fictional & who\\
    NORP & Political groups & who \\
    ORG & Companies, agencies, etc &  who \\
    DATE & Dates or periods & when \\
    TIME & Times smaller than a day & when \\
    FAC & Buildings, airports, highways & where \\
    GPE & Countries, cities, states  & where \\
    LOC & Locations, mountains, waters & where  \\
    PRODUCT & Objects, vehicles, foods & misc \\
    EVENT & Named wars, sports events & misc \\
    ART & Titles of books, songs & misc \\
    LAW & Laws & misc \\
    LAN & Any named language & misc \\
    PERCENT & Percentage, including ``\%'' & misc \\ 
    MONEY & Monetary values & misc \\
    QUANTITY & Measurements & misc \\
    ORDINAL & ``first'', ``second'', etc & misc \\
    CARDINAL & Numerals & misc \\
    \bottomrule
    \end{tabular}
    \end{adjustbox}
    \caption{Named Entities type, description and assigned component category.}
    \label{tab:ne_p}
\end{table}


One of the earliest works in news article image captioning,~\cite{Ramisa18}, 
proposed an encoder-decoder architecture with a deep convolutional model VGG~\cite{Simonyan15} and  Word2Vec~\cite{Mikolov13} as the image and text feature encoder, and an LSTM as the decoder.

\cite{Biten_2019_CVPR} 
 introduced the GoodNews dataset,
 and
proposed a two-step caption generation process using ResNet-152~\cite{Kaiming16} as the image representation and a sentence-level aggregated representation using GloVe embeddings~\cite{pennington2014glove}.
First, a caption is generated with 
placeholders for the different types of named entities: \textit{PERSON}, \textit{ORGANIZATION}, \textit{etc.} shown in the left column of Table \ref{tab:ne_p}.
Then, the placeholders are filled in by matching entities from the best ranked sentences of the article.
This two-step process aims to deal with rare named entities but prevents the captions from being linguistically rich and is can induce error propagation between steps.  


More recently, \cite{Liu2020VisualNewsB, Hu2020, Tran2020Tell} proposed one step, end-to-end methods. 
They all used ResNet-152 as image encoder, while for the text encoder: \cite{Hu2020} applied BiLSTM, \cite{Liu2020VisualNewsB} used BERT and \cite{Tran2020Tell} used RoBERTa.
\cite{Hu2020, Liu2020VisualNewsB} used LSTM as the decoder. 
\cite{Tran2020Tell} introduced the \nyt dataset, 
and a model named \tellfull, which we refer to as \tellns. This model exploits a Transformer decoder and byte-pair-encoding (BPE)~\cite{sennrich2016neural} allowing to generate captions with unseen or rare named entities from common tokens.
As in other multimodal tasks, where studies~\cite{shekhar2019,caglayan2019,li2020multimodal} have shown that the exploitation of both modalities is essential for achieving a good performance, \cite{Tran2020Tell} evaluated a text only model showing that it performs worse than the multimodal model.
We will also evaluate single visual and text modality models in our experiments.
Our work differs from previous work in news image captioning in that \name is an end-to-end framework that (i) integrates journalistic guidelines through a template guided caption generation process; and (ii) exploits a dedicated named entity representation and a long text encoding mechanism.
Our experiments show that our framework significantly outpeforms the state of the art.

\section{Template-Guided News Image Captioning}

In this section, we formally define the news captioning task and introduce the idea of template guidance and our \fullname (\name) approach.
We then propose two strategies 
to address the specific challenges 
of
named entities and long articles. 

\subsection{News Captioning Problem Formulation}
Given an image and article pair ($X^I$, $X^A$), the objective of news captioning is to generate a sentence $\mathbf{y}=\{\mathbf{y}_1,\ldots ,\mathbf{y}_N \}$ with a sequence of $N$ tokens, $\mathbf{y}_i \in V^K$ being the $i$-th token, $V^K$ being the vocabulary of $K$ tokens.
The problem can be solved by an encoder-decoder model. 
The decoder predicts the target sequence $\mathbf{y}$ conditioned on the source inputs $X^I$ and $X^A$.
The decoding probability $P(\mathbf{y}\vert X^I, X^A)$ is modeled using the probability of each target token $\mathbf{y}_n$ at time step $n$ conditioned on the source input $X^I$ and $X^A$ and the current partial target sequence $\mathbf{y}_{<n}$:
\begin{equation}
\resizebox{0.89\hsize}{!}{%
$P(\mathbf{y}\vert X^I, X^A;\bm{\theta}) = \prod_{n=1}^{N} P(\mathbf{y}_n\vert X^I, X^A, \mathbf{y}_{<n};\bm{\theta})$%
}
\end{equation}
where, $\bm{\theta}$ denotes the parameters of the model.

\subsection{Template Guidance}
\label{sec:template-guidance}
To make our model capable of generating captions following different templates, we introduce a new variable $\bm{\alpha}$
for template guidance.
The new decoding probability can be defined as:
\begin{equation}
\resizebox{0.89\hsize}{!}{
$P(\mathbf{y}\vert X^I, X^A) = \prod_{n=1}^{N} P(\mathbf{y}_n\vert X^I, X^A, \bm{\alpha}, \mathbf{y}_{<n})$
}
\end{equation}
where we ignore $\bm{\theta}$ for simplicity.

Based on our definition of templates, 
we could see $\bm{\alpha}$ as the high-level template class
defined by the combination of the active components. 
As there are $5$ template components, the total number of possible template classes is $2^5$.
However, this  poses two challenges to train our model: 
(i) data imbalance, as the most frequent template corresponds to 15.2\% of captions, while the least common ones appear less than 2\% of the time (more details in Tab. 3 of the supplementary material), and 
(ii) different high-level templates may be similar (i.e. having a single component difference) but would be considered totally different classes.

In order to address these issues we define $\bm{\alpha}$ as the set of active components of the template $\bm{\alpha}_{i=1}^5$, with $\bm{\alpha}_i$ being the probability of a template having component $i$. This formulation enables us to exploit the partial overlap in terms of components between the different templates.
Note that the percentage of each component, in Tab.~\ref{tab:tc}, is not as imbalanced as the full template classes. 
The template guidance $\bm{\alpha}$ can be provided by the news writer (`oracle' setting in the experiments) or can be estimated 
(`auto' setting) 
through a multi-label classification task as detailed in the next section and illustrated in the top-left of Fig.~\ref{fig:arch}(a). 


\subsection{Our Model Description}
\begin{figure*}
    \centering
    \includegraphics[width=0.95\textwidth]{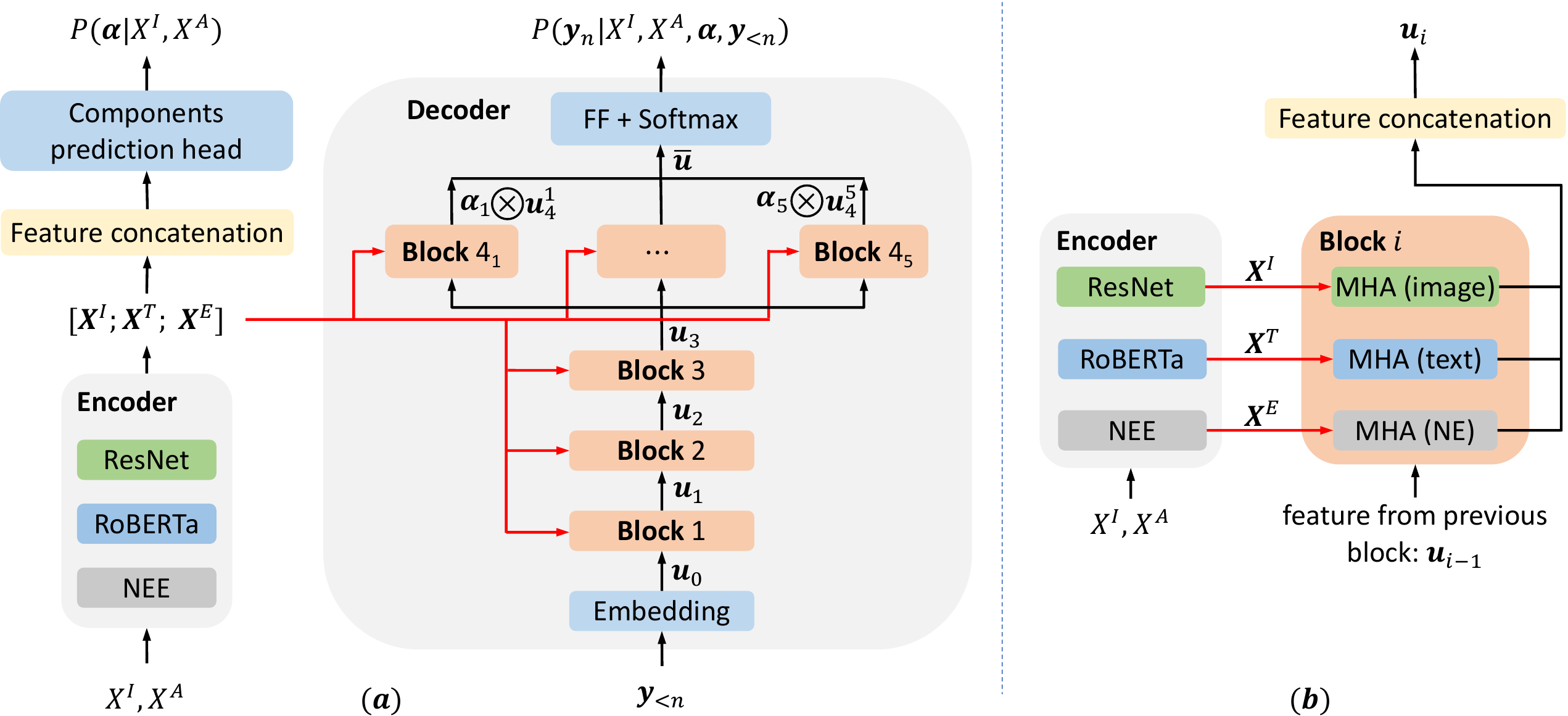}
    \caption{The architecture of our model. (\textit{a}) The Encoder takes image+text+named entities as input and generates features. The Decoder consists blocks 1-4, with blocks 1-3 shared for all template components \textit{who}, \textit{when}, \textit{where}, \textit{context} and \textit{misc}. Block 4 consists of 5 component-specific subblocks ($4_1$-$4_5$). A prediction head on top of the encoder predicts the probabilities of the $5$ components $\bm{\alpha}_{1:5}$, which then multiply the representations of the 5 subblocks $\mathbf{u}^{1:5}_4$. The final representation $\overline{\mathbf{u}}$ is obtained by averaging and used to predict the output token probabilities. (\textit{b}) Every block takes as input the representations from previous blocks as well as those from the Encoder via three Multi-Head Attention (MHA) modules designed for image, text and named entities separately.  
    }
    \label{fig:arch}
\end{figure*}

We propose a news image captioning model that generates captions through template guidance and can also generate accurate named entities and cover a larger extent of the article.
Our \name model, illustrated in Fig.~\ref{fig:arch}, is a transformer-based encoder-decoder, 
with an encoder extracting features from the image $X^I$ and the article $X^A$,
a prediction head estimating the probability of each component 
and a hybrid decoder to produce the caption.

The encoder consists of three parts: 
(i) a ResNet-152 pretrained on ImageNet extracting the image feature $\mathbf{X}^I \in \mathbb{R}^{d_I}$; 
(ii) RoBERTa producing the text features $\mathbf{X}^T \in \mathbb{R}^{d_T}$ from the article; and 
(iii) a Named Entity Embedder (NEE), detailed in Section~\ref{sec:nee}, applied to obtain the features  $\mathbf{X}^E \in \mathbb{R}^{d_E}$ of the named entities in the article.
The components prediction head, taking as input the concatenation of the image, article and named entities features, is a multi-layer perceptron with a sigmoid layer trained (using the components detected in the ground truth caption as target) to output the probability of each component $P(\bm{\alpha}\vert X^I,X^A)$.

The hybrid decoder consists of an embedding layer to get the embeddings of the output generated thus far (i.e., the partial generation), followed by $4$ blocks of $3$ Multi-Head Attention (MHA) modules, denoted as MHA (image/text/NE), to compute the attention across the partial generation and the input image, text and named entities. The final representation $\mathbf{u}_i$ for each block is the concatenation of the $3$ modules' output, Fig.~\ref{fig:arch}(b).
The first $3$ blocks are shared for all components, while
the $4$-th block consists of $5$ parallel component-specific blocks $4_1-4_5$ where block $4i$ outputs the representation $\mathbf{u}_4^i$ 
for the component $i$.
The final representation of the decoder is the average of the weighted sum of all components $\overline{\mathbf{u}}=\frac{1}{5}\sum_{i=1}^5\bm{\alpha}_i\mathbf{u}^i_4$.
Then the output probability $P(\mathbf{y}_{n}\vert X^I,X^A,\bm{\alpha},\mathbf{y}_{<n})$ is obtained by applying a feed-forward (FF) layer, and softmax over the target vocabulary.
Note that our ``template guided'' generation does not limit the number of occurrences of one component in the output caption 
and does not explicitly constrain the generation of specific components but rather the final representation $\overline{\mathbf{u}}$ will rely more on the component-specific representations corresponding to higher $\bm{\alpha}_i$ values. 


\subsubsection{Named Entity Embedding}
\label{sec:nee}
With over $96\%$ 
(see Tab. 1 in the supplementary material) of the news captions containing named entities, producing accurate named entities is essential to generating  good news captions.
However, text encoders like RoBERTa cannot properly represent named entities, and only handle them implicitly through BPE (Byte-Pair Encoding) subwords.

To deal explicitly with named entities, we learn entity embeddings from the Wikipedia knowledge base (KB),
following Wikipedia2vec~\cite{Wikipedia2Vec18} 
which embeds words and entities into a common space\footnote{\url{https://wikipedia2vec.github.io/wikipedia2vec/}}.
Given a vocabulary of words $V_{W}$ and a set of entities $V_{E}$, it learns a lookup embedding function $\mathcal{E}_{Wiki}:V_{W}\cup V_{E}\rightarrow \mathbb{R}^{d_{Wiki}}$.
There are three components in Wikipedia2Vec: (i) a skip-gram model for learning the word similarity in $V_{W}$, (ii) a KB graph model to learn the relatedness between pairs of entities  (vertices $V_{E}$ of the Wikipedia entity graph) and (iii) a version of Word2Vec where words are predicted from entities. 

Since predicting the correct named entities from context is very important for news captioning, we introduce a fourth component: (iv) a neural entity predictor (NEP).
Given a text (sequence of words) $t=\{w_1,\ldots,w_N \}$, we train Wikipedia2vec to predict the entities ${e_1,\ldots,e_m}$ that appear in the sequence.
With $E_{KB}$ being the set of all entities in KB, and $v_e$ and $v_t$ (computed as the element-wise mean of all the word vectors in $t$ followed by a fully connected layer) the vector representations of the entity $e$ and the text $t$, respectively, the probability of an entity $e$ appearing in text $t$ is defined as
\begin{equation}
    P(e\vert t)=\frac{\exp(v_e^{T}v_t)}{\sum_{e^{\prime}\in E_{KB}}\exp(v^T_{e^{\prime}}v_t)}.
    \label{eq:iv}
\end{equation}

We optimize the NEP model with a cross-entropy loss, but using Eq.~\ref{eq:iv} as is would be computationally expensive as it involves a summation over all entities in the KB.
We address this by replacing $E_{KB}$ in Eq. \ref{eq:iv} with $E^{\ast}$, the union of the positive entity $e$ and $50$
randomly chosen negative entities not in $t$.
Through exploiting the Named Entity Embedding (NEE), our model can represent and thus generate more accurate entities.
The NEE model is not jointly trained with the template components prediction and caption generation heads of \name, but pre-trained offline on 
Wikipedia KB. 

The Wikipedia KB contains a large set of NEs but cannot cover all NEs that could appear in a news article (about $40\%$ are not covered in our datasets). The embedding of a new NE cannot be obtained directly by lookup.
To alleviate this problem, we set the embedding of any missing NE with $v_t$ which is reasonable as we trained the NEP to maximize the correlation between $v_e$ and $v_t$ in Eq.~\ref{eq:iv}.



\subsubsection{Reading Longer Articles}
\label{sec:text}
\cite{Biten_2019_CVPR} use sentence-level features 
obtained by averaging the word features, of a pretrained GloVe \cite{pennington2014glove} model, in the sentence.
While this method can embed the whole article, the averaging makes the feature less informative.
\cite{Tran2020Tell} instead use RoBERTa 
as the text feature extractor, though this has the limitation of exploiting only 512 tokens.

However, processing only the first 512 tokens may ignore important contextual information 
appearing later in the news article.
To alleviate this problem, we propose a \textit{Multi-Span Text Reading} (MSTR) method to read more than $512$ tokens from the article.
MSTR splits the text into overlapping segments of $512$ tokens and pass them to the RoBERTa encoder independently.
The representation of any overlapping token in 2 segments is the element-wise interpolation of their representations.




\section{Experiments}

\begin{table*}[t]
\small
	\centering
	\begin{adjustbox}{max width=\textwidth}
	\begin{tabular}{cc|cccc|cc|cc}
		\cmidrule{3-10}
         & & \multicolumn{4}{c|}{\centering\textbf{\small{General Caption Ceneration}}}
		 & \multicolumn{2}{c|}{\textbf{\small{Named Entities}}} &
        \multicolumn{2}{c}{\textbf{\small{Components}}}\\
		 \cmidrule{3-10}
		 &  & \small{BLEU-4} & \small{ROUGE} & \small{METEOR} & \small{CIDEr} & \small{$P$} & \small{$R$} & \small{$\overline{P}$} & \small{$\overline{R}$} \\
		\midrule
		\multirow{12}{*}[-1cm]{\rotatebox[origin=c]{90}{\good}}
		& SAT~\cite{Kelvin2015} &  0.73 & 11.88 & 4.14 & 12.15 & 8.19 & 7.10 & -- & -- \\
		& Att2in2~\cite{Rennie2017} & 0.76 & 11.58 & 3.90 & 11.58 & -- & -- & -- & -- \\
		& BUTD~\cite{Anderson2018} & 0.71 & 11.06 & 3.74 & 11.02 & -- & -- & -- & -- \\
		& Adaptive Att \cite{Lu2017Adaptive} & 0.51 & 10.94 & 3.59 & 10.55 & -- & -- & -- & -- \\
		& Avg+CtxIns~\cite{Biten_2019_CVPR} & 0.89 & 12.20 & 4.37 & 13.10 & 8.23 & 6.06 & 20.51 & 18.72 \\
		& TBB+AttIns~\cite{Biten_2019_CVPR} & 0.76 & 12.20 & 4.17 & 12.70 & 8.87 & 5.64 & 20.23 & 18.45  \\
		\cmidrule{2-10}
		& VGG+LSTM \cite{Ramisa18} & 0.31 & 6.38 & 1.66 & 1.28 & -- & -- & -- & -- \\
		& VisualNews \cite{Liu2020VisualNewsB} & 5.1 & 19.3 & 8.8 & 43.7 & 19.6 & 17.9 & -- & --\\
		& Tell~\cite{Tran2020Tell} 
		& 5.45 & 20.70 & 9.74 & 48.50 & 21.10 & 17.40 & 69.52 & 63.31 \\
		& Tell (full)~\cite{Tran2020Tell} & 6.05 & 21.40 & 10.30 & 53.80 & 22.20 & 18.70 & 71.55 & 64.93 \\
		 \cmidrule{2-10}
		 & \name (zero-out text) & 1.71 & 13.04 & 5.23 & 9.61 & 4.42 & 3.01 & 18.92 & 16.77 \\
		 & \name (zero-out image) & 4.10 & 17.33 & 8.41 & 38.49 & 18.03 & 15.12 & 48.74 & 46.29 \\
		 & \name (image only) & 1.86 & 13.28 & 5.97 & 10.20 & 4.46 & 3.31 & 19.07 & 17.13  \\
		 & \name (text only) & 5.28 & 19.07 & 9.17 & 50.04 & 20.43 & 18.13 & 49.56 & 46.98  \\ 
		 & \name (auto) & 6.34 & 21.65 & 10.78 & 59.19 & 24.60 & 20.90 & 75.51 & 66.27 \\
		 &  \namens+\neeshort (auto) & 6.73 & 22.68 & 11.18 & 59.50 & 25.87 & 21.63 & 74.42 & 68.53 \\
		 & \namens+MSTR (auto) & 6.45 & 21.99 & 10.83 & 59.65 & 24.75 & 21.61 & 75.57 & \textbf{70.04} \\
		 & \namens+MSTR+\neeshort (auto) & \textbf{6.83} & \textbf{23.05} & \textbf{11.25} & \textbf{61.22} & \textbf{26.87} & \textbf{22.05} & \textbf{75.83} & 
		 68.85 \\
		 \cmidrule{2-10}
		 & \name (oracle) & \textit{7.06} & \textit{24.13} & \textit{11.72} & \textit{69.23} & \textit{28.40} & \textit{23.48} & \textit{92.96} & \textit{87.86} \\
		 & \namens+MSTR+\neeshort (oracle) & \textit{7.36} & \textit{24.25} & \textit{11.98} & \textit{69.76} & \textit{28.59} & \textit{23.68} & \textit{92.46} & \textit{87.55}  \\
		\midrule
		\midrule
		\multirow{8}{*}[-0.5cm]{\rotatebox[origin=c]{90}{\nyt}}
		 & Tell~\cite{Tran2020Tell} 
		 & 5.01 & 19.40 & 9.05 & 40.30 & 20.0 & 18.10 & 67.13 & 62.24 \\
		 & Tell (full)~\cite{Tran2020Tell} & 6.30 & 21.70 & 10.30 & 54.40 & 24.60 & 22.20 & 69.72 & 63.52 \\
		 \cmidrule{2-10}
		 & \name (zero-out text) & 1.42 & 12.66 & 5.08 & 9.33 & 4.23 & 2.89 & 18.87 & 16.53 \\
		 & \name (zero-out image) & 3.88 & 15.64 & 7.76 & 32.01 & 21.15 & 14.84 & 53.71 & 51.29 \\
		 & \name (image only) & 1.50 & 12.58 & 5.68 & 9.93 & 4.49 & 2.88 & 19.40 & 17.12  \\
		 & \name (text only) & 4.95 & 18.47 & 8.54 & 41.27 & 20.52 & 18.48  & 54.89 & 52.31 \\
		 
		 & \name (auto) & 6.39 & 22.38 & 10.75 & 56.54 & 27.35 & 23.73 & 73.37 & 65.79 \\
		 &  \namens+\neeshort (auto) & 6.66 & 22.72 & 10.85 & 59.02 & 26.81 & 23.20 & 73.02 & \textbf{66.54} \\
		 &  \namens+MSTR (auto) & 6.44 & 22.63 & 10.88 & 57.61 & 26.41 & 23.67 & 73.36 & 66.30 \\
		 & \namens+MSTR+\neeshort (auto) & \textbf{6.79} & \textbf{22.80} & \textbf{10.93} & \textbf{59.42} & \textbf{28.63} & \textbf{24.49} & \textbf{73.51} & 65.49 \\
		 \cmidrule{2-10}
		 & \name (oracle) & \textit{7.44} & \textit{24.09} & \textit{11.93} & \textit{65.53} & \textit{28.53} & \textit{26.09} & \textit{90.76} & \textit{87.99} \\
		 & \namens+MSTR+\neeshort (oracle) & \textit{7.68} & \textit{24.09} & \textit{12.09} & \textit{66.15} & \textit{28.79} & \textit{26.35} & \textit{90.07} & \textit{87.92} \\
		\bottomrule
	\end{tabular}
	\end{adjustbox}
	\caption {Results on \good and \nyt. We highlight the \textbf{best} model in bold. Note that we directly use the results reported in~\cite{Tran2020Tell} for the baseline models.\label{tab:results}}
\end{table*}

We evaluate \name on two large-scale publicly available news captioning datasets: 
\good \cite{Biten_2019_CVPR} 
and 
\nyt 
\cite{Tran2020Tell}
both collected using The New York Times public API\footnote{https://developer.nytimes.com/apis}, with the latter being larger and containing longer articles. 
We follow the evaluation protocols defined by the authors of each dataset and used by previous works with $421$K training, $18$K validation, and $23$K test captions for \good and $763$K training, $8$K validation and $22$K test captions for \nyt.
We provide further details about the datasets in the supplementary material.


\subsection{Methods \& Metrics}

We implement \name as a Transfomer-based 
encoder-decoder architecture similar to Tell but with our proposed template guidance.
We introduce \namens+\neeshort as \name with enriched named entity embeddings (Section~\ref{sec:nee}), and \namens+MSTR as \name with multi-span text reading technique (Section~\ref{sec:text}).
To evaluate how \name exploits template guidance, we introduce the \name (oracle) and \namens+MSTR+\neeshort (oracle) variants, where ground truth template components are provided through $\bm{\alpha}$. 
We evaluate if our model exploits both the text and image input in two ways. 
We first report results of our multimodal model where at test time we zero-out text features (i.e. $\mathbf{X}^T$ and $\mathbf{X}^E$ are set to all zero vectors) \name (zero-out text) or image features 
\name (zero-out image). 
We also train single-modality models with only an image encoder (\name image only) or a text encoder (\name text only).


We compare against two types of baselines.
(i) Two-step generation methods: that are based on conventional image captioning models \cite{Kelvin2015,Rennie2017,Anderson2018,Lu2017Adaptive,Biten_2019_CVPR} to first generate captions with
placeholders and then insert named entities into these placeholders.
(ii) End-to-end models: VGG+LSTM~\cite{Ramisa18}, VisualNews~\cite{Liu2020VisualNewsB} that uses ResNet as image encoder,
BERT article encoder and bi-LSTM as decoder, and Tell, with two variants: 
(a) Tell, which uses RoBERTa and ResNet-152 as the encoders and Transformer as the decoder, it is equivalent to \name without template guidance as they use the same encoders and training settings.
(b) Tell (full), which includes two additional visual encoders: YOLOv3 and MTCNN, and Location-Aware and Weighted RoBERTa for text encoding.

For the general caption generation quality evaluation, we use the BLEU-4 \cite{papineni2002}, ROUGE \cite{lin2004rouge}, METEOR \cite{denkowski2014meteor} and CIDEr \cite{Vedantam15cider} metrics.
We also use named entity precision/recall
to evaluate the named entity generation quality.
To better understand how well the generated captions follow the ground truth templates, we calculate precision and recall for the five components \textit{who}, \textit{when}, \textit{where}, \textit{context} and \textit{misc} and use the averaged precision and recall\footnote{Per-component results are provided in Table 4 of the supplementary material.} as the final metric.

\subsection{Implementation and Training details}

Following \cite{Tran2020Tell}, we set the hidden size of the input features $d_I=2048$, $d_T=1024$ and $d_E=300$ and the number of heads $H=16$.
We use the Adam optimizer~\cite{kingma2015} with $\beta_1=0.9$, $\beta_2=0.98$, $\epsilon=10^{-6}$.
The number of tokens in the vocabulary $K=50264$ and $d^{Wiki}=300$.
We limit the text length in MSTR to 1,000 tokens 
as preliminary studies have shown similar performance with longer text input but at the expense of significant increased training time (Tab. 6 in supplementary). 
In practice, for an article longer than 512 tokens, we read two overlapping text segments of 512 tokens, one starting from the beginning and another from the end and thus can have $[24-511]$ overlapping tokens.
The components prediction head in Fig.~\ref{fig:arch} is a linear layer followed by an output layer of $1024$ dimensions.

The training pipeline uses PyTorch~\cite{paszke2017automatic} and
the AllenNLP framework~\cite{gardner2018allennlp}. The RoBERTa model and
dynamic convolution code are adapted from fairseq~\cite{ott2019fairseq}.
We use a maximum batch size of $16$ and training is stopped after the model has seen 6.6 million examples, corresponding to $16$ epochs on \good and $9$ epochs on \nyt. 
Training is done with mixed precision to reduce the memory footprint and allow our full model to be trained on a single V-100 GPU for 4 to 6 days on both datasets.

\subsection{Evaluation}



\begin{figure*}[t]
    \centering
    \small
    \includegraphics[width=\textwidth]{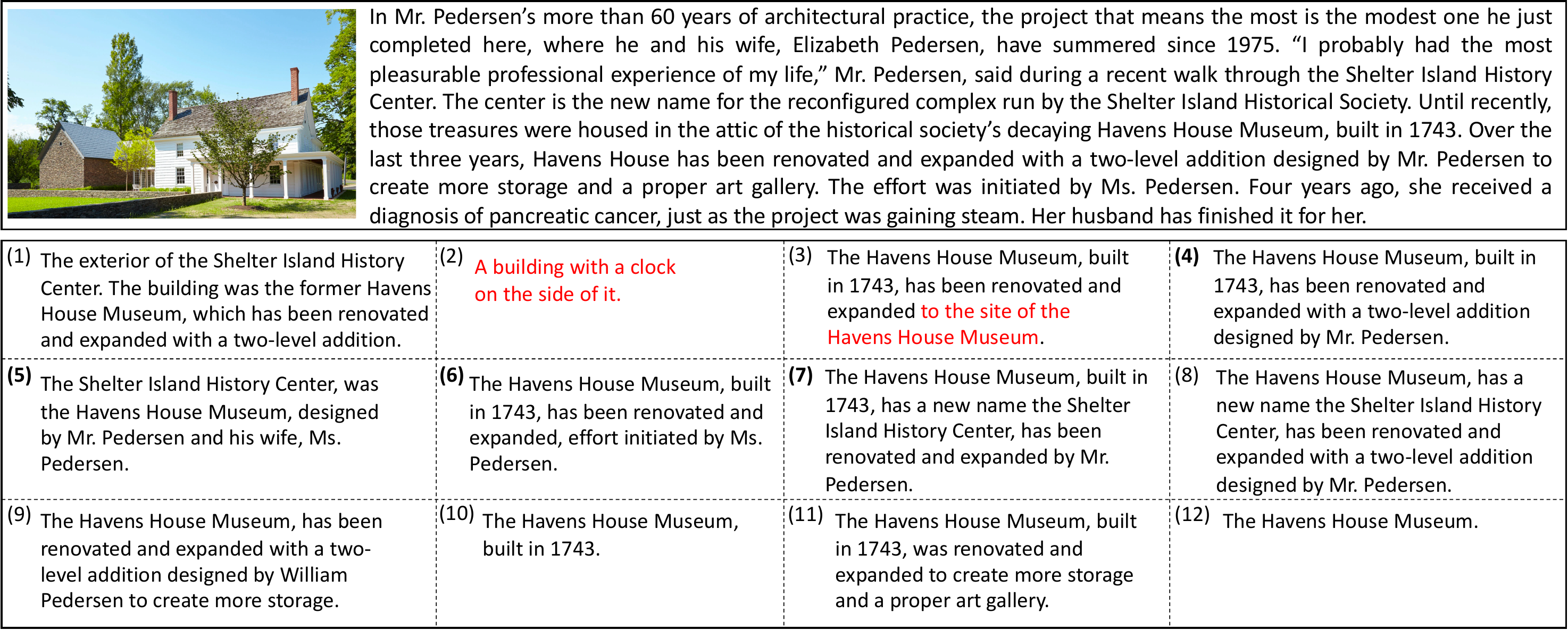}
    \caption{An example of news caption generation. The captions are generated by: (1) human (ground truth caption). (2) conventional image captioning model SAT. (3) Tell. (4)  \name. (5)  \namens+\neeshort. (6) \namens+MSTR. 
    \namens+MSTR+\neeshort (7) auto, (8) oracle, with template (9) \textit{who} + \textit{context}, (10) \textit{who} + \textit{when}, (11) \textit{who} + \textit{when} + \textit{context}, and (12) \textit{who}.
     For the generated captions, we highlight wrong statements in {\color{red}red}.
    }
    \label{fig:qualitative}
\end{figure*}

\subsubsection{General Caption Generation}
We first discuss the results with the general caption generation metrics BLEU-4, ROUGE, METEOR and CIDEr reported in Table~\ref{tab:results}.
We report the mean values of three runs, and the maximum standard deviations of our variants on BLEU, ROUGE, METEOR, CIDEr are 0.013, 0.019, 0.016 and 0.069, which shows the stability of our results and that our method improvements are notable.
For the \good dataset, \name (auto) provides an improvement of $0.89$, $0.95$, $1.04$, $10.69$ points over Tell on the four metrics respectively, while the full model \namens+MSTR+\neeshort (auto) has an even bigger improvement of $1.38$, $2.35$, $1.51$, $12.72$.
The improvement is especially impressive for the CIDEr score.
\name performs much better than all the two-step captioning methods (first group of results) and VGG+LSTM.
For the \nyt dataset, we compare our models only to Tell since other models perform much worse.  
Here, our full model 
achieves $6.79$, $22.80$, $10.93$ and $59.42$ 
with $1.78$, $3.40$, $1.88$, $19.12$ points improvement over Tell.
Our \namens+MSTR+\neeshort (auto) outperforms Tell (full) which exploits additional visual features on both datasets .
This demonstrates the effectiveness of our model in generating good captions.
By providing the oracle  $\bm{\alpha}$, the \namens+MSTR+\neeshort (oracle) can achieve even higher performance on almost all metrics, showing the value of our template guidance process. 

From the single modality evaluation, we observe that models that exploit the text only (\name (zero-out image) and \name (text only)) perform better than those relying on the image only (\name (zero-out text) and \name (image only)) but all have lower performance than multimodal models, confirming that both modalities are important for news image captioning.

\subsubsection{Named Entity Generation}
One of the main objectives of news captioning is to generate captions with accurate named entities.
As shown in Tab.~\ref{tab:results}, compared to Tell, \namens+MSTR+\neeshort (auto) increases the named entity precision and recall scores by $5.77\%$ 
and $4.65\%$ 
on \good, and $8.63\%$ 
and $6.39\%$ 
on \nyt.
The oracle versions of our models attain even higher performances.

\subsubsection{Template Components Evaluation}
The average precision and recall of the template components, reported in the two rightmost columns of Tab.~\ref{tab:results},
of \namens+MSTR+\neeshort (auto)  increases by 
$6.3\%$
and 
$5.5\%$ 
on \good dataset and 
$6.4\%$ 
and 
$3.3\%$ 
on \nyt dataset compared to Tell.
By providing the oracle $\bm{\alpha}$, 
even better results are obtained, demonstrating that our model can exploit template guidance.


\subsubsection{Qualitative \& Human Evaluation}
In Figure~\ref{fig:qualitative}  
we show the image, article (shortened for visualization) and the captions generated by a conventional image captioning model SAT~\cite{Kelvin2015}, Tell~\cite{Tran2020Tell} and different \name variants.
The captions generated by all \name variants are meaningful and closer to the ground truth than the baselines.
Interestingly, most captions generated by \name variants include people's names, e.g. Mr. or Ms. \textit{Pedersen}
in addition to the building names
probably because people's names are the most common type for the component \textit{who} in the datasets (see Tab. 1 of the supplementary material).
As MSTR can read longer text than Tell, 
\namens+MSTR can exploit 
the end of the article and generates the text span \textit{effort initiated by Ms. Pedersen}.
The caption generated by \namens+MSTR+\neeshort has all the key factors in the ground truth caption (\textit{the Havens House Museum}, \textit{the Shelter Island History Center}, \textit{been renovated and expanded}) 
demonstrating the strengths of our model.
The captions generated using the oracle $\bm{\alpha}$ (8) as well as some other manually defined $\bm{\alpha}$ (9-12)
 illustrate the benefits 
and 
flexibility of our template guidance in \name.


Finally, we conducted a human evaluation through crowd-sourcing on Amazon Mechanical Turk on 200 random image-article pairs sampled from the test set of the NYT800K dataset. 
For each image-article pair, three different raters
were requested to rate the ground truth caption, the caption generated by Tell, and captions generated by 4
variants
of our model, on a 4 point scale.
Raters were asked to evaluate separately  how well the caption was describing the \textit{image}, how relevant it was to the \textit{article}, and how easy to understand the \textit{sentence} was.
We report the average of the three ratings in Tab.~\ref{tab:human}, showing that all variants of our model produce captions that are rated better than Tell and closer to the ground truth captions ratings on the three aspects.
Details on the annotation instructions and results are given in the supplementary material.

\begin{table}[!t]
\centering
\small
\begin{tabular}{cccccccl}
\toprule
\textbf{Model} & image & article & sentence \\
\midrule
Ground Truth & 2.96 & 2.86 & 3.08 \\ 
\midrule
Tell & 2.80 & 2.80 & 2.92 \\
\midrule
\name & 2.87 & 2.86 & 2.97 \\
\namens+\neeshort & 2.88 & \textbf{2.92} & \textbf{2.99} \\
\namens+MSTR & \textbf{2.89} & 2.86 & 2.98 \\
\namens+MSTR+\neeshort & 2.86 & 2.88 & \textbf{2.99} \\
\bottomrule
\end{tabular}

\caption{\textbf{Human evaluation on the generated captions.} We highlight the \textbf{best} model in bold.}
\label{tab:human}
\end{table}

\section{Conclusion}
News image captioning is a challenging task as it requires exploiting both image and text content to produce rich and well structured captions including relevant named entities and information gathered from the whole article.  
In this work, we presented \fullname, aiming to solve the news image captioning task by integrating domain specific knowledge in both the representation and caption generation process. On the representation side, we introduced two techniques: named entity embedding (\neeshortns) and multi-span text reading (MSTR).
Our decoding process explicitly integrates the key components a journalist would seek to describe to  improve the caption generation quality.
Our method obtains remarkable gains on both \good and \nyt 
datasets relative to the state-of-the-art.
\chapter{Conclusions}
The direct $P(y \vert x)$ probabilistic models form the fundamental aspect of modern artificial intelligence.
Such models can be made incredibly flexible by parameterizing the conditional distributions with differentiable deep neural networks. 

Optimization of such models using maximum likelihood estimation objective is straightforward.
Many excellent papers have been presented to solve these problems either in Computer Vision (e.g., AlexNet, VGG, ResNet, etc.), or in Natural Language Processing (e.g., LSTM, GRU, Transformer, etc.).
However, two major drawbacks exists in these simple graphical models.
Firstly, the training processes usually involve a huge amount of data.
Labeling that amount of data is a non-trivial task.
Without enough data, the deep learning methods will suffer from the overfitting problem.
Secondly, most of the methods cannot control or explain the results in a straight-forward way.
For example, in the neural machine translation models, the next word is predicted as the one with the most probability.
How can we associate this word with some more easily understandable factors, like syntax?
Can we achieve the goal of control the generation by controlling the syntax?
Existing methods for such problems were either relatively inefficient, complex or not applicable to models with neural networks as components.

In this work, our main contribution is to propose a straightforward method by introducing a latent variable $z$, which represents the knowledge extracted using the external tools or sources.
The new graphical model therefore becomes $P(y \vert x, z)$.
In this work, we explored several ways of integrating this latent variable.
In the image to image translation problem, we use the domain labels as the latent variable.
The goal of multi-domain image to image translation is to translate the images from one domain to any other domains.
The SOTA methods generally train several bi-domain image translation models which are quite inefficient.
Using domain labels as the external knowledge can help control the translation process, helping the model to understand which domain to translate to.
The model is shared across all domains, giving the flexibility of training.
In the neural machine translation problem, the SOTA methods just use source and target sentences as the training data, without using any kind of syntax.
We utilized Part-Of-Speech sequences as the external knowledge and design a method to integrate the POS sequence with the encoder-decoder model.
The intuition behind this design rooted in the learning of English.
Do we really need to know syntax to learn good English? 
Apparently, syntax is important if we really want to have a systematic English knowledge.
Similarly, for the news image captioning problem, to find a good $z$, the first question the author asked himself was ``How do the journalists writing captions?''
Then the author found the answer by utilizing the online books teaching how to write captions.
Thus, when we are tackling new problems, the first thing to do is to forget about deep learning and the existing SOTA methods, but to focus on the problem itself and ask ``How did human beings solve the problem before we had deep learning?''
With this intuition in mind, it would be easier to find a solution to introduce $z$ to the problem.
Then, we can start to solve $P(y\vert x, z)$ either in supervised learning or unsupervised learning.
In this work, we try to optimize the lower bound of the log-likelihood and to dissemble $P(y\vert x, z)$ into several part and solve each part separately.

Finally, We hope this work can inspire people to have new perspectives of solving problems. 
We suspect that further improvement is feasible when this idea is applied to other research problems.

\newpage
\chapter{List of Publications}
As required, we provide a list of publications whose content was used in this thesis, and provide the contributions of co-authors.

The author has four published papers, as follows:
\begin{itemize}
    \item Xuewen Yang, Heming Zhang, Yingru Liu, CH Wu, Jianchao Tan, Jue Wang, Xin Wang. Fashion Captioning: Towards generating accurate descriptions with semantic rewards. In Proceedings of the ECCV 2020.
    \item Xuewen Yang, Yingru Liu, Dongliang Xie, Xin Wang, Niranjan Balasubramanian, et al. Latent Part-of-Speech Sequences for Neural Machine Translation. In Proceedings of the EMNLP 2019.
    \item Xuewen Yang, Dongliang Xie, Xin Wang, Jiangbo Yuan, Wanying Ding, Pengyun Yan. Learning tuple compatibility for conditional outfit recommendation. In Proceedings of the ACM Multimedia 2020.
    \item Xuewen Yang, Dongliang Xie, Xin Wang. Crossing-domain generative adversarial networks for unsupervised multi-domain image-to-image translation. In Proceedings of the ACM Multimedia 2018.
\end{itemize}

The author has three pre-print papers that are still in submissions.
\begin{itemize}
    \item Xuewen Yang, Yingru Liu, Xin Wang, ReFormer: The Relational Transformer for Image Captioning. arXiv 2021.
    \item Xuewen Yang, Svebor Karaman, Journalistic Guidelines Aware News Image Captioning, In submission to EMNLP 2021.
    \item Xuewen Yang, Xin Wang. Recognizing License Plates in Real-Time. arXiv 2017.
\end{itemize}

The author has three papers as the co-author of others:
\begin{itemize}
    \item Yingru Liu, Y Xing, Xuewen Yang, Xin Wang, et al. Learning Continuous-Time Dynamics by Stochastic Differential Networks. arXiv 2020.
    \item Yingru Liu, Xuewen Yang, Dongliang Xie, Xin Wang, L Shen, H Huang, Niranjan Balasubramanian. Adaptive Activation Network and Functional Regularization for Efficient and Flexible Deep Multi-Task Learning. AAAI 2020.
    \item Heming Zhang, Xuewen Yang, Jianchao Tan, CH Wu, Jue Wang, CCJ Kuo. Learning Color Compatibility in Fashion Outfit. arXiv 2020.
\end{itemize}


\bibliographystyle{unsrtnat}
\renewcommand{\baselinestretch}{1}
\normalsize
\addcontentsline{toc}{chapter}{Bibliography}


\bibliography{dissertation}

\end{document}


\maketitle

\section{Appendix}

\subsection{Implementation and Training Details}
We implement all Transformer-based models using \texttt{Fairseq}~\footnote{https://github.com/pytorch/fairseq} Pytorch framework. 

For all translation tasks, we choose the \textit{base} configuration of Transformer with $d_{model}=512$.
During training, we choose Adam optimizer~\cite{Kingma14} with $\beta_1 = 0.9$, $\beta_2 = 0.98$. The initial learning rate is 0.0002 with 4000 warm-up steps. The learning rate is scheduled with the same rule as in~\cite{Vaswani17}. Each batch on one GPU contains roughly 2000 tokens for IWSLT tasks and 800 tokens for the WMT En-De task. We train IWSLT tasks using two 1080Ti GPUs and train WMT task using 8 K80 GPUs. The hyperparameter $\lambda$ is set to 0.2. For inference, we use beam search with beam size 5 to generate candidates. 
\subsection{Dataset Details}
We evaluate our model on two small translation datasets - IWSLT'14 German-English (De-En) and English-French (En-Fr)~\cite{Cettolo14} and a much bigger one - WMT'14 English-German (En-De). 

\textbf{IWSLT'14 En-De/En-Fr} 
We use the datasets extracted from IWSLT 2014 machine translation evaluation campaign~\cite{Cettolo14}, which consists of 153K/220K training sentence pairs for En-De/En-Fr tasks. For En-De, we use 7K data split from the training set as the validation set and use the concatenation of dev2010, tst2010, tst2011 and tst2012 as the test set, which is widely used in prior studies~\cite{huang2018towards, Tianyu19, bahdanau2016actorcritic, Ranzato16}.
For En-Fr, the tst2014 is taken as the validation set and tst2015 is used as the test set, which is the same with prior studies~\cite{denkowski2017, Cheng18Towards}.
We also lowercase the sentences of En-De and En-Fr following general practice. Before encoding sentences using sub-word types based on byte-pair encoding~\cite{sennrich2016neural}, which is a common practice in NMT, we parse POS tag sequences of the sentences using Stanford Parser~\cite{Chen14}. The POS tag sequences produce POS vocabulary of size 32 for both English and French and 32 for German. Sentences are then encoded using sub-word types. To make the lengths of POS tag sequences equal to their corresponding sub-word sentences, if several sub-words belong to the same word, they are given the same POS tag. For IWSLT'14 En-De dataset, we build a English sub-word vocabulary of size 6632 and a German sub-word vocabulary of size 8848. For En-Fr dataset, we build a English sub-word vocabulary of size 7172 and a French sub-word vocabulary of size 8740.

\textbf{WMT'14 English-German (En-De)} 

We use the same dataset as~\cite{Vaswani17}, which consists of 4.5M sentence pairs. We use the concatenation of newstest2012 and newstest2013 as the validation set and newstest2014 as the test set. Sentences are encoded using byte-pair encoding with a shared vocabulary of about 40K sub-word tokens. The method to generate POS tag sequences is the same, except that we merge some POS tags of similar meaning to one and get a POS tag vocabulary of size 16 for both German and English. This operation reduces computational cost, and gives us a bigger batch for training.
\bibliography{emnlp-ijcnlp-2019}
\bibliographystyle{acl_natbib}


\maketitle

\begin{abstract}
    This supplementary material provides more information about the two datasets \good and \nyt, template statistics and prediction results, implementation and training details, model difference between Tell and \name, details on the human evaluation and ablation evaluations of another sequence length efficient Transfomer - Longformer as well as different sequence length for MSTR.
\end{abstract}

\section{Dataset}
The Visual Genome dataset~\cite{Krishna17} consists of $51$ relation types, shown in Tab.~\ref{tab:vg_rel}.

\begin{table}[!p]
\caption{List of relation labels in Visual Genome dataset.}
\centering
\begin{tabular}{cp{8cm}cl}
\toprule
background, above, across, against, along, and, at,\\
attached to, behind, belonging to, between, carrying, \\
covered in, covering, eating, flying in, for, from, \\
growing on, hanging from, has, holding, in, in front of, \\
laying on, looking at, lying on, made of, mounted on, \\
near, of, on, on back of, over, painted on, parked on, \\
part of, playing, riding, says, sitting on, standing on, \\
to, under, using, walking in, walking on, watching, \\
wearing, wears, with \\
\bottomrule
\end{tabular}
\label{tab:vg_rel}
\end{table}

\begin{table*}[!t]
\centering
\begin{tabular}{ccccccccccl}
\toprule
\multirow{2}{*}{Method} & \multicolumn{2}{c|}{B-1}  & \multicolumn{2}{c|}{B-4} & \multicolumn{2}{c|}{M} & \multicolumn{2}{c|}{R} & \multicolumn{2}{c|}{C}   \\
\cmidrule{2-11}
&c5 & c40 &c5 & c40 &c5 & c40 &c5 & c40 &c5 & c40  \\
\midrule
Up-Down~\cite{Anderson2018} &  80.2 & 95.2 & 36.9 & 68.5 & 27.6 & 36.7 & 57.1 & 72.4 & 117.9 & 120.5 \\
Att2all~\cite{Rennie2017} &  78.1 & 93.7 & 35.2 & 64.5 & 27.0 & 35.5 & 56.3 & 70.7 & 114.7 & 116.7 \\
AoA~\cite{huang2019attention} &  81.0 & 95.0 & 39.4 & 71.2 & 29.1 & 38.5 & 58.9 & 74.5 & 126.9 & 129.6 \\
\midrule
ETA~\cite{Li_2019_ICCV} & 81.2 & 95.0 & 38.9 & 70.2 & 28.6 & 38.0 & 58.6 & 73.9 & 122.1 & 124.4  \\
Image-T~\cite{He2020image} & 81.2 & 95.4 & 39.6 & 71.5 & 29.1 & 38.4 & 59.2 & 74.5 & 127.4 & 129.6 \\
$\mathcal{M}^2$-T~\cite{cornia2020m2} & 81.6 & 96.0 & 39.7 & 72.8 & 29.4 & 39.0 & 59.2 & 74.8 & 129.3 & 132.1 \\
\midrule
GCN~\cite{Yao2018} & 80.8 & 95.9 & 38.7 & 69.7 & 28.5 & 37.6 & 58.5 & 73.4 & 125.3 & 126.5 \\
SGAE~\cite{yang2019} &  -- -- & -- -- & 37.8 & 68.7 & 28.1 & 37.0 & 58.2 & 73.1 & 122.7 & 125.5 \\
VSUA~\cite{guo2019vsua} & 79.9 & 94.7 & 37.4 & 68.3 & 28.2 & 37.1 & 57.9 & 72.8 & 123.1 & 125.5 \\
\midrule
\name & \textbf{82.0} & \textbf{96.7} & \textbf{40.1} & \textbf{73.2} & \textbf{29.8} & \textbf{39.5} & \textbf{59.9} & \textbf{75.2} & \textbf{129.9} & \textbf{132.8}  \\
\bottomrule
\end{tabular}
\caption{\small{Results on COCO dataset. We report the single model results on the COCO online test server. We highlight the \textbf{best} model in bold.}}
\label{tab:coco_online}
\end{table*}

\section{Extra Experiments}
We evaluate the model on the COCO online test server, composed of 40 775 images for which annotations are not made publicly available.
The MSCOCO online testing results are
listed in Tab.~\ref{tab:coco_online}, our \name outperforms previous transformer based model on several evaluation metrics.

\bibliography{eacl2021}
\bibliographystyle{acl_natbib}



\maketitle



\begin{abstract}
    This supplementary material provides more information about the two datasets \good and \nyt as well as the training details.
\end{abstract}


\section{Dataset}
The Visual Genome dataset~\cite{Krishna17} consists of $51$ relation types, shown in Tab.~\ref{tab:vg_rel}.

\begin{table}[!p]
\caption{List of relation labels in Visual Genome dataset.}
\centering
\begin{tabular}{cp{8cm}cl}
\toprule
background, above, across, against, along, and, at,\\
attached to, behind, belonging to, between, carrying, \\
covered in, covering, eating, flying in, for, from, \\
growing on, hanging from, has, holding, in, in front of, \\
laying on, looking at, lying on, made of, mounted on, \\
near, of, on, on back of, over, painted on, parked on, \\
part of, playing, riding, says, sitting on, standing on, \\
to, under, using, walking in, walking on, watching, \\
wearing, wears, with \\
\bottomrule
\end{tabular}
\label{tab:vg_rel}
\end{table}

\begin{table*}[!t]
\centering
\begin{tabular}{ccccccccccl}
\toprule
\multirow{2}{*}{Method} & \multicolumn{2}{c|}{B-1}  & \multicolumn{2}{c|}{B-4} & \multicolumn{2}{c|}{M} & \multicolumn{2}{c|}{R} & \multicolumn{2}{c|}{C}   \\
\cmidrule{2-11}
&c5 & c40 &c5 & c40 &c5 & c40 &c5 & c40 &c5 & c40  \\
\midrule
Up-Down~\cite{Anderson2018} &  80.2 & 95.2 & 36.9 & 68.5 & 27.6 & 36.7 & 57.1 & 72.4 & 117.9 & 120.5 \\
Att2all~\cite{Rennie2017} &  78.1 & 93.7 & 35.2 & 64.5 & 27.0 & 35.5 & 56.3 & 70.7 & 114.7 & 116.7 \\
AoA~\cite{huang2019attention} &  81.0 & 95.0 & 39.4 & 71.2 & 29.1 & 38.5 & 58.9 & 74.5 & 126.9 & 129.6 \\
\midrule
ETA~\cite{Li_2019_ICCV} & 81.2 & 95.0 & 38.9 & 70.2 & 28.6 & 38.0 & 58.6 & 73.9 & 122.1 & 124.4  \\
Image-T~\cite{He2020image} & 81.2 & 95.4 & 39.6 & 71.5 & 29.1 & 38.4 & 59.2 & 74.5 & 127.4 & 129.6 \\
$\mathcal{M}^2$-T~\cite{cornia2020m2} & 81.6 & 96.0 & 39.7 & 72.8 & 29.4 & 39.0 & 59.2 & 74.8 & 129.3 & 132.1 \\
\midrule
GCN~\cite{Yao2018} & 80.8 & 95.9 & 38.7 & 69.7 & 28.5 & 37.6 & 58.5 & 73.4 & 125.3 & 126.5 \\
SGAE~\cite{yang2019} &  -- -- & -- -- & 37.8 & 68.7 & 28.1 & 37.0 & 58.2 & 73.1 & 122.7 & 125.5 \\
VSUA~\cite{guo2019vsua} & 79.9 & 94.7 & 37.4 & 68.3 & 28.2 & 37.1 & 57.9 & 72.8 & 123.1 & 125.5 \\
\midrule
\name & \textbf{82.0} & \textbf{96.7} & \textbf{40.1} & \textbf{73.2} & \textbf{29.8} & \textbf{39.5} & \textbf{59.9} & \textbf{75.2} & \textbf{129.9} & \textbf{132.8}  \\
\bottomrule
\end{tabular}
\caption{\small{Results on COCO dataset. We report the single model results on the COCO online test server. We highlight the \textbf{best} model in bold.}}
\label{tab:coco_online}
\end{table*}

\section{Extra Experiments}
We evaluate the model on the COCO online test server, composed of 40 775 images for which annotations are not made publicly available.
The MSCOCO online testing results are
listed in Tab.~\ref{tab:coco_online}, our \name outperforms previous transformer based model on several evaluation metrics.

\bibliography{eacl2021}
\bibliographystyle{acl_natbib}


\title{Supplementary Material for \\
ReFormer: The Relational Transformer for Image Captioning}

\author{First Author\\
Institution1\\
Institution1 address\\
{\tt\small firstauthor@i1.org}
\and
Second Author\\
Institution2\\
First line of institution2 address\\
{\tt\small secondauthor@i2.org}
}

\maketitle
\ificcvfinal\thispagestyle{empty}\fi

\begin{abstract}
    This supplementary material provides more information about the Visual Genome dataset and the online evaluation on COCO test server.
\end{abstract}

\section{Dataset}
The Visual Genome dataset~\cite{Krishna17} consists of $51$ relation types, shown in Tab.~\ref{tab:vg_rel}.

\begin{table}[!p]
\caption{List of relation labels in Visual Genome dataset.}
\centering
\begin{tabular}{cp{8cm}cl}
\toprule
background, above, across, against, along, and, at,\\
attached to, behind, belonging to, between, carrying, \\
covered in, covering, eating, flying in, for, from, \\
growing on, hanging from, has, holding, in, in front of, \\
laying on, looking at, lying on, made of, mounted on, \\
near, of, on, on back of, over, painted on, parked on, \\
part of, playing, riding, says, sitting on, standing on, \\
to, under, using, walking in, walking on, watching, \\
wearing, wears, with \\
\bottomrule
\end{tabular}
\label{tab:vg_rel}
\end{table}

\begin{table*}[!t]
\centering
\begin{tabular}{ccccccccccl}
\toprule
\multirow{2}{*}{Method} & \multicolumn{2}{c|}{B-1}  & \multicolumn{2}{c|}{B-4} & \multicolumn{2}{c|}{M} & \multicolumn{2}{c|}{R} & \multicolumn{2}{c|}{C}   \\
\cmidrule{2-11}
&c5 & c40 &c5 & c40 &c5 & c40 &c5 & c40 &c5 & c40  \\
\midrule
Up-Down~\cite{Anderson2018} &  80.2 & 95.2 & 36.9 & 68.5 & 27.6 & 36.7 & 57.1 & 72.4 & 117.9 & 120.5 \\
Att2all~\cite{Rennie2017} &  78.1 & 93.7 & 35.2 & 64.5 & 27.0 & 35.5 & 56.3 & 70.7 & 114.7 & 116.7 \\
AoA~\cite{huang2019attention} &  81.0 & 95.0 & 39.4 & 71.2 & 29.1 & 38.5 & 58.9 & 74.5 & 126.9 & 129.6 \\
\midrule
ETA~\cite{Li_2019_ICCV} & 81.2 & 95.0 & 38.9 & 70.2 & 28.6 & 38.0 & 58.6 & 73.9 & 122.1 & 124.4  \\
Image-T~\cite{He2020image} & 81.2 & 95.4 & 39.6 & 71.5 & 29.1 & 38.4 & 59.2 & 74.5 & 127.4 & 129.6 \\
$\mathcal{M}^2$-T~\cite{cornia2020m2} & 81.6 & 96.0 & 39.7 & 72.8 & 29.4 & 39.0 & 59.2 & 74.8 & 129.3 & 132.1 \\
\midrule
GCN~\cite{Yao2018} & 80.8 & 95.9 & 38.7 & 69.7 & 28.5 & 37.6 & 58.5 & 73.4 & 125.3 & 126.5 \\
SGAE~\cite{yang2019} &  -- -- & -- -- & 37.8 & 68.7 & 28.1 & 37.0 & 58.2 & 73.1 & 122.7 & 125.5 \\
VSUA~\cite{guo2019vsua} & 79.9 & 94.7 & 37.4 & 68.3 & 28.2 & 37.1 & 57.9 & 72.8 & 123.1 & 125.5 \\
\midrule
\name & \textbf{82.0} & \textbf{96.7} & \textbf{40.1} & \textbf{73.2} & \textbf{29.8} & \textbf{39.5} & \textbf{59.9} & \textbf{75.2} & \textbf{129.9} & \textbf{132.8}  \\
\bottomrule
\end{tabular}
\caption{\small{Results on COCO dataset. We report the single model results on the COCO online test server. We highlight the \textbf{best} model in bold.}}
\label{tab:coco_online}
\end{table*}

\section{Extra Experiments}
We evaluate the model on the COCO online test server, composed of 40 775 images for which annotations are not made publicly available.
The MSCOCO online testing results are
listed in Tab.~\ref{tab:coco_online}, our \name outperforms previous transformer based model on several evaluation metrics.

{\small
\bibliographystyle{ieee_fullname}
\bibliography{egbib}
}